\def\year{2021}\relax
\documentclass[letterpaper]{article} 
\usepackage{aaai21}  
\usepackage{times}  
\usepackage{helvet} 
\usepackage{courier}  
\usepackage[hyphens]{url}  
\usepackage{graphicx} 
\usepackage[numbers]{natbib}
\usepackage{caption} 
\usepackage{multirow}
\usepackage[table,xcdraw]{xcolor}
\usepackage{graphicx}
\usepackage{relsize}
\usepackage{amsmath}
\usepackage{amssymb}
\usepackage{amsfonts}
\usepackage{subcaption}
\usepackage{bbm}
\usepackage{bm}
\usepackage{placeins}
\usepackage{xcolor}
\definecolor{navyblue}{rgb}{0.0, 0.0, 0.5}
\usepackage{booktabs,tabularx}
\newcolumntype{C}{>{\centering\arraybackslash}X}

\title{Clinical Risk Prediction with Temporal Probabilistic \\ Asymmetric Multi-Task Learning}

\author{
  A. Tuan Nguyen \thanks{Equal contribution} \thanks{This work was done while the author is in KAIST} \textsuperscript{\rm 1, \rm 4}, 
  Hyewon Jeong \footnotemark[1] \textsuperscript{\rm 1},
  Eunho Yang \textsuperscript{\rm 1, \rm 2, \rm 3},
  Sung Ju Hwang \textsuperscript{\rm 1, \rm 2, \rm 3}
  \thanks{Correspondence to: A. Tuan Nguyen, Hyewon Jeong, and Sung Ju Hwang.} \\
}
\affiliations{
    \textsuperscript{\rm 1} School of Computing, 
    \textsuperscript{\rm 2} AI Graduate School, 
    Korea Advanced Institute of Science and Technology\\
    \textsuperscript{\rm 3} Aitrics,
    \textsuperscript{\rm 4} Department of Computer Science, University of Oxford\\
    tuan.nguyen@cs.ox.ac.uk, \{jhw162, eunhoy, sjhwang82\}@kaist.ac.kr \\
}

\begin{document}

\maketitle

\begin{abstract} 
Although recent multi-task learning methods have shown to be effective in improving the generalization of deep neural networks, they should be used with caution for safety-critical applications, such as clinical risk prediction. This is because even if they achieve improved task-average performance, they may still yield degraded performance on individual tasks, which may be critical (e.g., prediction of mortality risk). Existing asymmetric multi-task learning methods tackle this \emph{negative transfer} problem by performing knowledge transfer from tasks with low loss to tasks with high loss. However, using loss as a measure of reliability is risky since low loss could result from overfitting. In the case of time-series prediction tasks, knowledge learned for one task (e.g., predicting the sepsis onset) at a specific timestep may be useful for learning another task (e.g., prediction of mortality) at a later timestep, but lack of loss at each timestep makes it challenging to measure the reliability at each timestep. To capture such dynamically changing asymmetric relationships between tasks in time-series data, we propose a novel temporal asymmetric multi-task learning model that performs knowledge transfer from certain tasks/timesteps to relevant uncertain tasks, based on the feature-level uncertainty. We validate our model on multiple clinical risk prediction tasks against various deep learning models for time-series prediction, which our model significantly outperforms without any sign of negative transfer. Further qualitative analysis of learned knowledge graphs by clinicians shows that they are helpful in analyzing the predictions of the model.
\end{abstract}
\section{Introduction}
\label{introduction}
Multi-task learning (MTL)~\cite{caruana1997multitask} is a method to train a model, or multiple models jointly for multiple tasks to obtain improved generalization, by sharing knowledge among them. One of the most critical problems in multi-task learning is the problem known as \emph{negative transfer}, where unreliable knowledge from other tasks adversely affects the target task. This negative transfer could be fatal for safety-critical applications such as clinical risk prediction, where we cannot risk losing performance on any of the tasks. To prevent negative transfer, researchers have sought ways to allow knowledge transfer only among closely related tasks, by either identifying the task groups or learning optimal sharing structures among tasks \cite{duong2015low,misra2016cross}. However, it is not only the task relatedness that matters, but also the relative reliability of the task-specific knowledge. Recent asymmetric multi-task learning (AMTL) models~\cite{lee2016asymmetric,lee2017deep} tackle this challenge by allowing tasks with low loss to transfer more. 

\begin{figure*}
	\begin{subfigure}{.21\textwidth}
		\centering
		\includegraphics[width=\linewidth]{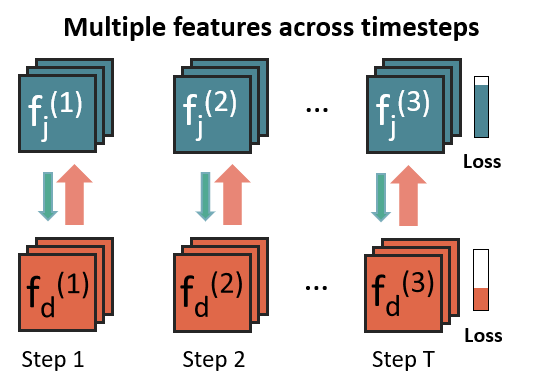}
		\caption{\footnotesize Loss-based AMTL}
		\label{AMTL-samestep}
	\end{subfigure}
	\begin{subfigure}{.2\textwidth}
		\centering
		\includegraphics[width=\linewidth]{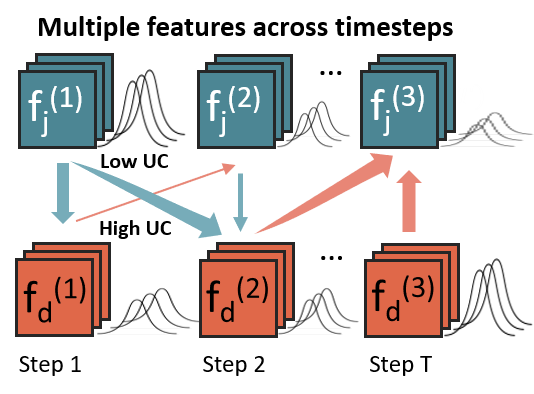}
		\caption{\footnotesize TPAMTL}
		\label{AMTL-ours}
	\end{subfigure}
	\begin{subfigure}{.2025\textwidth}
		\centering
		\includegraphics[width=\linewidth]{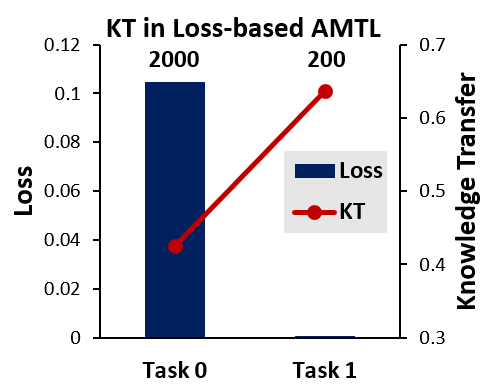}
		\caption{ }
		\label{AMTL-Failure}
	\end{subfigure}
	\begin{subfigure}{.16\textwidth}
		\centering
		\includegraphics[width=\linewidth]{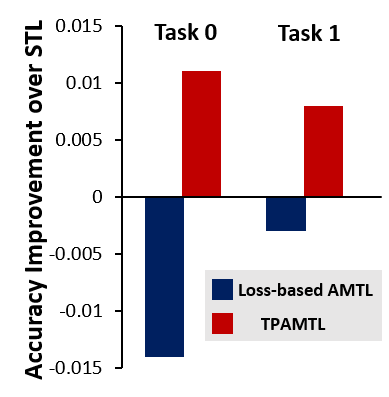}
		\caption{ }
		\label{acc}
	\end{subfigure}
	\begin{subfigure}{.2025\textwidth}
		\centering
		\includegraphics[width=\linewidth]{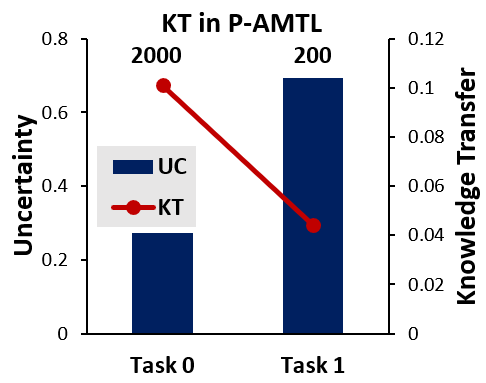}
		\caption{ }
		\label{PAMTL}
	\end{subfigure}
	\caption{\small \textbf{Concept:} \textbf{(a)} Existing AMTL models \cite{lee2016asymmetric, lee2017deep} utilize task loss to perform static knowledge transfer (KT) from one task to another; thus it cannot capture dynamically changing relationships between timesteps and tasks in the time-series domain. \textbf{(b)} Our model performs dynamic KT among tasks and across timesteps based on the feature-level uncertainty (UC). \textbf{(c-e)} Failure case of Loss-based AMTL (\textcolor{navyblue}{Table} \ref{Table1: mnist}), with a small, imbalanced number of training instances (Task 0: 2000 instances, Task1: 200 instances).} 
	\label{fig:attention}
\end{figure*}

While the asymmetric knowledge transfer between tasks is useful, it does not fully exploit the asymmetry in time-series analysis, which has an additional dimension of the time axis. With time-series data, knowledge transfer direction may need to be different depending on the timestep. For instance, suppose that we predict an event of infection or mortality within $48$ hours of admission in intensive care units (ICU) based on electronic health records (EHR). At earlier timesteps, prediction of \emph{Infection} may be more reliable than \emph{Mortality}, where we may want knowledge transfer to happen from task \emph{Infection} to \emph{Mortality}; at later timesteps, we may want the next situation to happen. Moreover, knowledge transfer may happen across timesteps. For example, a high risk of \emph{Infection} in early timestep will alert high risk of \emph{Mortality} at later timesteps. To exploit such temporal relationships between tasks, we need a model that does not perform static knowledge transfer between two tasks (\textcolor{navyblue}{Figure} \ref{AMTL-samestep}), but transfers knowledge across the timesteps of two different tasks, while dynamically changing the knowledge transfer amount and direction at each timestep (\textcolor{navyblue}{Figure} \ref{AMTL-ours}).  To this end, we propose a multi-task learning framework for time-series data, where each task not only learns its own latent features at each timestep but also leverages aggregated latent features from the other tasks at the same or different timesteps via attention allocation (\textcolor{navyblue}{Figure} \ref{model-architecture}). 

Yet this brings in another challenge. On what basis should we promote asymmetric knowledge transfer? For asymmetric knowledge transfer between tasks, we could use task loss as a proxy of knowledge reliability~\cite{lee2016asymmetric,lee2017deep}. However, loss is not a direct measure of reliability, as loss might not be available at every step for time-series prediction. Also, a model trained with few instances (Task 1 in \textcolor{navyblue}{Figure} \ref{AMTL-Failure} - \ref{PAMTL}) may have a small loss and thus transfer more knowledge to other tasks (\textcolor{navyblue}{Figure}~\ref{AMTL-Failure}), but the knowledge from this model could be highly biased and unreliable as it may have overfitted (\textcolor{navyblue}{Figure}~\ref{acc}). Thus, we propose a novel probabilistic Bayesian framework for asymmetric knowledge transfer, which leverages feature-level \emph{uncertainty}, instead of task loss, to measure the reliability of the knowledge (\textcolor{navyblue}{Figure}~\ref{AMTL-ours}). Basically, if a latent feature learned at a certain timestep has large uncertainty, our model will allocate small attention values for the feature; that is, the attention will be attenuated based on the uncertainty, where knowledge transfers from the task with low uncertainty to high uncertainty (\textcolor{navyblue}{Figure}~\ref{PAMTL}; Please see \textcolor{navyblue}{Table} \ref{Table1: mnist}).

We experimentally validate our \emph{Temporal Probabilistic Asymmetric Multi-Task Learning (TP-AMTL)} model on \textbf{four clinical risk prediction datasets} against multiple baselines. The results show that our model obtains significant improvements over strong multi-task learning baselines with no \textbf{negative transfer} on any of the tasks (\textcolor{navyblue}{Table} \ref{infection-1000}). We further show that both the asymmetric knowledge transfer between tasks at two different timesteps and the uncertainty-based attenuation of attention weights are effective in improving generalization. Finally, with the actual knowledge transfer graph plotted with uncertainty obtained for each timestep, we could interpret the model behaviors according to actual clinical events in clinical risk prediction tasks (\textcolor{navyblue}{Figure}~\ref{transfer-uc}, \textcolor{navyblue}{Supplemenatary Figure} 9, 10). This interpretability makes it more suitable for clinical risk prediction in real-world situations.

Our contribution is threefold: 

\begin{itemize}

	\item We propose a \textbf{novel probabilistic formulation for asymmetric knowledge transfer}, where the amount of knowledge transfer depends on the feature level uncertainty.

    \item We extend the framework to \textbf{an asymmetric multi-task learning framework for time-series analysis}, which utilizes feature-level uncertainty to perform knowledge transfer among tasks and across timesteps, thereby exploiting both the task-relatedness and temporal dependencies.
	
	\item We validate our model on \textbf{four clinical risk prediction tasks} against ten baselines, which it significantly outperforms with no \textbf{negative transfer}. With the help of clinicians, we further analyze the learned knowledge transfer graph to discover meaningful relationships between clinical events.
	
\end{itemize}

\section{Related Work}

\subsection{Multi-task Learning} 
While the literature on multi-task learning \cite{caruana1997multitask, argyriou2008convex} is vast, we selectively mention the prior works that are closely related to ours. Historically, multi-task learning models have focused on \emph{what to share} \cite{yang2016deep, yang2016trace, ruder2017learning}, as the jointly learned models could share instances, parameters, or features \cite{kang2011learning, kumar2012learning, maurer2013sparse}. With deep learning, multi-task learning can be implemented rather straightforwardly by making multiple tasks to share the same deep network. However, since solving different tasks will require diversified knowledge, complete sharing of the underlying network may be suboptimal and brings in a problem known as \emph{negative transfer}, where certain tasks are negatively affected by knowledge sharing. To prevent this, researchers are exploring more effective knowledge sharing structures. Soft parameter sharing method with regularizer \cite{duong2015low} can enforce the network parameters for each task to be similar, while a method to learn the optimal combination of shared and task-specific representations is also proposed \cite{misra2016cross} in computer vision. Losses can be weighed based on the uncertainty of the task in a multi-task framework \cite{kendall2018multi}, reducing negative transfer from uncertain tasks. While finding a good sharing structure can alleviate negative transfer, negative transfer will persist if we perform symmetric knowledge transfer among tasks. To resolve this symmetry issue, the asymmetric MTL model with inter-task knowledge transfer \cite{lee2016asymmetric} was proposed, which allows task-specific parameters for tasks with smaller loss to transfer more. \citet{lee2017deep} proposed a model for asymmetric task-to-feature transfer that allows reconstructing features with task-specific features while considering their loss, which is more suitable for deep neural networks and scalable. 

\subsection{Clinical time-series analysis} 
While our method is generic and applicable to any time-series prediction task, we mainly focus on clinical time-series analysis in this paper. Multiple clinical benchmark datasets \cite{citi2012physionet, johnson2016mimic} have been released and publicly available. Also, several recent works have proposed clinical prediction benchmarks with publicly available datasets \cite{che2018recurrent, harutyunyan2019multitask, johnson2017reproducibility, pirracchio2016mortality, purushotham2017benchmark}. We construct our datasets and tasks specific to our problem set (\textcolor{navyblue}{Experiments} section), in part referring to previous benchmark tasks. Furthermore, there has been some progress on this topic recently, mostly focusing on the interpretability and reliability of the model. An attention-based model \cite{choi2016retain} that generates attention for both the timesteps (hospital visits) and features (medical examination results) was proposed to provide interpretations of the predictions. However, attentions are often unreliable since they are learned in a weakly-supervised manner, and a probabilistic attention mechanism \cite{heo2018uncertainty} was also proposed to obtain reliable interpretation and prediction that considers uncertainty as to how to trust the input. Our work shares the motivation with these prior works as we target interpretability and reliability. Recently, SAnD~\cite{song2018attend}) proposes to utilize a self-attention architecture for clinical prediction tasks, and Adacare~\cite{ma2020adacare} proposes scale-adaptive recalibration module with dilated convolution to capture time-series biomarker features. Yet, they are inherently susceptible to negative transfer as all tasks share a single base network (\textcolor{navyblue}{Table}~\ref{infection-1000}). 

\section{Approach}

\subsection{Probabilistic Asymmetric Multi-Task Learning}

In this section, we describe our framework of probabilistic asymmetric multi-task learning (P-AMTL) in a general setting. Suppose that we have $D$ tasks with datasets $\{\mathbf{X}_d,\mathbf{Y}_d\}_{d=1}^D$, in which the sets $\mathbf{X}_1,\mathbf{X}_2,...,\mathbf{X}_D$ can be identical, overlapping or even disjoint. We further suppose that we have $D$ different probabilistic networks $\{p_d(.)\}_{d=1}^D$ , each of which generates high-level latent features of task $d$ (task-specific) via $\mathbf{Z}_d\sim p_d(\mathbf{X}_d)$. In a single-task learning setting, these latent features $\mathbf{Z}_d$ are in turn used to make predictions for task $d$. However, in our asymmetric multi-task learning framework, we want to borrow some learned features from other tasks to share knowledge and to improve generalization performance. Specifically, in order to perform prediction for task $d$, we leverage latent features learned from other tasks, $\mathbf{Z}_{j,d} \sim p_j(\mathbf{X}_d), \forall j \neq d$. Given the source features $\mathbf{Z}_{j,d}$ and the target features $\mathbf{Z}_{d}$, the model needs to decide on the following:

\textbf{1) The amount of knowledge to transfer} 
Existing asymmetric multi-task learning models \cite{lee2016asymmetric,lee2017deep} often use task loss to decide on the amount of knowledge transfer, in a way that tasks with low training loss are allowed to transfer more, while tasks with high loss only receive knowledge from other tasks. However, the task loss may not be a proper measure of the knowledge from the task and also unavailable in some cases (\textcolor{navyblue}{Figure} \ref{AMTL-Failure}, cases in \textcolor{navyblue}{Introduction}). To overcome these limitations, we propose to learn the amount of knowledge transfer based on the feature-level UC. Our model learns the transfer weight from $\mathbf{Z}_{j,d}$ to $\mathbf{Z}_d$ by a small network $F_{j,d}$ (Equation \ref{transfer-amount}). This learnable network takes both $\mathbf{Z}_{j,d}$, $\mathbf{Z}_d$ and their variance $\bm{\sigma}^2_{j,d}$ and $\bm{\sigma}^2_{d}$ as its input. Note that if the variance is not available from the output of $\{p_d(.)\}_{d=1}^D$ directly, we can perform Monte-Carlo sampling $k$ times on $\mathbf{Z}_{j,d}$ \text{ and } $\mathbf{Z}_d$ to compute the estimates of variances. In practice, we can implement each $F_{j,d}$ as a multi-layer perceptron with the input as the concatenation of $\mathbf{Z}_{j,d},\mathbf{Z}_d,\bm{\sigma}^2_{j,d} \text{ and }\bm{\sigma}^2_{d}$.
\begin{figure*}
	\centering
	\includegraphics[width=\textwidth]{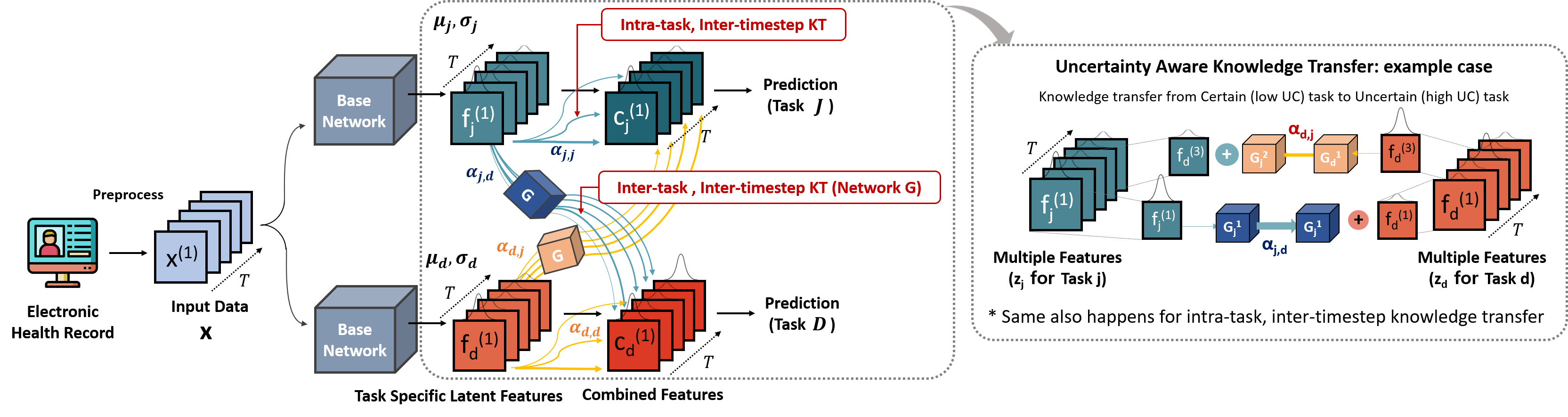}
	\caption{\small\textbf{Temporal probabilistic asymmetric knowledge transfer}. This figure illustrates how we apply the probabilistic asymmetric knowledge transfer between tasks at the same timestep and across different timesteps. (Right) Features of task $j$ at timestep $1$ is more reliable than features of task $d$ at timestep $1$, so the model will learn to transfer more from task $j$ to task $d$ and transfer less from task $d$ to task $j$.}
	\label{model-architecture}
\end{figure*}

\begin{equation}
\label{transfer-amount}
  \alpha_{j,d} = F_{j,d}(\mathbf{Z}_{j,d},\mathbf{Z}_d,\bm{\sigma}^2_{j,d},\bm{\sigma}^2_{d})
\end{equation}

\textbf{2) The form of transferred knowledge}
Since the learned features for different tasks may have completely different representations, directly adding $\alpha_{j,d}\mathbf{Z}_{j,d}$ to $\mathbf{Z}_d$ would be sub-optimal. For this combining process, we train two additional networks $G^1_k$ and $G^2_k$ for each task $k$ where $G^1_k$ is used to convert the learned task-specific features from $p_k(.)$ to a shared latent space and $G^2_k$ is used to convert the features from that shared latent space back to the task-specific latent space. Finally, we compute the combined feature map for task $d$ as (\textcolor{navyblue}{Figure}~\ref{model-architecture} \textcolor{navyblue}{(Right)}): 

\begin{equation}
  \mathbf{C}_d = \mathbf{Z}_d + G^2_d \left( \sum_{j\neq d}\alpha_{j,d}*G^1_j(\mathbf{Z}_{j,d})\right)
\end{equation}

The combined feature map $\mathbf{C}_d$ can then be used for the final prediction for task $d$. The combined feature maps for all other tasks are computed in the same manner.

\subsection{Temporal Probabilistic Asymmetric Multi-Task Learning}

We now apply our probabilistic asymmetric multi-task learning framework for the task of time-series prediction. 

Our goal is to jointly train time-series prediction models for multiple tasks at once. Suppose that we are given training data for $D$ tasks, $\mathbbm{D}=\{(\mathbf{X}_1, \mathbf{Y}_1), \dots, (\mathbf{X}_D, \mathbf{Y}_D)\}$. Further suppose that each data instance $\mathbf{x} \text{ where } \mathbf{x} \in \mathbf{X}_d$ for some task $d$, consists of $T$ timesteps. That is, $\mathbf{x} = (\mathbf{x}^{(1)}, \mathbf{x}^{(2)}, \dots, \mathbf{x}^{(T)})$, where $\mathbf{x}^{(t)} \in \mathbb{R}^{m}$ denotes the data instance for the timestep $t$. Here we assume the number of timesteps $T$ is identical across all tasks for simplicity, but there is no such restriction in our model. Additionally, $y_d$ is the label for task $d$; $y_d \in \{0, 1\}$ for binary classification tasks, and $y_d \in \mathbb{R}$ for regression tasks. Given time-series data and tasks, we want to learn the task-specific latent features for each task and timestep, and then perform asymmetric knowledge transfer between them. Our framework is comprised of the following components:

\paragraph{Shared Low-Level Layers} We allow our model to share low-level layers for all tasks in order to learn a common data representation before learning task-specific features. At the lowest layer, we have a shared linear data embedding layer to embed the data instance for each timestep into a continuous shared feature space. Given a time-series data instance $\mathbf{x}$, we first linearly transform the data point for each timestep $t$, $\mathbf{x}^{(t)} \in \mathbbm{R}^m$, which contains $m$ variables:

\begin{equation}
	(\mathbf{v}^{(1)},\mathbf{v}^{(2)},...,\mathbf{v}^{(T)}) = \mathbf{v} = \mathbf{x} \mathbf{W}_{emb} \in \mathbbm{R}^{T\times k} 
	\label{eq1}
\end{equation}
where $\mathbf{W}_{emb} \in \mathbbm{R}^{m\times k}$ and $k$ is the hidden dimension. This embedded input is then fed into shared RNN layer for pre-processing:

\begin{equation}
	\mathbf{r} = (\mathbf{r}^{(1)},\mathbf{r}^{(2)},...,\mathbf{r}^{(T)}) = RNN(\mathbf{v}^{(1)},\mathbf{v}^{(2)},...,\mathbf{v}^{(T)})
\end{equation}

\paragraph{Task- and Timestep Embedding Layers (`Base Network' in \textcolor{navyblue}{Figure}~\ref{model-architecture})} After embedding and pre-processing the input into a continuous space, we further encode them into task- and timestep-specific features. Since hard-sharing layers may result in negative transfer between tasks, we allow \emph{separate} embedding layers for each task to encode task-specific knowledge in our `Base Network'. For each task $d$, the separate network consists of $L$ feed-forward layers, which learn disentangled knowledge for each timestep. These $L$ feed-forward layers for task embedding can be formulated as:
\begin{small}
    \begin{equation}
    \mathbf{h}_d = \sigma((...\sigma(\sigma(\mathbf{r}\mathbf{W}^1_d+\mathbf{b}^1_d)\mathbf{W}^2_d+\mathbf{b}^2_d)...)\mathbf{W}^L_d+\mathbf{b}^L_d) \in \mathbbm{R}^{T\times k}
    \end{equation}
\end{small}
where $\mathbf{W}_d^i \in \mathbbm{R}^{k\times k}, \mathbf{b}_d^i \in \mathbbm{R}^{k}$ and $\sigma$ is a non-linear activation function (e.g. leaky relu).

\paragraph{Modeling feature-level uncertainty} While the embedding above can capture knowledge for each task and timestep, we want to further model their uncertainties as well, to measure the reliability of the knowledge captured. Towards this objective, we model the latent variables as probabilistic random variables, with two types of UC \cite{kendall2017uncertainties}: 1) \emph{epistemic uncertainty}, which comes from the model's unreliability from the lack of training data, and 2) \emph{aleatoric uncertainty}, that comes from the inherent ambiguity in the data. We capture the former by using dropout variational inference \cite{gal2016dropout}, and the latter by explicitly learning the model variance as a function of the input (\textcolor{navyblue}{Figure}~\ref{model-architecture}).
Suppose that we have the task-specific latent features: $\mathbf{z}_d \sim p(\mathbf{z}_d|\mathbf{x},\bm{\omega}) \text{ where } \bm{\omega}$ is the set of all parameters. This can be formulated as below:
\begin{align}
    \mathbf{z}_d|\mathbf{x},\bm{\omega} \sim \mathcal{N}\left(\mathbf{z}_d;\bm{\mu}_d,diag(\bm{\sigma}^2_d)\right)\\
    \bm{\mu}_d = \sigma(\mathbf{h}_d\mathbf{W}_d^\mu+\mathbf{b}_d^\mu)\\
    \bm{\sigma}_d = softplus(\mathbf{h}_d\mathbf{W}_d^\sigma+\mathbf{b}_d^\sigma)
\end{align}
As mentioned before, we use dropout approximation \cite{gal2016dropout} with parameter \textbf{$M$} as the variational distribution $q_M(\bm{\omega})$ to approximate $p(\bm{\omega}|\mathbbm{D})$.

\paragraph{Asymmetric knowledge transfer across tasks and time steps}
\label{apply-kt}
Now we apply the proposed probabilistic asymmetric knowledge transfer method to perform knowledge transfer across timesteps, both within each task and across tasks, to exploit intra- and inter-task temporal dependencies. In order to transfer knowledge from task $j$ to task $d$ with temporal dependencies, we allow the latent features of task $d$ at time step $t$ ($\mathbf{f}_d^{(t)}$, with $\mathbf{z}_d = (\mathbf{f}_d^{(1)},\mathbf{f}_d^{(2)},...,\mathbf{f}_d^{(T)})$) to obtain knowledge from task $j$ at all previous time steps (\textcolor{navyblue}{Figure}~\ref{model-architecture}), and then combine them into a single feature map $\mathbf{C}_d^{(1)},\mathbf{C}_d^{(2)},...,\mathbf{C}_d^{(T)}$:

\begin{small}
    \begin{equation}
      \mathbf{C}_d^{(t)}=\mathbf{f}_d^{(t)}+G^2_{d}\left(\mathlarger{\sum}^D_{j=1} \mathlarger{\sum}_{i=1}^t
      \alpha^{(i,t)}_{j,d}*G^1_{j}\left(\mathbf{f}_j^{(i)}\right)\right) \forall t 
    \end{equation}    
\end{small}
As mentioned in the previous subsection, the transfer weight $\alpha^{(i,t)}_{j,d}$ is computed by a network $F_{j,d}$ with input
$\mathbf{f}_d^{(t)}$, $\mathbf{f}_j^{(i)}$ and their variance $\bm{\sigma}_d^{(t)}$, $\bm{\sigma}_j^{(i)}$ (again, we perform MC sampling to get the variance).

Here, we choose to constrain the knowledge transfer to happen only from past to future timesteps because of the time complexity at inference time. With our proposed model, for \textbf{each update} at the clinical environment in an online manner, we only need to transfer the knowledge from previous time steps to the current one, making the complexity to be $\mathbf{O(T)}$. This is on a par with other models like RETAIN \cite{choi2016retain} or UA \cite{heo2018uncertainty}, making it highly scalable (\textcolor{navyblue}{Table} 1 of \textbf{Supplementary File}). However, if we allow the knowledge to transfer from future timestep to past timestep, we also need to update the knowledge at previous timesteps for a single update. The time complexity of the model in this case is $O(T^2)$, which is undesirable. In the ablation study, we show that this constraint also brings in a small performance gain. The total complexity of the \textbf{whole} training or inference is still $O(T^2)$ due to the inter-timestep transfer, but this is on par with state-of-the-art models such as Transformer \cite{vaswani2017attention} and SAnD \citep{song2018attend}.

Finally, we use the combined features $\mathbf{C}_d^{(1)}, \mathbf{C}_d^{(2)}$,...,$\mathbf{C}_d^{(T)}$, which contain temporal dependencies among tasks, for prediction for each task $d$. We use an attention mechanism
\begin{equation}
	\bm{\beta}_d^{(t)}=tanh\left( \mathbf{C}_d^{(t)}\mathbf{W}_d^\beta+\mathbf{b}_d^\beta\right) \quad\forall t \in \{1,2,...,T\}
\end{equation}
where $\mathbf{W}_d^\beta \in \mathbbm{R}^{k\times k}$ and $\mathbf{b}_d^\beta \in \mathbbm{R}^k$. Then the model can perform prediction as
\begin{small}
    \begin{equation}
    	p_d=p(\widehat{y_d}|\mathbf{x})=Sigmoid\left(\frac{1}{T}\left(\mathlarger{\sum}^T_{t=1}\bm{\beta}_d^{(t)}\odot \mathbf{v}^{(t)}\right)\mathbf{W}_d^o+b_d^o\right)
    \end{equation}   
\end{small}

for classification tasks, where $\odot$ denotes the element-wise multiplication between attention ${\beta_d}^{(t)}$ and shared input embedding $\mathbf{v}^{(t)}$ (from \textcolor{navyblue}{Eq.}~\ref{eq1}), $\mathbf{W}_d^o \in \mathbbm{R}^{k\times 1}$ and $b_d^o \in \mathbbm{R}^1$. Predictions for other tasks are done similarly. Note that our model does not require each instance to have the labels for every tasks. We can simply maximize the likelihood $p(y_d|\mathbf{x})$ whenever the label $y_d$ is available for input $x$ for task $d$. Furthermore, our model does not require the instances to have the same number of timesteps $T$.

The loss function of our objective function is:
\begin{small}
    \begin{align}
        \sum_{x,y} &\left[-\sum_{d\in Ta(x)} (y_d \log p_d + (1-y_d)\log (1-p_d))\right]+\beta_{w\_decay}||\theta||_2^2
    \end{align}
\end{small}
where $Ta(x)$ is the set of tasks which we have the labels available for data instance $x$, $\beta_{w\_decay}$ is the coefficient for weight decay and $\theta$ is the whole parameters of our model.



\section{Experiments \footnote{For the \textbf{1)} details on the baselines, base network configurations, and hyper-parameters, \textbf{2)} details and experimental results (quantitative and qualitative interpretation) on two datasets (\textbf{MIMIC III - Heart Failure} and \textbf{Respiratory Failure}), please see the \textbf{supplementary file}. These additional result further supports our model and shows that our model also generalize well to various, larger dataset.}}

\subsection{Probabilistic Asymmetric Multi-task Learning}
\label{section: p-amtl}

We first validate the effectiveness of the uncertainty-based knowledge transfer for asymmetric multi-task learning, using a non-temporal version of our model. We use a variant \cite{mnistvar} of the MNIST dataset \cite{lecun-mnisthandwrittendigit-2010} which contains images of handwritten digits $0$-$9$ with random rotations and background noise. From this dataset, we construct $5$ tasks; each task is a binary classification of determining whether the given image belongs to class $0$-$4$. We sample 5000, 5000, 1000, 1000, and 500 examples for task 0, 1, 2, 3, and 4 respectively, such that asymmetric knowledge transfer becomes essential to obtain good performance on all tasks. As for the base network, we use a multi-layer perceptron, which outputs mean and variance of the task-specific latent features.\\
\textbf{1) Single Task Learning (STL)} learns an independent model for each task. \\
\textbf{2) Multi Task Learning (MTL)} learns a single base network with $5$ disjoint output layers for the $5$ tasks. \\
\textbf{3) AMTL-Loss.} A model that is similar to P-AMTL, with the transfer weight from task $j$ to task $d$ learned by a network $F_{j,d}$ with the average task loss over all instances as the input. \\
\textbf{4) P-AMTL.} Our probabilistic asymmetric MTL model.

Results (\textcolor{navyblue}{Table}~\ref{Table1: mnist}) show that MTL outperforms STL, but suffers from negative transfer (Task 4, highlighted in \textcolor{red}{red} (\textcolor{navyblue}{Table}~\ref{Table1: mnist})). AMTL-Loss underperforms MTL, which shows that the loss is not a good measure of reliability; a model that is overfitted to a task will have a small loss, but its knowledge may be unreliable (\textcolor{navyblue}{Figure}\ref{AMTL-Failure}). Finally, our model outperforms all baselines without any sign of negative transfer, demonstrating the superiority of uncertainty-based knowledge transfer.

\begin{table}[ht!]
	\footnotesize
	\caption{AUROC for the MNIST-Variation Experiment}
 	\label{Table1: mnist}
	\centering
	\resizebox{\columnwidth}{!}{
		\begin{tabular}{c|ccccc|c}
			\toprule
			Models & Task 0 & Task 1 & Task 2 & Task 3 & Task 4  &  Average \\
			\midrule
			STL  & 0.7513$\pm$0.02 & \textbf{0.7253$\pm$0.01} & \textbf{0.5401$\pm$0.01} & 0.5352$\pm$0.02 & 0.6639$\pm$0.01 & 0.6432$\pm$0.01 \\
			MTL & 0.8266$\pm$0.01 & \textcolor{red}{0.7021$\pm$0.01} & \textbf{0.5352$\pm$0.01} & \textbf{0.5987$\pm$0.01} & \textcolor{red}{0.6203$\pm$0.02} &0.6565$\pm$0.01 \\
			AMTL-Loss & \textcolor{red}{0.7317$\pm$0.02} & \textbf{0.7236$\pm$0.01} & \textcolor{red}{0.5309$\pm$0.01} & 0.5166$\pm$0.02 & 0.6698$\pm$0.01 & \textcolor{red}{0.6345$\pm$0.01} \\
    		\midrule
		    \textbf{P-AMTL (ours)} & \textbf{0.8469$\pm$0.01} & \textbf{0.7267$\pm$0.01} & \textbf{0.5382$\pm$0.01} & \textbf{0.5950$\pm$0.01} & \textbf{0.6822$\pm$0.01} & \textbf{0.6778$\pm$0.01} \\
			\bottomrule
		\end{tabular}
}
\end{table}

\subsection{Clinical risk prediction from EHR}
\label{section: tp-amtl}
Clinical risks can be defined in various ways, but in this paper, we define \emph{clinical risks} as the existence of the event (e.g., Heart Failure, Respiratory Failure, Infection, Mortality) that may lead to deterioration of patients' health condition within a given time window (e.g., $48$ hour).

\subsubsection{Tasks and Datasets  }
We experiment on four datasets where we compile for clinical risk prediction from two open-source EHR datasets. Every dataset used in this paper contain tasks with clear temporal dependencies between them. The input features have been pre-selected by the clinicians, since the excessive features were not clinically meaningful, and may harm the prediction performance. Missing feature values are imputed with zero values. Pleare refer to the \textbf{supplementary file} for the explanation and evaluation on two datasets \textbf{MIMIC III - Heart Failure} and \textbf{Respiratory Failure}.

\textbf{1) MIMIC III - Infection.} We compile a dataset out of the MIMIC III dataset~\cite{johnson2016mimic}, which contains records of patients admitted to ICU of a hospital. We use records of patients over the age $15$, where we hourly sample to construct $48$ timesteps from the first $48$ hours of admission. Following clinician's guidelines, we select $12$ infection-related variables for the features at each timestep (See \textcolor{navyblue}{Table} 8, 9 of \textbf{supplementary file}). Tasks considered for this dataset are the clinical events before and after infection; \emph{Fever} (Task 1) as the sign of infection with elevated body temperature, \emph{Infection} (Task 2) as the confirmation of infection by the result of microbiology tests, and finally, \emph{Mortality} (Task 3) as a possible outcome of infection (See \textcolor{navyblue}{Figure} 1 of \textbf{supplementary file} for the task configuration). After pre-processing, approximately $2000$ data points with a sufficient amount of features were selected, which was randomly split to approximately $1000/500/500$ for training/validation/test. 

\textbf{2) PhysioNet} \cite{citi2012physionet}. Total of $4,000$ ICU admission records were included, each containing 48 hours of records (sampled hourly) and $31$ physiological signs including variables displayed in \textcolor{navyblue}{Table} \ref{Clinicalevents}. We select $29$ infection-related variables for the features available at each timestep (See \textbf{supplementary file}). Task used in the experiment includes four binary classification tasks, namely, 1) \emph{Stay$<3$}: whether the patient would stay in ICU for less than three days, 2) \emph{Cardiac:} whether the patient is recovering from cardiac surgery, 3) \emph{Recovery:} whether the patient is staying in Surgical ICU to recover from surgery, and 4) \emph{Mortality prediction (Mortality)} (See \textcolor{navyblue}{Figure} 2 of \textbf{supplementary file} for the task configuration). We use a random split of $2800/400/800$ for training/validation/test.

\subsubsection{Baselines} We compared the single- and multi- task learning baselines to see the effect of negative transfer. Please see the \textbf{supplementary file} for descriptions of the baselines, experimental details, and the hyper-parameters used.

\vspace{5pt}
\noindent \textbf{Single Task Learning (STL) baselines} \\
\textbf{1) STL-LSTM}. The single-task learning method with long short-term memory network to capture temporal dependencies. \\
\textbf{2) STL-Transformer}. Similar STL setting with \textbf{1)}, but with Transformer~\cite{vaswani2017attention} as the base network. \\
\textbf{3) STL-RETAIN}~\cite{choi2016retain}.  The attentional RNN for interpretability of clinical prediction with EHR.\\
\textbf{4) STL-UA}~\cite{heo2018uncertainty}. Uncertainty-Aware probabilistic attention model. \\
\textbf{5) STL-SAnD}~\cite{choi2016retain}. Self-attention model for multi-task time series prediction. \\
\textbf{6) STL-AdaCare}~\cite{ma2020adacare} Feature-adaptive modeling with squeeze and excitation block, based on dilated convolution network.

\vspace{5pt}
\noindent \textbf{Multi Task Learning (MTL) baselines.} \\
MTL baselines are the naive hard-sharing multi-task learning method where all tasks share the same network except for the separate output layers for prediction.\\
\textbf{7) MTL-LSTM, 8) MTL-Transformer, 9) MTL-RETAIN, 10) MTL-UA, 11) MTL-SAnD, 12) AdaCare} Multi-task learning setting with 7) LSTM, 8) Transformer~\cite{vaswani2017attention}, 9) RETAIN~\cite{choi2016retain}, 10) UA \cite{heo2018uncertainty}, 11) SAnD \cite{song2018attend}, 12) AdaCare~\cite{ma2020adacare} as the base network, respectively. \\
\textbf{13) AMTL-LSTM}~\cite{lee2016asymmetric}. This learns the knowledge transfer graph between task-specific parameters shared across all timesteps with static knowledge transfer between tasks based on the task loss (\textcolor{navyblue}{Figre}~\ref{AMTL-samestep}).\\
\textbf{14) MTL-RETAIN-Kendall}~\cite{kendall2018multi}. The uncertainty-based loss-weighing scheme with base MTL-RETAIN. \\
\textbf{15) TP-AMTL.} Our probabilistic temporal AMTL model that performs both intra- and inter-task knowledge transfer. 

\subsubsection{Quantitative evaluation}

\begin{table*}[ht]
 \small
 \centering
 \caption{\small Task performance on the MIMIC-III Infection and PhysioNet dataset. We report average AUROC and standard error over five runs (MTL model accuracies lower than those of their STL counterparts are colored in \textcolor{red}{red}).} 
 \label{infection-1000}
 \centering	
 \resizebox{\textwidth}{!}{
 	\begin{tabular}{cccccc|ccccc}
 		\toprule
 		\multicolumn{2}{c}{} & \multicolumn{4}{c|}{MIMIC-III Infection} & \multicolumn{5}{c}{PhysioNet} \\
 		\cmidrule(r){3-6}
 		\cmidrule(r){7-11}
 		& Models & Fever & Infection & Mortality  & Average  & Stay $< 3$    & Cardiac   & Recovery   & Mortality  & Average\\
 		\midrule
 		\multicolumn{1}{c|}{} 
 		& LSTM
 		& 0.6738 $\pm$ 0.02 
 		& 0.6860 $\pm$ 0.02  
 		& 0.6373 $\pm$ 0.02
 		& 0.6657 $\pm$ 0.02
		& 0.7673 $\pm$ 0.09 
		& 0.9293 $\pm$ 0.01  
		& 0.8587 $\pm$ 0.01 
		& 0.7100 $\pm$ 0.01  
		& 0.8163 $\pm$ 0.03
 		\\
		\multicolumn{1}{c|}{} 
		& Transformer \cite{vaswani2017attention} 
		&\textbf{0.7110 $\pm$ 0.01} 
		& 0.6500 $\pm$ 0.01 
		& 0.6766 $\pm$ 0.01 
		& 0.6792 $\pm$ 0.01
		& \textbf{0.8953 $\pm$ 0.01} 
		& 0.9283 $\pm$ 0.02 
		& 0.8721 $\pm$ 0.01
		& 0.6796 $\pm$ 0.02
		& 0.8380 $\pm$ 0.01 
		\\
 		\multicolumn{1}{c|}{STL} 
 		& RETAIN \cite{choi2016retain}
 		& 0.6826 $\pm$ 0.01 
 		& 0.6655 $\pm$ 0.01  
 		& 0.6054 $\pm$ 0.02  
 		& 0.6511 $\pm$ 0.01
		& 0.7407 $\pm$ 0.04  
		& 0.9236 $\pm$ 0.01  
		& 0.8148 $\pm$ 0.04 
		& 0.7080 $\pm$ 0.02  
		& 0.7968 $\pm$ 0.03
 		\\
  		\multicolumn{1}{c|}{} 
  		& UA \cite{heo2018uncertainty}
  		& 0.6987 $\pm$ 0.02 
  		& 0.6504 $\pm$ 0.02  
  		& 0.6168 $\pm$ 0.05  
  		& 0.6553 $\pm$ 0.02
		& 0.8556 $\pm$ 0.02 
		& 0.9335 $\pm$ 0.01 
		& 0.8712 $\pm$ 0.01 
		& 0.7283 $\pm$ 0.01 
		& 0.8471 $\pm$ 0.01
  		\\
        \multicolumn{1}{c|}{}
        & SAnD \cite{song2018attend}
        & 0.6958 $\pm$ 0.02
        & 0.6829 $\pm$ 0.01    
        & 0.7073 $\pm$ 0.02
        & 0.6953 $\pm$ 0.01
 		& \textbf{0.8965 $\pm$ 0.02} 
 		& 0.9369 $\pm$ 0.01  
 		& 0.8838 $\pm$ 0.01  
 		& 0.7330 $\pm$ 0.01 
 		& 0.8626 $\pm$ 0.01
        \\
   		\multicolumn{1}{c|}{} 
  		& AdaCare \cite{ma2020adacare}
  		& 0.6354 $\pm$ 0.02
  		& 0.6256 $\pm$ 0.03
  		& 0.6217 $\pm$ 0.01
  		& 0.6275 $\pm$ 0.08
  		& 0.7508 $\pm$ 0.06
  		& 0.8610 $\pm$ 0.01
  		& 0.7700 $\pm$ 0.03
   		& 0.6595 $\pm$ 0.02 
  		& 0.7603 $\pm$ 0.07 
  		\\
 		\midrule
 		\multicolumn{1}{c|}{} 
 		& LSTM 
 		& 0.7006 $\pm$ 0.03 
 		& 0.6686 $\pm$ 0.02  
 		& 0.6261 $\pm$ 0.03 
 		& 0.6651 $\pm$ 0.02
		& 0.7418 $\pm$ 0.09
		& 0.9233 $\pm$ 0.01  
		& 0.8472 $\pm$ 0.02 
		& 0.7228 $\pm$ 0.01  
		& \textcolor{red}{0.8088 $\pm$ 0.03}
 		\\
 		\multicolumn{1}{c|}{} 
 		& Transformer\cite{vaswani2017attention}  
 		& 0.7025 $\pm$ 0.01 
 		& 0.6479 $\pm$ 0.02  
 		& \textcolor{red}{0.6420 $\pm$ 0.02}  
 		& 0.6641 $\pm$ 0.02
		& \textcolor{red}{0.8532 $\pm$ 0.03} 
		& 0.9291 $\pm$ 0.01 
		& 0.8770 $\pm$ 0.01 
		& 0.7358 $\pm$ 0.01
		& 0.8488 $\pm$ 0.01
 		\\
 		\multicolumn{1}{c|}{} 
 		& RETAIN \cite{choi2016retain}  
 		& 0.7059 $\pm$ 0.02 
 		& 0.6635 $\pm$ 0.01  
 		& 0.6198 $\pm$ 0.05  
 		& 0.6630 $\pm$ 0.02
		& 0.7613 $\pm$ 0.03  
		& \textcolor{red}{0.9064 $\pm$ 0.01}  
		& 0.8160 $\pm$ 0.04 
		& 0.6944 $\pm$ 0.03  
		& 0.7945 $\pm$ 0.03
 		\\
   		\multicolumn{1}{c|}{MTL} 
   		& UA  \cite{heo2018uncertainty}
   		& \textbf{0.7124 $\pm$ 0.01} 
   		& 0.6489 $\pm$ 0.02  
   		& 0.6325 $\pm$ 0.04  
   		& 0.6646 $\pm$ 0.02
		& 0.8573 $\pm$ 0.03 
		& 0.9348 $\pm$ 0.01 
		& 0.8860 $\pm$ 0.01 
		& \textbf{0.7569 $\pm$ 0.02}
		& 0.8587 $\pm$ 0.02
   		\\
 		\multicolumn{1}{c|}{}
 	    &  SAnD \cite{song2018attend}
        &  0.7041 $\pm$ 0.01
        &  0.6818 $\pm$ 0.02
        &  0.6880 $\pm$ 0.01
        &  0.6913 $\pm$ 0.01
 		&  0.8800 $\pm$ 0.03
 		& \textbf{0.9410 $\pm$ 0.00} 
 		& \textcolor{red}{0.8607 $\pm$ 0.01} 
 		& \textbf{0.7612 $\pm$ 0.02} 
 		& 0.8607 $\pm$ 0.06
        \\
   		\multicolumn{1}{c|}{} 
  		& AdaCare \cite{ma2020adacare}
  		& \textcolor{red}{0.5996 $\pm$ 0.01}
  		& 0.6163 $\pm$ 0.02
  		& 0.6283 $\pm$ 0.01  
  		& 0.6148 $\pm$ 0.00
  		& 0.8746 $\pm$ 0.01
  		& \textcolor{red}{0.7211 $\pm$ 0.01}
  		& \textcolor{red}{0.6348 $\pm$ 0.02}
  		& 0.7457 $\pm$ 0.03
  		& \textcolor{red}{0.7440 $\pm$ 0.08}
  		\\
 		\multicolumn{1}{c|}{} 
 		& AMTL-LSTM\cite{lee2016asymmetric} 
 		& 0.6858 $\pm$ 0.01 
 		& 0.6773 $\pm$ 0.01  
 		& 0.6765 $\pm$ 0.01  
 		& 0.6798 $\pm$ 0.01
		& 0.7600 $\pm$ 0.08    
		& 0.9254 $\pm$ 0.01   
		& 0.8066 $\pm$ 0.01 
		& 0.7167 $\pm$ 0.01    
		& 0.8022 $\pm$ 0.03 
 		\\
  		\multicolumn{1}{c|}{} 
  		& RETAIN-Kendall \cite{kendall2018multi}
  		& 0.6938 $\pm$ 0.01 
  		& \textcolor{red}{0.6182 $\pm$ 0.03}  
  		& 0.5974 $\pm$ 0.02  
  		& 0.6364 $\pm$ 0.02
		& 0.7418 $\pm$ 0.02 
		& 0.9219 $\pm$ 0.02 
		& 0.7883 $\pm$ 0.03 
		& \textcolor{red}{0.6787 $\pm$ 0.02} 
		& \textcolor{red}{0.7827 $\pm$ 0.02}
  		\\
  		\midrule
 		& TP-AMTL (our model) 
 		& \textbf{0.7081 $\pm$ 0.01}    
 		& \textbf{0.7173 $\pm$ 0.01} 
 		& \textbf{0.7112 $\pm$ 0.01} 
 		& \textbf{0.7102 $\pm$ 0.01}
 		& \textbf{0.8953 $\pm$ 0.01}    
		& \textbf{0.9416 $\pm$ 0.01} 
		& \textbf{0.9016 $\pm$ 0.01} 
		& \textbf{0.7586 $\pm$ 0.01} 
		& \textbf{0.8743 $\pm$ 0.01}
 		\\
 		\bottomrule
 	\end{tabular}
 }
\end{table*}
\begin{figure*}[ht]
	\begin{subfigure}{.5\textwidth}
		\centering
		\includegraphics[width=\linewidth]{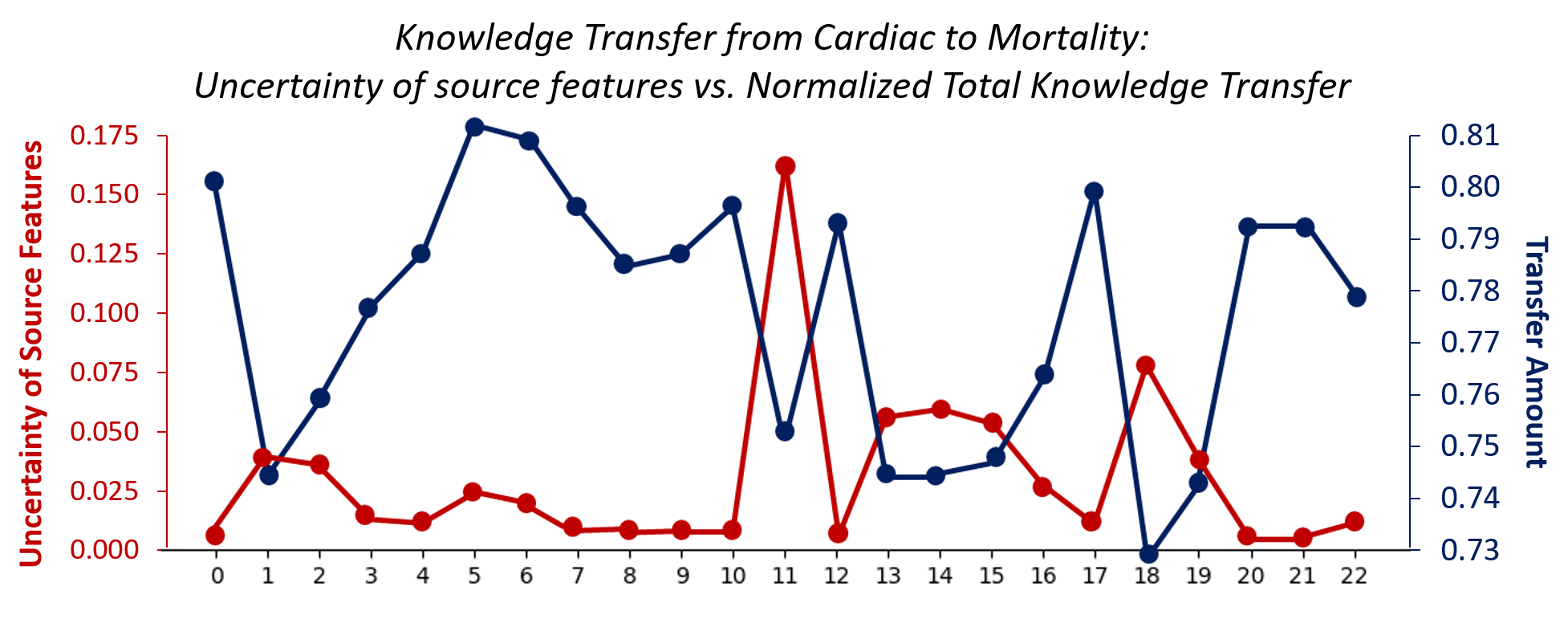}
		\captionsetup{justification=centering,margin=0.5cm}
		\caption{\footnotesize Outgoing Transfer from different Sources}
		\label{transfer-source}
	\end{subfigure}%
	\begin{subfigure}{.5\textwidth}
		\centering
		\includegraphics[width=\linewidth]{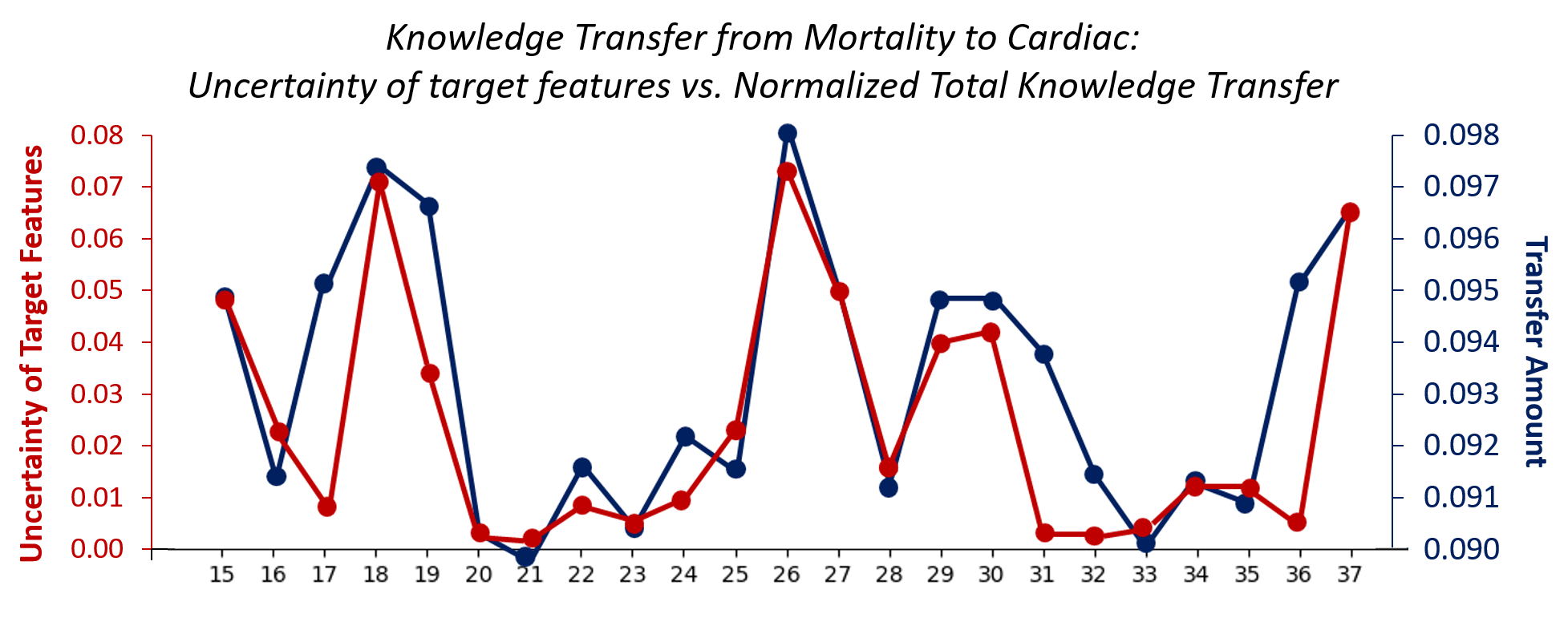}
		\captionsetup{justification=centering,margin=0.5cm}
		\caption{\footnotesize Incoming Transfer to different Targets}
		\label{transfer-target}
	\end{subfigure}
	\caption{\footnotesize Examples showing the relationship between the amount of KT and UC of source and target features. (a) The sources with low UC transfer more knowledge. (b) The targets with high UC receive more knowledge.}
	\label{transfer-uc}
\end{figure*}

We first evaluate the prediction accuracy of the baseline STL and MTL models and ours on the four clinical time-series datasets, by measuring the Area Under the ROC curve (AUROC) (MIMIC-III Infection and PhysioNet (\textcolor{navyblue}{Table}~\ref{infection-1000})). We observe that hard-sharing MTL models outperform STL on some tasks, but suffers from performance degeneration on others (highlighted in \textcolor{red}{red} in \textcolor{navyblue}{Table}~\ref{infection-1000}), which shows a clear sign of  negative transfer. MTL models especially work poorly on MIMIC-III infection, which has clear temporal relationships between tasks. Probabilistic models (e.g., \textbf{UA} \cite{heo2018uncertainty}) generally outperform their deterministic counterparts (e.g., \textbf{RETAIN} \cite{choi2016retain}). However, \textbf{MTL-RETAIN-Kendall} \cite{kendall2018multi}, which learns the weight for each task loss based on uncertainty, significantly underperforms even the \textbf{STL-LSTM}, which may be due to the fact that losses in our settings are at almost similar scale unlike with the task losses in~\cite{kendall2018multi} that have largely different scales. Although the self-attention based model~\textbf{SAnD} \cite{song2018attend} shows impressive performance on some of the tasks from  PhysioNet, it also suffers from performance degeneration in the MTL setting, resulting in lower overall performance. \textbf{AMTL-LSTM} \cite{lee2016asymmetric} improves on some tasks, but degenerates the performance on the others, which we attribute to the fact that it does not consider inter-timestep transfer. Additionally, \textbf{AdaCare} with dilated convolution showed severely degenerated performance except for one task. On the other hand, our model, \textbf{TP-AMTL}, obtains significant improvements over all STL and MTL baselines on both datasets. It also does not show performance degeneration on any of the tasks, suggesting that it has successfully dealt away with negative transfer in multi-task learning with time-series prediction models. Experimental results on \textbf{ablation study} (regarding inter-, intra-task, future-to-past knowledge transfer, various uncertainty types)  and additional two datasets (\textbf{MIMIC III - Heart Failure} and \textbf{Respiratory Failure}) are available in the \textbf{supplementary file}, which further supports our model and shows that our model also generalize well to various, larger datasets.

To further analyze the relationships between uncertainty and knowledge transfer, we visualize knowledge transfer from multiple sources (\textcolor{navyblue}{Figure}~\ref{transfer-source}) normalized over the number of targets, and to multiple targets (\textcolor{navyblue}{Figure}~\ref{transfer-target}) normalized over the number of sources, along with their uncertainties. Specifically, the uncertainty of a task at a certain timestep is represented by the average of the variance of all feature distributions. The normalized amount of knowledge transfer from task $j$ at time step $t$ to task $d$ is computed as $\scriptstyle{(\alpha^{(t,t)}_{j,d} + \alpha^{(t,t+1)}_{j,d} + ... + \alpha^{(t,T)}_{j,d})/(T-t+1)}$. Similarly, the normalized amount of knowledge transfer to task $d$ at time step $t$ from task $j$ is $\scriptstyle{(\alpha^{(1,t)}_{j,d} + \alpha^{(2,t)}_{j,d} + ... + \alpha^{(t,t)}_{j,d})/t}$. We observe that source features with low uncertainty transfer knowledge more, while at the target, features with high uncertainty receive more knowledge transfer. However, note that they are not perfectly correlated, since the amount of knowledge transfer is also affected by the pairwise similarities between the source and the target features as well. 

\subsubsection{Interpretations of the Transfer Weights}

\begin{table*}[ht!]
	\footnotesize
	\caption{\footnotesize Clinical Events in selected medical records for case studies. \textbf{SBP} - Systolic arterial blood pressure, \textbf{DBP} - Diastolic arterial blood pressure, \textbf{BT} - Body Temperature, \textbf{WBC} - White Blood Cell Count, \textbf{FiO2} - Fractional inspired Oxygen, \textbf{BUN} - Blood Urine Nitrogen}
	\label{Clinicalevents}
	\hspace{-0.2in}
	\resizebox{\textwidth}{!}{%
		\begin{tabular}{lll|llllllll|llllllllll}
			&  &  & SBP & DBP & BT & WBC & Culture Results &  &  &  & SBP & DBP & Temp & FiO2 & Lactate & $HCO_3^-$ & BUN & Creatinine & &  \\ \cline{3-8} \cline{11-20}
			&  & 1:00 & 100 & 53 & 40.1 & 12500 & N/A &  &  & \textbf{7:38} & \textbf{N/A} & \textbf{37} & \textbf{37} & \textbf{0.35} & \textbf{5.3} & \textbf{N/A} & \textbf{N/A} & \textbf{N/A} & &\\
			&  & \textbf{2:57} & \textbf{89} & \textbf{46} & \textbf{N/A} & \textbf{N/A} & \textit{\textbf{(+)Klebsiella Pneumoniae}} &  &  & 8:38 & 140 & 55 & 36.6 & N/A & N/A & \textbf{10.6} & \textbf{85} & \textbf{4.2} & & \\
			&  & 5:00 & 120 & 64 & N/A & N/A & N/A & &  & 9:38 & 142 & 42 & N/A & 0.4 & 6.1 & N/A & N/A & N/A & &  
		\end{tabular}
	}
\end{table*}
\begin{figure*}[ht!]
	\begin{subfigure}{.5\textwidth}
		\centering
		\includegraphics[width=\linewidth]{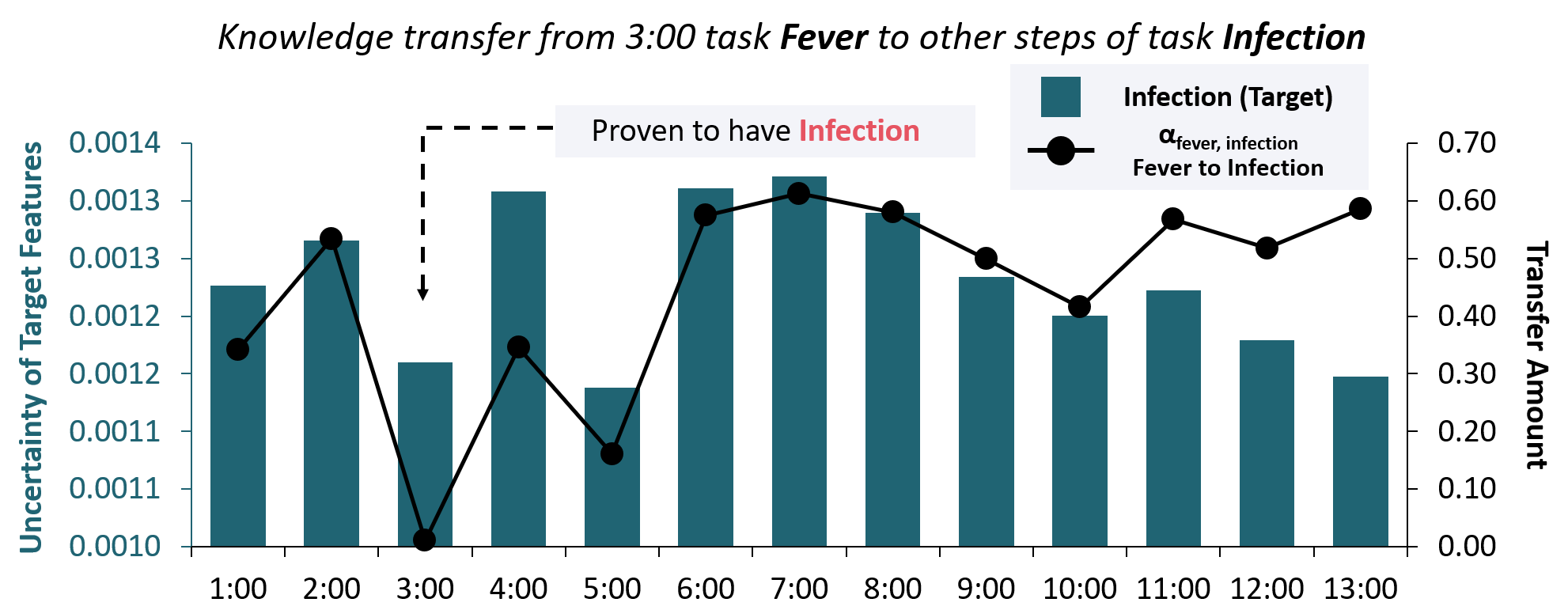}
		\caption{\footnotesize MIMIC-III Infection}
		\label{Knowledge_Transfer_MIMIC}
	\end{subfigure}%
	\begin{subfigure}{.5\textwidth}
		\centering
		\includegraphics[width=\linewidth]{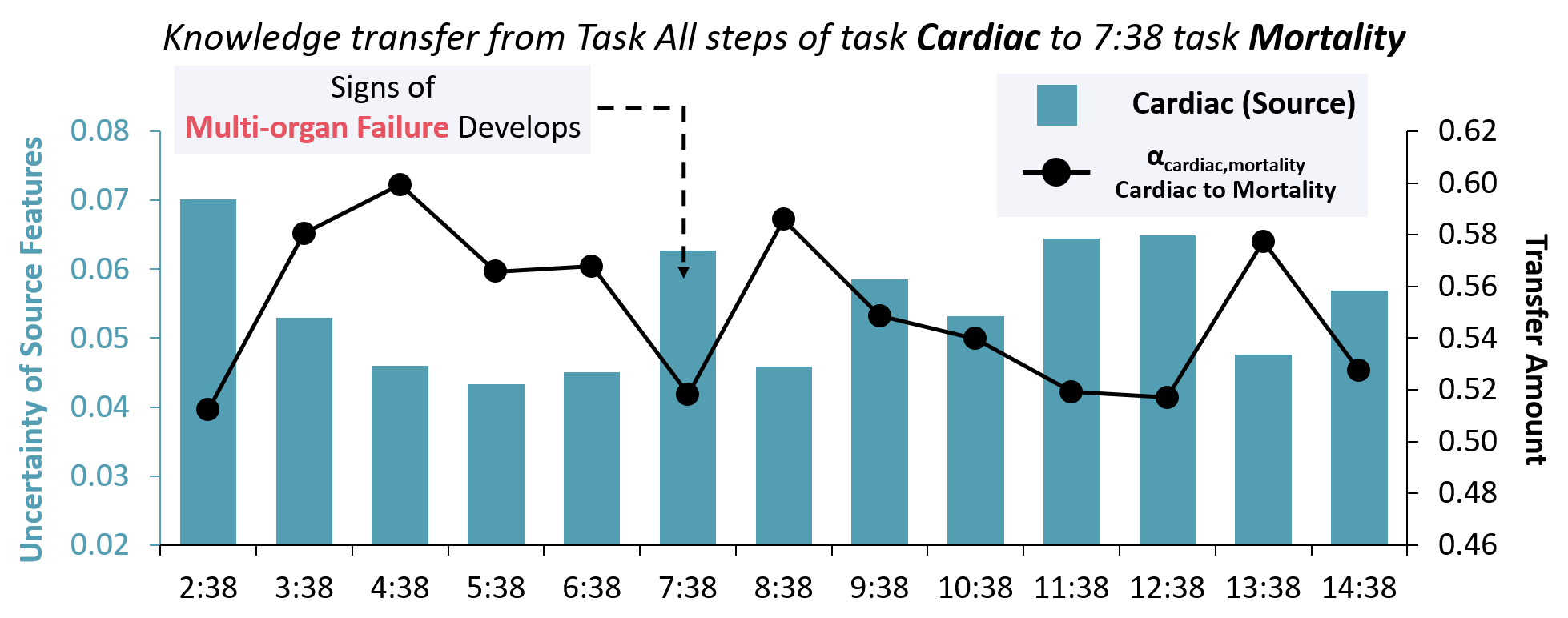}
		\caption{\footnotesize PhysioNet}
		\label{Knowledge_Transfer_Physionet}
	\end{subfigure}
	\vspace{-0.1in}
	\caption{\footnotesize \textbf{Visualizations of the amount of UC and normalized KT for example cases} where the trends of both UC and KT at certain timesteps are correlated with noticeable clinical events (indicated with dotted arrow).}
	\label{fig:Interpretation}
\end{figure*}

With the help of a physician, we further analyze how transfer weights and uncertainty are related with the patient's actual medical conditions (see \textcolor{navyblue}{Table}~\ref{Clinicalevents} and \textcolor{navyblue}{Figure}~\ref{fig:Interpretation}). We first consider an example record of a patient from the MIMIC-III Infection dataset who was suspected of infection on admission, and initially had fever, which was confirmed to be the symptom of bacterial infection later. \textcolor{navyblue}{Figure}~\ref{Knowledge_Transfer_MIMIC} shows the amount of knowledge transfer from task \emph{Fever} at $3$:$00$ to all later timesteps of task \emph{Infection}. At this timestep, the patient's condition changes significantly. We observe that the patient had a fever, and the WBC level has increased to the state of leukocytosis, and both the SBP and DBP decrease over time. Most importantly, the patient is diagnosed to have an infection, as the culture results turns out to be positive for \textit{Klebsiella Pneumoniae} at 2:57. With the drop of uncertainty of the task \emph{Infection} around the time window where the event happens (dotted arrow in \textcolor{navyblue}{Figure}~\ref{Knowledge_Transfer_MIMIC}), the amount of knowledge transfer from \emph{Fever} to \emph{Infection} drops as well, as the knowledge from the source task becomes less useful.

As for another case study, we consider a record of a patient from PhyisoNet dataset who recovered from cardiac surgery and passed away during admission (\textcolor{navyblue}{Table}~\ref{Clinicalevents} and \textcolor{navyblue}{Figure}~\ref{Knowledge_Transfer_Physionet}). From \textcolor{navyblue}{Table}~\ref{Clinicalevents}, we observe that sign of multi-organ failure develops, as features related to respiratory ($FiO_2$, $HCO_3^-$, Lactate), renal (BUN, Creatinine), and cardiac (DBP) function deteriorates. As patient's condition after surgery gets worse, uncertainty of \emph{Cardiac} starts to decrease at later timesteps (dotted arrow in \textcolor{navyblue}{Figure}~\ref{Knowledge_Transfer_Physionet}) and knowledge transfer from \emph{Cardiac} to \emph{Mortality} increases as the uncertainty of the source task \emph{Cardiac} starts to drop, since the knowledge from the source task becomes more reliable. Therefore, by analyzing the learned knowledge graph using our model, we could identify timesteps where meaningful interactions occur between tasks. For more case study examples of \textbf{MIMIC-III Heart Failure} and \textbf{Respiratory Failure} datasets, please see the \textbf{supplementary file}.

\subsection{Ablation Study}
\label{sec:ablation}

\begin{table*}[ht!]
 \footnotesize
 \centering
 \caption{\small Ablation Study result on the MIMIC-III Infection and PhysioNet Dataset.} 
 \label{ablation}
 \centering
 \resizebox{\textwidth}{!}{
 	\begin{tabular}{ccccc|ccccc}
 		\toprule
		& \multicolumn{4}{c|}{MIMIC-III Infection} & \multicolumn{5}{c}{PhysioNet} \\
		\cmidrule(r){2-5}
	\cmidrule(r){6-10}
		Model  & Fever & Infection & Mortality  & Average & Stay 3    & Cardiac   & Recovery   & Mortality   & Average\\
		\midrule
		AMTL-notransfer
		& 0.7048 $\pm$ 0.01
		& 0.6889 $\pm$ 0.01
		& 0.6969 $\pm$ 0.01
		& 0.6968 $\pm$ 0.01
		& \textbf{0.8901 $\pm$ 0.01}
		& 0.9212 $\pm$ 0.01
		& \textbf{0.8986 $\pm$ 0.01}
		& \textbf{0.7515 $\pm$ 0.01}
		& \textbf{0.8653 $\pm$ 0.01}
		\\
		AMTL-intratask
		& \textbf{0.7082 $\pm$ 0.01}
		& \textbf{0.7071 $\pm$ 0.01}   
		& 0.6814 $\pm$ 0.03
		& 0.6989 $\pm$ 0.01
		& 0.8829 $\pm$ 0.01    
		& \textbf{0.9338 $\pm$ 0.01}   
		& 0.8812 $\pm$ 0.01 
		& \textbf{0.7521 $\pm$ 0.01}    
		& 0.8625 $\pm$ 0.01
		\\
		AMTL-samestep
		& \textbf{0.7130 $\pm$ 0.00}  
		& 0.6946 $\pm$ 0.01 
		& 0.6983 $\pm$ 0.03
		& \textbf{0.7019 $\pm$ 0.01}
		& 0.8669 $\pm$ 0.01   
		& 0.9273 $\pm$ 0.01 
		& 0.8902 $\pm$ 0.01
		& 0.7382 $\pm$ 0.01    
		& 0.8557 $\pm$ 0.01
		\\
		TD-AMTL
		& 0.6636 $\pm$ 0.03    
		& 0.6874 $\pm$ 0.01   
		& 0.6953 $\pm$ 0.02 
		& 0.6821 $\pm$ 0.00
		& 0.7381 $\pm$ 0.06    
		& 0.9155 $\pm$ 0.01   
		& 0.8629 $\pm$ 0.01 
		& 0.7365 $\pm$ 0.01
		& 0.8133 $\pm$ 0.02
		\\
		TP-AMTL (constrained)
		& \textbf{0.7159 $\pm$ 0.00} 
		& \textbf{0.7003 $\pm$ 0.01}
		& 0.6561 $\pm$ 0.01 
		& 0.6908 $\pm$ 0.00
		& \textbf{0.8999 $\pm$ 0.01}    
		& 0.9186 $\pm$ 0.01   
		& 0.8892 $\pm$ 0.01 
		& \textbf{0.7610 $\pm$ 0.01} 
		& \textbf{0.8672 $\pm$ 0.00} 
		\\
		TP-AMTL (epistemic)
		& 0.6997 $\pm$ 0.01
		& \textbf{0.7196 $\pm$ 0.00} 
		& \textbf{0.7106 $\pm$ 0.02}
		& \textbf{0.7100 $\pm$ 0.01}
		& \textbf{0.8952 $\pm$ 0.01}    
		& \textbf{0.9341 $\pm$ 0.01}   
		& \textbf{0.8934 $\pm$ 0.01}
		& \textbf{0.7547 $\pm$ 0.01} 
		& \textbf{0.8693 $\pm$ 0.01} 
		\\
		TP-AMTL (aleatoric)
		& 0.6984 $\pm$ 0.02    
		& \textbf{0.7032 $\pm$ 0.01}
		& \textbf{0.7217 $\pm$ 0.02} 
		& 0.7078 $\pm$ 0.00
		& 0.8012 $\pm$ 0.03    
		& 0.9183 $\pm$ 0.01   
		& 0.8537 $\pm$ 0.02 
		& 0.7401 $\pm$ 0.03 
		& 0.8283 $\pm$ 0.01
		\\
		\midrule
		TP-AMTL (full model)
		& \textbf{0.7165 $\pm$ 0.00}    
		& \textbf{0.7093 $\pm$ 0.01}   
		& \textbf{0.7098 $\pm$ 0.01} 
		& \textbf{0.7119 $\pm$ 0.00}
		& \textbf{0.8953 $\pm$ 0.01}    
		& \textbf{0.9416 $\pm$ 0.01}   
		& \textbf{0.9016 $\pm$ 0.01} 
		& \textbf{0.7586 $\pm$ 0.01} 
		& \textbf{0.8743 $\pm$ 0.01} 
		\\
 		\bottomrule
 	\end{tabular}
 }
\vspace{-0.1in}
\end{table*}


We compare our model against several variations of our model with varying knowledge transfer direction with respect to task (inter-task) and time (inter-timestep), and with temporal constraint (only future-to-past). Also we examine two kinds of uncertainty (epistemic, aleatoric) with our model (\textcolor{navyblue}{Table}~\ref{ablation}).

\subsubsection{Inter-task and inter-timestep knowledge transfer.}
\hfill\\
\textbf{1) AMTL-notransfer:} The variant of our model without knowledge transfer. \\
\textbf{2) AMTL-intratask:} The variant of our model that knowledge only transfers within a same task. \\
\textbf{3) AMTL-samestep:} Another variant of our model that knowledge transfers only within a same time-step.\\
\textbf{4) TD-AMTL:} The deterministic counterpart of our model.

Our model outperforms \textbf{AMTL-intratask} and \textbf{AMTL-samestep}, which demonstrates the effectiveness of inter-task and inter-step knowledge transfer (\textcolor{navyblue}{Table}~\ref{ablation}). \textbf{TD-AMTL} largely underperforms any variants, which may be due to overfitting of the knowledge transfer model, that can be effectively prevented by our bayesian framework. 

\subsubsection{Future-to-past transfer.}
\hfill\\
\textbf{5) TP-AMTL (constrained):} the model with temporal constraint

The unconstrained model outperforms \textbf{TP-AMTL (constrained)} (\textcolor{navyblue}{Table}~\ref{ablation}), where transfer can only happen from the later timestep to earlier ones.

\subsubsection{Two kinds of uncertainty.}
\hfill\\
\textbf{6) TP-AMTL (epistemic)} uses only MC-dropout to model epistemic uncertainty and $p_\theta(\mathbf{z}_d|\mathbf{x},\bm{\omega})$ is simplified into $\mathcal{N}\left(\mathbf{z}_d;\bm{\mu}_d,\bm{0}\right)$ (i.e. its pdf becomes the dirac delta function at $\bm{\mu}_d$ and $\mathbf{z}_d$ is always $\bm{\mu}_d$)\\
\textbf{7) TP-AMTL (aleatoric)} uses only $p_\theta(\mathbf{z}_d|\mathbf{x},\bm{\omega})$ to model the aleatoric uncertainty, without MC-dropout.

For both MIMIC-III and PhysioNet datasets, epistemic uncertainty attributes more to the performance gain (\textcolor{navyblue}{Table}~\ref{ablation}). However, it should be noted that the impacts of two kinds of uncertainty vary from dataset to dataset. By modelling both kinds of uncertainty, the model is guaranteed to get the best performance.
\section{Conclusion}
We propose a novel probabilistic asymmetric multi-task learning framework that allows asymmetric knowledge transfer between tasks at different timesteps, based on the uncertainty. While existing asymmetric multi-task learning methods consider asymmetric relationships between tasks as fixed, the task relationship may change at different timesteps in time-series data. Moreover, knowledge obtained for a task at a specific timestep could be useful for other tasks in later timesteps. Thus, to model the varying direction of knowledge transfer and across-timestep knowledge transfer, we propose a novel probabilistic multi-task learning framework that performs knowledge transfer based on the uncertainty of the latent representations for each task and timestep. We validate our model on clinical time-series prediction tasks on four datasets, on which our model shows strong performance over the baseline symmetric and asymmetric multi-task learning models, without any sign of negative transfer. Several case studies with learned knowledge graphs show that our model is interpretable, providing useful and reliable information on model predictions. This interpretability of our model will be useful in building a safe time-series analysis system for large-scale settings where both the number of time-series data instances and timestep are extremely large, such that manual analysis is impractical.
\section{Acknowledgement}
This work was supported by the Institute for Information \& communications Technology Planning \& Evaluation (IITP) grant funded by the Korea government (MSIT) (No.2017-0-01779, A machine learning and statistical inference framework for explainable artificial intelligence(XAI)), the Engineering Research Center Program through the National Research Foundation of Korea (NRF) funded by the Korean Government MSIT (NRF-2018R1A5A1059921), and the Institute of Information \& communications Technology Planning \& Evaluation (IITP) grant funded by the Korea government (MSIT) (No.2019-0-00075, Artificial Intelligence Graduate School Program (KAIST)).

\bibliography{TPAMTL}

\begin{thebibliography}{38}
\providecommand{\natexlab}[1]{#1}
\providecommand{\url}[1]{\texttt{#1}}
\providecommand{\urlprefix}{URL }
\expandafter\ifx\csname urlstyle\endcsname\relax
  \providecommand{\doi}[1]{doi:\discretionary{}{}{}#1}\else
  \providecommand{\doi}{doi:\discretionary{}{}{}\begingroup
  \urlstyle{rm}\Url}\fi

\bibitem[{Argyriou, Evgeniou, and Pontil(2008)}]{argyriou2008convex}
Argyriou, A.; Evgeniou, T.; and Pontil, M. 2008.
\newblock Convex multi-task feature learning.
\newblock \emph{Machine Learning} 73(3): 243--272.

\bibitem[{Caruana(1997)}]{caruana1997multitask}
Caruana, R. 1997.
\newblock Multitask learning.
\newblock \emph{Machine learning} 28(1): 41--75.

\bibitem[{Che et~al.(2018)Che, Purushotham, Cho, Sontag, and
  Liu}]{che2018recurrent}
Che, Z.; Purushotham, S.; Cho, K.; Sontag, D.; and Liu, Y. 2018.
\newblock Recurrent neural networks for multivariate time series with missing
  values.
\newblock \emph{Scientific reports} 8(1): 1--12.

\bibitem[{Choi et~al.(2016)Choi, Bahadori, Sun, Kulas, Schuetz, and
  Stewart}]{choi2016retain}
Choi, E.; Bahadori, M.~T.; Sun, J.; Kulas, J.; Schuetz, A.; and Stewart, W.
  2016.
\newblock Retain: An interpretable predictive model for healthcare using
  reverse time attention mechanism.
\newblock In \emph{Advances in Neural Information Processing Systems},
  3504--3512.

\bibitem[{Citi and Barbieri(2012)}]{citi2012physionet}
Citi, L.; and Barbieri, R. 2012.
\newblock PhysioNet 2012 Challenge: Predicting mortality of ICU patients using
  a cascaded SVM-GLM paradigm.
\newblock In \emph{2012 Computing in Cardiology}, 257--260. IEEE.

\bibitem[{Duong et~al.(2015)Duong, Cohn, Bird, and Cook}]{duong2015low}
Duong, L.; Cohn, T.; Bird, S.; and Cook, P. 2015.
\newblock Low resource dependency parsing: Cross-lingual parameter sharing in a
  neural network parser.
\newblock In \emph{Proceedings of the 53rd Annual Meeting of the Association
  for Computational Linguistics and the 7th International Joint Conference on
  Natural Language Processing (Volume 2: Short Papers)}, volume~2, 845--850.

\bibitem[{Gal and Ghahramani(2016)}]{gal2016dropout}
Gal, Y.; and Ghahramani, Z. 2016.
\newblock Dropout as a bayesian approximation: Representing model uncertainty
  in deep learning.
\newblock In \emph{international conference on machine learning}, 1050--1059.

\bibitem[{Guo et~al.(2020)Guo, Fan, Chen, Wu, Zhang, He, Wang, Wan, Wang, and
  Lu}]{guo2020cardiovascular}
Guo, T.; Fan, Y.; Chen, M.; Wu, X.; Zhang, L.; He, T.; Wang, H.; Wan, J.; Wang,
  X.; and Lu, Z. 2020.
\newblock Cardiovascular implications of fatal outcomes of patients with
  coronavirus disease 2019 (COVID-19).
\newblock \emph{JAMA cardiology} .

\bibitem[{Harutyunyan et~al.(2019)Harutyunyan, Khachatrian, Kale, Ver~Steeg,
  and Galstyan}]{harutyunyan2019multitask}
Harutyunyan, H.; Khachatrian, H.; Kale, D.~C.; Ver~Steeg, G.; and Galstyan, A.
  2019.
\newblock Multitask learning and benchmarking with clinical time series data.
\newblock \emph{Scientific data} 6(1): 1--18.

\bibitem[{Heo et~al.(2018)Heo, Lee, Kim, Lee, Kim, Yang, and
  Hwang}]{heo2018uncertainty}
Heo, J.; Lee, H.~B.; Kim, S.; Lee, J.; Kim, K.~J.; Yang, E.; and Hwang, S.~J.
  2018.
\newblock Uncertainty-aware attention for reliable interpretation and
  prediction.
\newblock In \emph{Advances in Neural Information Processing Systems},
  909--918.

\bibitem[{Johnson, Pollard, and Mark(2017)}]{johnson2017reproducibility}
Johnson, A.~E.; Pollard, T.~J.; and Mark, R.~G. 2017.
\newblock Reproducibility in critical care: a mortality prediction case study.
\newblock In \emph{Machine Learning for Healthcare Conference}, 361--376.

\bibitem[{Johnson et~al.(2016)Johnson, Pollard, Shen, Li-wei, Feng, Ghassemi,
  Moody, Szolovits, Celi, and Mark}]{johnson2016mimic}
Johnson, A.~E.; Pollard, T.~J.; Shen, L.; Li-wei, H.~L.; Feng, M.; Ghassemi,
  M.; Moody, B.; Szolovits, P.; Celi, L.~A.; and Mark, R.~G. 2016.
\newblock MIMIC-III, a freely accessible critical care database.
\newblock \emph{Scientific data} 3: 160035.

\bibitem[{Kang, Grauman, and Sha(2011)}]{kang2011learning}
Kang, Z.; Grauman, K.; and Sha, F. 2011.
\newblock Learning with Whom to Share in Multi-task Feature Learning.
\newblock In \emph{ICML}, volume~2, 4.

\bibitem[{Kendall and Gal(2017)}]{kendall2017uncertainties}
Kendall, A.; and Gal, Y. 2017.
\newblock What uncertainties do we need in bayesian deep learning for computer
  vision?
\newblock In \emph{Advances in neural information processing systems},
  5574--5584.

\bibitem[{Kendall, Gal, and Cipolla(2018)}]{kendall2018multi}
Kendall, A.; Gal, Y.; and Cipolla, R. 2018.
\newblock Multi-task learning using uncertainty to weigh losses for scene
  geometry and semantics.
\newblock In \emph{Proceedings of the IEEE Conference on Computer Vision and
  Pattern Recognition}, 7482--7491.

\bibitem[{Kumar and Daume~III(2012)}]{kumar2012learning}
Kumar, A.; and Daume~III, H. 2012.
\newblock Learning task grouping and overlap in multi-task learning.
\newblock \emph{arXiv preprint arXiv:1206.6417} .

\bibitem[{LeCun and Cortes(2010)}]{lecun-mnisthandwrittendigit-2010}
LeCun, Y.; and Cortes, C. 2010.
\newblock {MNIST} handwritten digit database.
\newblock http://yann.lecun.com/exdb/mnist/.
\newblock \urlprefix\url{http://yann.lecun.com/exdb/mnist/}.

\bibitem[{Lee, Yang, and Hwang(2016)}]{lee2016asymmetric}
Lee, G.; Yang, E.; and Hwang, S. 2016.
\newblock Asymmetric multi-task learning based on task relatedness and loss.
\newblock In \emph{International Conference on Machine Learning}, 230--238.

\bibitem[{Lee, Yang, and Hwang(2017)}]{lee2017deep}
Lee, H.~B.; Yang, E.; and Hwang, S.~J. 2017.
\newblock Deep Asymmetric Multi-task Feature Learning.
\newblock \emph{arXiv preprint arXiv:1708.00260} .

\bibitem[{Li et~al.(2020)Li, Qin, Xu, Yin, Wang, Kong, Bai, Lu, Fang, Song
  et~al.}]{li2020artificial}
Li, L.; Qin, L.; Xu, Z.; Yin, Y.; Wang, X.; Kong, B.; Bai, J.; Lu, Y.; Fang,
  Z.; Song, Q.; et~al. 2020.
\newblock Artificial intelligence distinguishes COVID-19 from community
  acquired pneumonia on chest CT.
\newblock \emph{Radiology} 200905.

\bibitem[{Ma et~al.(2020)Ma, Gao, Wang, Zhang, Wang, Ruan, Tang, Gao, and
  Ma}]{ma2020adacare}
Ma, L.; Gao, J.; Wang, Y.; Zhang, C.; Wang, J.; Ruan, W.; Tang, W.; Gao, X.;
  and Ma, X. 2020.
\newblock AdaCare: Explainable Clinical Health Status Representation Learning
  via Scale-Adaptive Feature Extraction and Recalibration.
\newblock In \emph{Proceedings of the AAAI Conference on Artificial
  Intelligence}, volume~34, 825--832.

\bibitem[{Maurer, Pontil, and Romera-Paredes(2013)}]{maurer2013sparse}
Maurer, A.; Pontil, M.; and Romera-Paredes, B. 2013.
\newblock Sparse coding for multitask and transfer learning.
\newblock In \emph{International conference on machine learning}, 343--351.

\bibitem[{McCall(2020)}]{mccall2020covid}
McCall, B. 2020.
\newblock COVID-19 and artificial intelligence: protecting health-care workers
  and curbing the spread.
\newblock \emph{The Lancet Digital Health} 2(4): e166--e167.

\bibitem[{Misra et~al.(2016)Misra, Shrivastava, Gupta, and
  Hebert}]{misra2016cross}
Misra, I.; Shrivastava, A.; Gupta, A.; and Hebert, M. 2016.
\newblock Cross-stitch networks for multi-task learning.
\newblock In \emph{Proceedings of the IEEE Conference on Computer Vision and
  Pattern Recognition}, 3994--4003.

\bibitem[{Novel et~al.(2020)}]{novel2020epidemiological}
Novel, C. P. E. R.~E.; et~al. 2020.
\newblock The epidemiological characteristics of an outbreak of 2019 novel
  coronavirus diseases (COVID-19) in China.
\newblock \emph{Zhonghua liu xing bing xue za zhi= Zhonghua liuxingbingxue
  zazhi} 41(2): 145.

\bibitem[{Pirracchio(2016)}]{pirracchio2016mortality}
Pirracchio, R. 2016.
\newblock Mortality prediction in the icu based on mimic-ii results from the
  super icu learner algorithm (sicula) project.
\newblock In \emph{Secondary Analysis of Electronic Health Records}, 295--313.
  Springer.

\bibitem[{Purushotham et~al.(2017)Purushotham, Meng, Che, and
  Liu}]{purushotham2017benchmark}
Purushotham, S.; Meng, C.; Che, Z.; and Liu, Y. 2017.
\newblock Benchmark of deep learning models on large healthcare mimic datasets.
\newblock \emph{arXiv preprint arXiv:1710.08531} .

\bibitem[{Reichlin et~al.(2009)Reichlin, Hochholzer, Bassetti, Steuer, Stelzig,
  Hartwiger, Biedert, Schaub, Buerge, Potocki et~al.}]{reichlin2009early}
Reichlin, T.; Hochholzer, W.; Bassetti, S.; Steuer, S.; Stelzig, C.; Hartwiger,
  S.; Biedert, S.; Schaub, N.; Buerge, C.; Potocki, M.; et~al. 2009.
\newblock Early diagnosis of myocardial infarction with sensitive cardiac
  troponin assays.
\newblock \emph{New England Journal of Medicine} 361(9): 858--867.

\bibitem[{Richardson et~al.(2020)Richardson, Griffin, Tucker, Smith, Oechsle,
  Phelan, and Stebbing}]{richardson2020baricitinib}
Richardson, P.; Griffin, I.; Tucker, C.; Smith, D.; Oechsle, O.; Phelan, A.;
  and Stebbing, J. 2020.
\newblock Baricitinib as potential treatment for 2019-nCoV acute respiratory
  disease.
\newblock \emph{Lancet (London, England)} 395(10223): e30.

\bibitem[{Ruder et~al.(2017)Ruder, Bingel, Augenstein, and
  S{\o}gaard}]{ruder2017learning}
Ruder, S.; Bingel, J.; Augenstein, I.; and S{\o}gaard, A. 2017.
\newblock Learning what to share between loosely related tasks.
\newblock \emph{ArXiv} .

\bibitem[{Song et~al.(2018)Song, Rajan, Thiagarajan, and
  Spanias}]{song2018attend}
Song, H.; Rajan, D.; Thiagarajan, J.~J.; and Spanias, A. 2018.
\newblock Attend and diagnose: Clinical time series analysis using attention
  models.
\newblock In \emph{Thirty-Second AAAI Conference on Artificial Intelligence}.

\bibitem[{UdeM(2014)}]{mnistvar}
UdeM. 2014.
\newblock Variations on the MNIST Digits.
\newblock
  \urlprefix\url{https://sites.google.com/a/lisa.iro.umontreal.ca/public_static_twiki/variations-on-the-mnist-digits}.

\bibitem[{Vaswani et~al.(2017)Vaswani, Shazeer, Parmar, Uszkoreit, Jones,
  Gomez, Kaiser, and Polosukhin}]{vaswani2017attention}
Vaswani, A.; Shazeer, N.; Parmar, N.; Uszkoreit, J.; Jones, L.; Gomez, A.~N.;
  Kaiser, {\L}.; and Polosukhin, I. 2017.
\newblock Attention is all you need.
\newblock In \emph{Advances in neural information processing systems},
  5998--6008.

\bibitem[{Wang and Wong(2020)}]{wang2020covid}
Wang, L.; and Wong, A. 2020.
\newblock COVID-Net: A tailored deep convolutional neural network design for
  detection of COVID-19 cases from chest radiography images.
\newblock \emph{arXiv} arXiv--2003.

\bibitem[{Woo et~al.(2004)Woo, Lau, Tsoi, Chan, Wong, Che, Tam, Tam, Cheng,
  Hung et~al.}]{woo2004relative}
Woo, P.~C.; Lau, S.~K.; Tsoi, H.-w.; Chan, K.-h.; Wong, B.~H.; Che, X.-y.; Tam,
  V.~K.; Tam, S.~C.; Cheng, V.~C.; Hung, I.~F.; et~al. 2004.
\newblock Relative rates of non-pneumonic SARS coronavirus infection and SARS
  coronavirus pneumonia.
\newblock \emph{The Lancet} 363(9412): 841--845.

\bibitem[{Xie et~al.(2020)Xie, Tong, Guan, Du, Qiu, and
  Slutsky}]{xie2020critical}
Xie, J.; Tong, Z.; Guan, X.; Du, B.; Qiu, H.; and Slutsky, A.~S. 2020.
\newblock Critical care crisis and some recommendations during the COVID-19
  epidemic in China.
\newblock \emph{Intensive care medicine} 1--4.

\bibitem[{Yang and Hospedales(2016{\natexlab{a}})}]{yang2016deep}
Yang, Y.; and Hospedales, T. 2016{\natexlab{a}}.
\newblock Deep multi-task representation learning: A tensor factorisation
  approach.
\newblock \emph{arXiv preprint arXiv:1605.06391} .

\bibitem[{Yang and Hospedales(2016{\natexlab{b}})}]{yang2016trace}
Yang, Y.; and Hospedales, T.~M. 2016{\natexlab{b}}.
\newblock Trace norm regularised deep multi-task learning.
\newblock \emph{arXiv preprint arXiv:1606.04038} .

\end{thebibliography}
\clearpage
\label{appendix}
\section{Detailed Description of Datasets and Experimental Setup}
\subsection{Datasets and Tasks}
Here we include detailed feature information of PhysioNet dataset and explanation on MIMIC III-Respiratory failure and MIMIC III-Heart Failure.

\label{tasks}
\subsubsection{MIMIC III \cite{johnson2016mimic}-Infection}
\begin{figure}[ht!]
	\centering
	\includegraphics[width=0.45\textwidth]{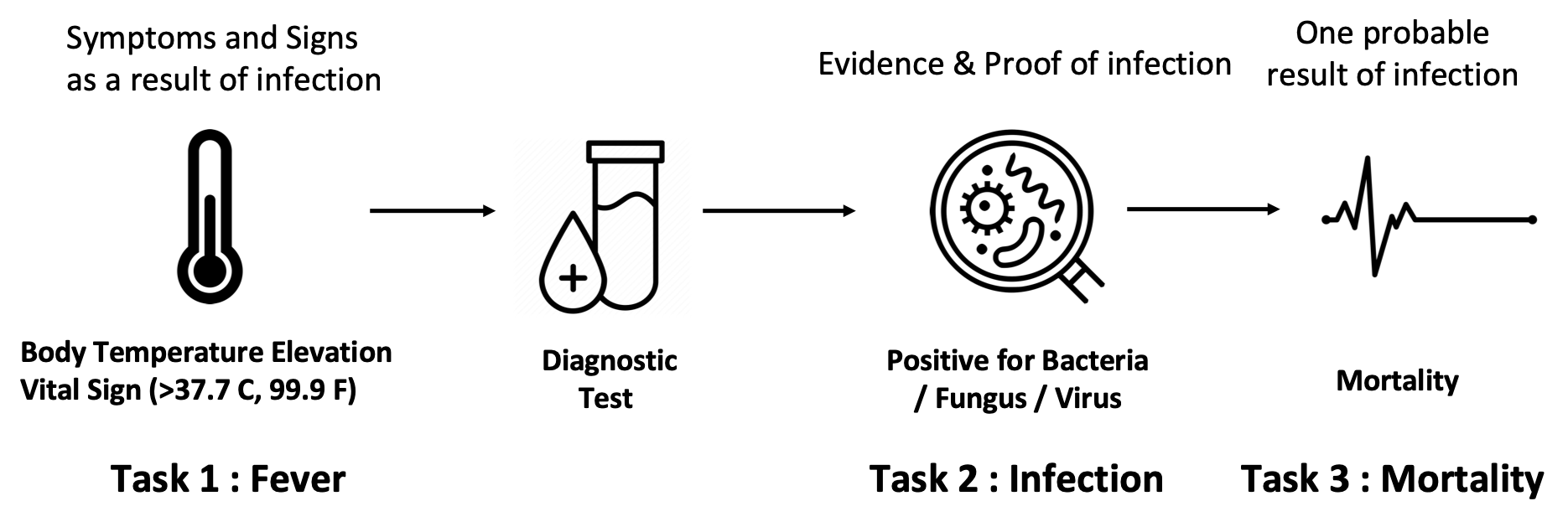}
	\caption{\small \textbf{MIMIC III-Infection: Task overview.} Tasks used in the experiment included the diagnostic process of the patient infectious status. When a patient in ICU is infected with any of the pathogens, body temperature elevates (\textit{Fever (Task 1)}) as a sign of infection. Next, \textit{Infection (Task 2)}, is confirmed when the blood culture result turns out to be positive with bacteria, fungus, or virus. Lastly, \textit{Mortality (Task 3)} can be resulted from infection.}
	\label{infection_task}
\end{figure}

Here we include the configuration of MIMIC III-Infection dataset for reference (\textcolor{navyblue}{Figure} \ref{infection_task}).

\subsubsection{PhysioNet} 
\begin{figure}[ht!]
	\centering
	\includegraphics[width=0.4\textwidth]{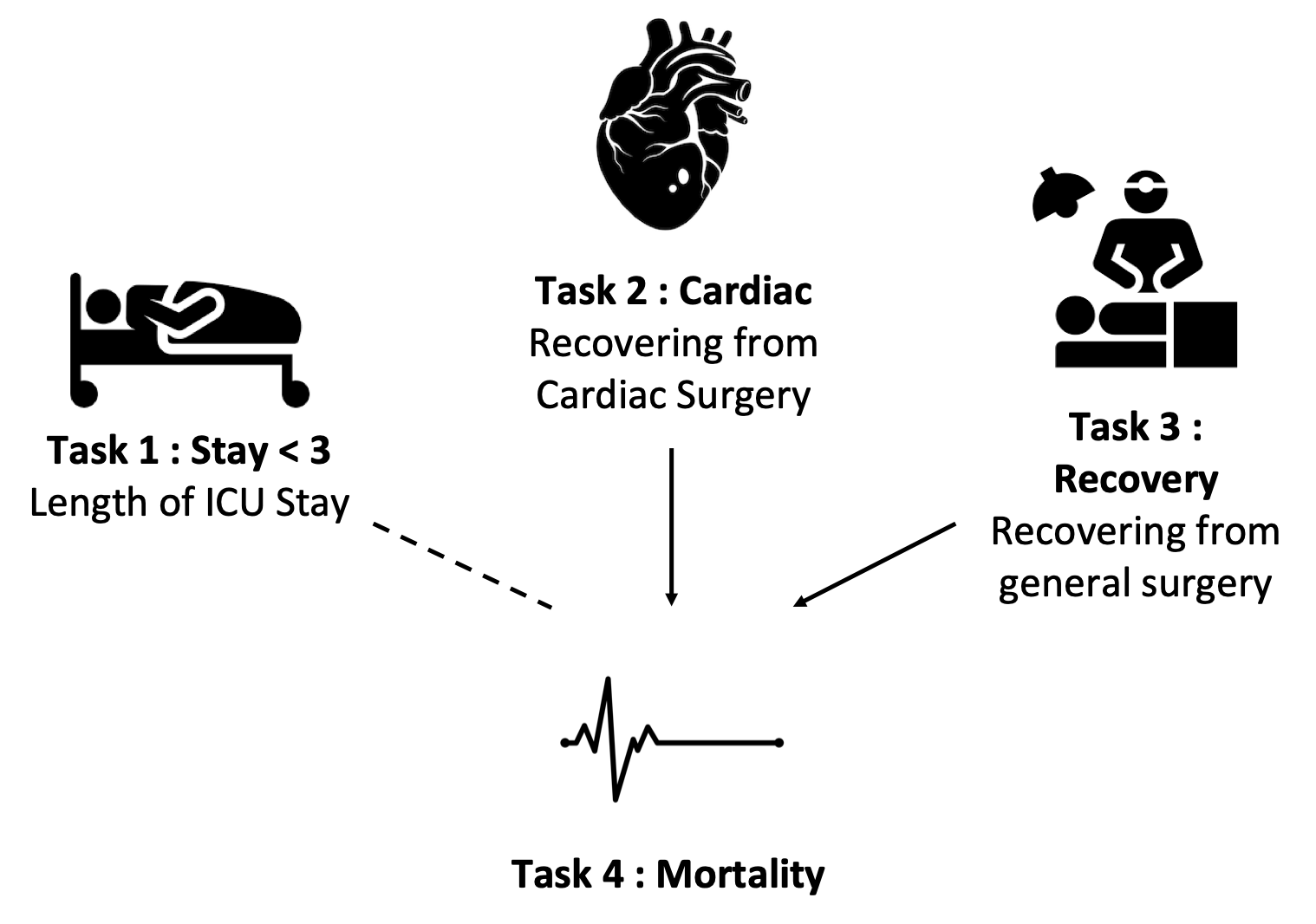}
	\caption{\small \textbf{PhysioNet : Task overview.} Tasks generated from PhysioNet dataset including three tasks that have temporal relationship with task \textit{Mortality (Task 3)}. Shorter length of stay in ICU (\textit{Stay $<3$ (Task 1)}) has reverse temporal relationship with \textit{Mortality}. Also, patient condition related to surgical operation(\textit{Cardiac (Task 2)} surgery, and general surgery (\textit{Recovery (Task 3)})) are temporally related to \textit{Mortality}.}
	\label{physionet_task}
\end{figure}

\cite{citi2012physionet} (\textcolor{navyblue}{Figure}~\ref{physionet_task})
We list 29 pre-selected physiological signs in this section: \textit{age, gender, height, weight, Systolic Blood Pressure, Diastolic Blood Pressure, mean arterial pressure, heart rate, respiratory rate, body temperature, glucose, bilirubin, serum electrolytes (sodium, potassium, magnesium, bicarbonate), lactate, pH, Hematocrit, platelets, Partial Pressure of Oxygen (PaO$_2$), Partial Pressure of carbon dioxide (PaCO$_2$), Oxygen Saturation (SaO$_2$), Fraction of Inspired Oxygen (FiO$_2$), Glasgow Coma Scale (GCS), blood urea nitrogen (BUN), Creatinine, Urine, Mechanical Ventilation Status.}

To test the generalization performance of our model, we compiled two additional datasets, namely \textbf{Heart Failure} and \textbf{Respiratory Failure}, out of the MIMIC III dataset~\cite{johnson2016mimic}.
 
\subsubsection{MIMIC III-Heart Failure} From the MIMIC III dataset, we collected $3,577$ data instances, each of which corresponds to the record of a patient (age between 18 and 100) admitted to the ICU of a hospital. This dataset contains 15 features which are associated to the risk of heart failure occurrence, including \textit{Heart rate (HR)}, \textit{Systolic Blood Pressure (SBP)}, \textit{Diastolic Blood Pressure (DBP)}, \textit{Body Temperature (BT)}, \textit{Fraction of inspired oxygen (FiO$_2$)}, \textit{Mixed venous oxygen saturation (M$_v$O$_2$)}, \textit{Oxygen Saturation of arterial blood (S$_a$O$_2$)}, \textit{Brain natriuretic peptide (BNP)}, \textit{Ejection Fraction (EF)}, \textit{Glasgow Coma scale (GCS) - Verbal, Motor, Eye}. We considered four tasks that might lead to heart failure. The first task (\textit{Ischemic}) is the patient condition where a patient is diagnosed with ischemic heart disease. The second task (\textit{Valvular}) is related to the diagnosis of valvular heart disease, and the third task (\textit{Heart Failure}) contains the condition where a patient is diagnosed with various types of heart failure. Lastly, \textit{Mortality} (Task 4) can be a possible outcome of heart failure. We used random split of approximately 1850/925/925 instances for train, valid, test set. ICD code for the diagnosis of each disease is summarized in \textcolor{navyblue}{Table}~\ref{hf_label} and \textcolor{navyblue}{Table}~\ref{hf_feature} summarizes all feature information used to construct this dataset.

\begin{figure}[ht!]
	\centering
	\includegraphics[width=0.45\textwidth]{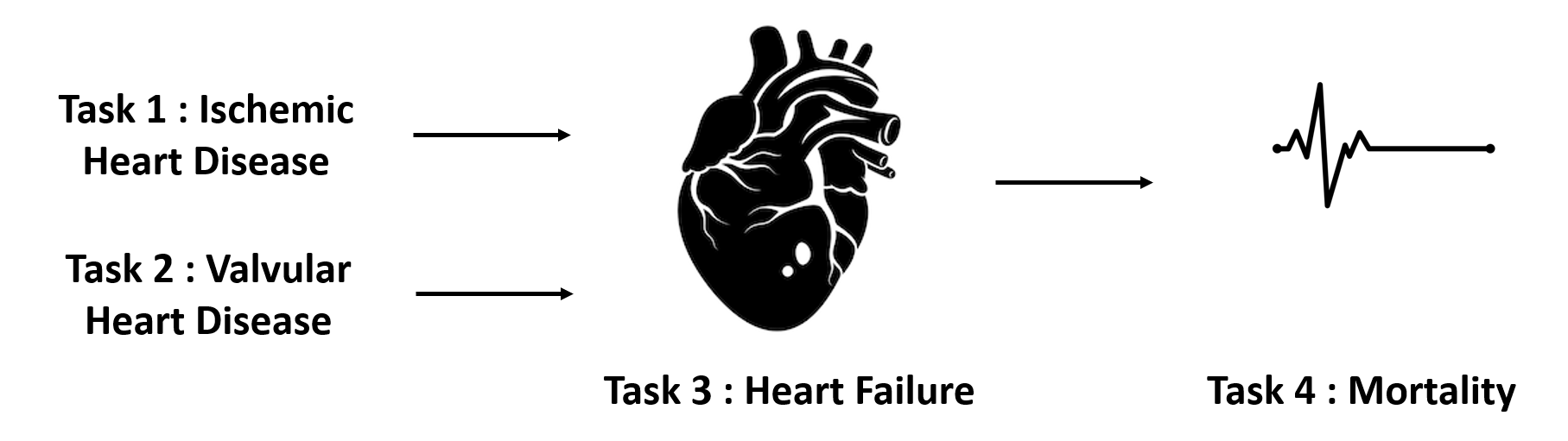}
	\caption{\small \textbf{MIMIC III-Heart Failure: Task overview.} Tasks used in the experiment include the tasks that are temporally related. The first and second tasks \textit{Ischemic (Ischemic Heart Disease, Task 1)} and \textit{Valvular (Valvular Heart Disease, Task 2)} can both result in \textit{Heart Failure (Task 3)} of a patient. Also, \textit{Mortality (Task4)} is one possible outcome of \textit{Heart Failure}.}
	\label{hf_task}
\end{figure}

\subsubsection{MIMIC III-Respiratory Failure} We collected total of $37,818$ distinct ICU admissions of adult patients, between the age of 18 and 100. To further test the generalization performance of our model to larger dataset, we test the model performance on three partial datasets. First we run the experiment with full admission instances sampled for 48 hours after admission (\textcolor{navyblue}{Table}~\ref{respiratoryfailure_original}), full $37,818$ instances of data sampled for 48 hours (\textcolor{navyblue}{Table}~\ref{respiratoryfailure_30000}). The disease and clinical event label, ICD code and item ID are summarized in \textcolor{navyblue} {Table} ~\ref{rf_label}. This dataset contains 29 features, which are predictive of respiratory failure occurrence, including \textit{Fraction of inspired oxygen (Fi$O_2$)}, \textit{Oxygen Saturation of arterial blood ($S_a$O$_2$) and venous blood ($S_v O_2$)},  Partial Pressure of oxygen in alveoli ($P_A O_2$), Arterial/Venous oxygen content ($C_a O_2$), which is summarized in \textcolor{navyblue} {Table}~\ref{rf_feature}. Tasks considered in this task includes four respiratory conditions that are required for the prediction of respiratory failure (Task 5) and mortality (Task 7) (\textcolor{navyblue}{Table}~\ref{rf_label}): Hypoxemia (Task1, $P_aO_2 < 60mmHg$), Hypercapnia (Task2, $P_aCO_2 > 50mmHg$), VQ Mismatch (Task 3, $A_aDO_2 > 10$), Acidosis (Task 4, $pH < 7.35$), Respiratory Failure (Task 5), Cyanosis (Task 6, $S_aO_2 < 60mmHg$). The collected dataset was randomly split into $26,472$/$5,672$/$5,674$ for training/validation/test in full dataset, and $2,800$/$600$/$600$ for total $4,000$ instances.

\begin{figure}[ht!]
	\centering
	\includegraphics[width=0.45\textwidth]{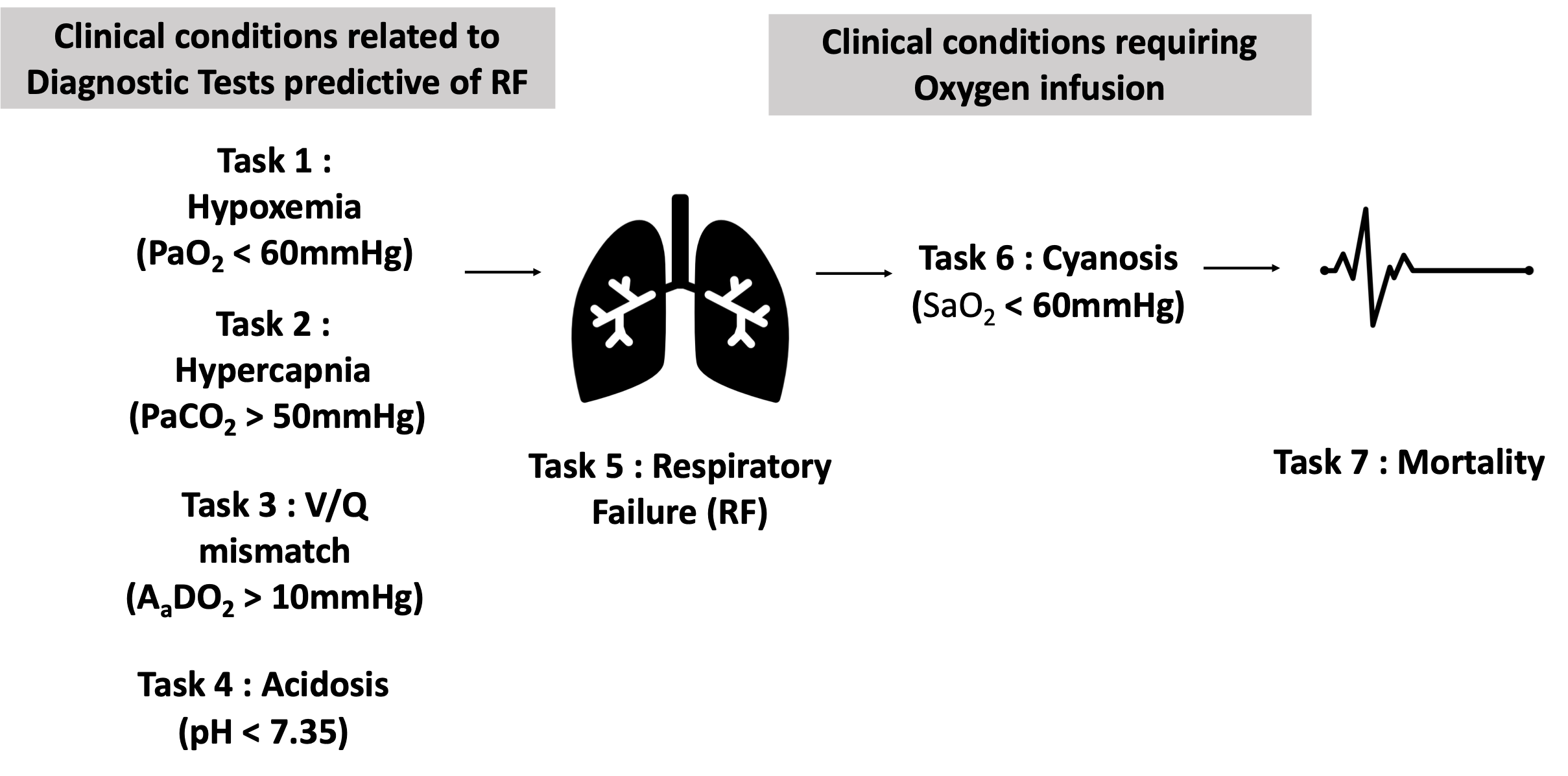}
	\caption{\small \textbf{MIMIC III-Respiratory Failure: Task overview.} Tasks used in the experiment includes the tasks that are 1) clinical conditions related to diagnostic tests of respiratory failure (\textit{Hypoxemia (Task 1)}, \textit{Hypercarbia (Task 2)}, \textit{V/Q mismatch (Task 3)}, and \textit{Acidosis (Task 4)}), 2) \textit{Respiratory Failure (Task 5)}, 3) clinical condition related to treatment planning of patients (\textit{Cyanosis (Task 6)}), 4) \textit{Mortality (Task 7)}.}
	\label{rf_task}
\end{figure}

\subsection{Baselines}
\label{baselines}
\subsubsection{STL-LSTM.} The single- and multi-task learning method which uses RNNs to capture the temporal dependencies.

\subsubsection{MTL-LSTM.} The naive hard-sharing multi-task learning method where all tasks share the same network except for the separate output layers for prediction, whose base network is Long Short-Term Memory Network (LSTM).  

\begin{gather*}
  (\mathbf{v}_d^{(1)},\mathbf{v}_d^{(2)},...,\mathbf{v}_d^{(T)}) = \mathbf{v}_d = \mathbf{x} \mathbf{W}^d_{emb} \in \mathbbm{R}^{T\times k} \\
  (\mathbf{h}_d^{(1)},\mathbf{h}_d^{(2)},...,\mathbf{h}_d^{(T)}) = LSTM_d(\mathbf{v}_d^{(1)},\mathbf{v}_d^{(2)},...,\mathbf{v}_d^{(T)})\\
  \beta_d^{(t)} = \tanh{(\bm{h}_d^{(t)})}\\
  p(\widehat{y_d}|\mathbf{x})=Sigmoid\left(\frac{1}{T}\left(\mathlarger{\sum}_{t=1}^T\beta_d^{(t)}\odot \mathbf{v}^{(t)}\right)\mathbf{W}_d^o+b_d^o\right)
\end{gather*}

\begin{figure}[ht!]
	\begin{subfigure}{.45\textwidth}
		\centering
		\includegraphics[width=\linewidth]{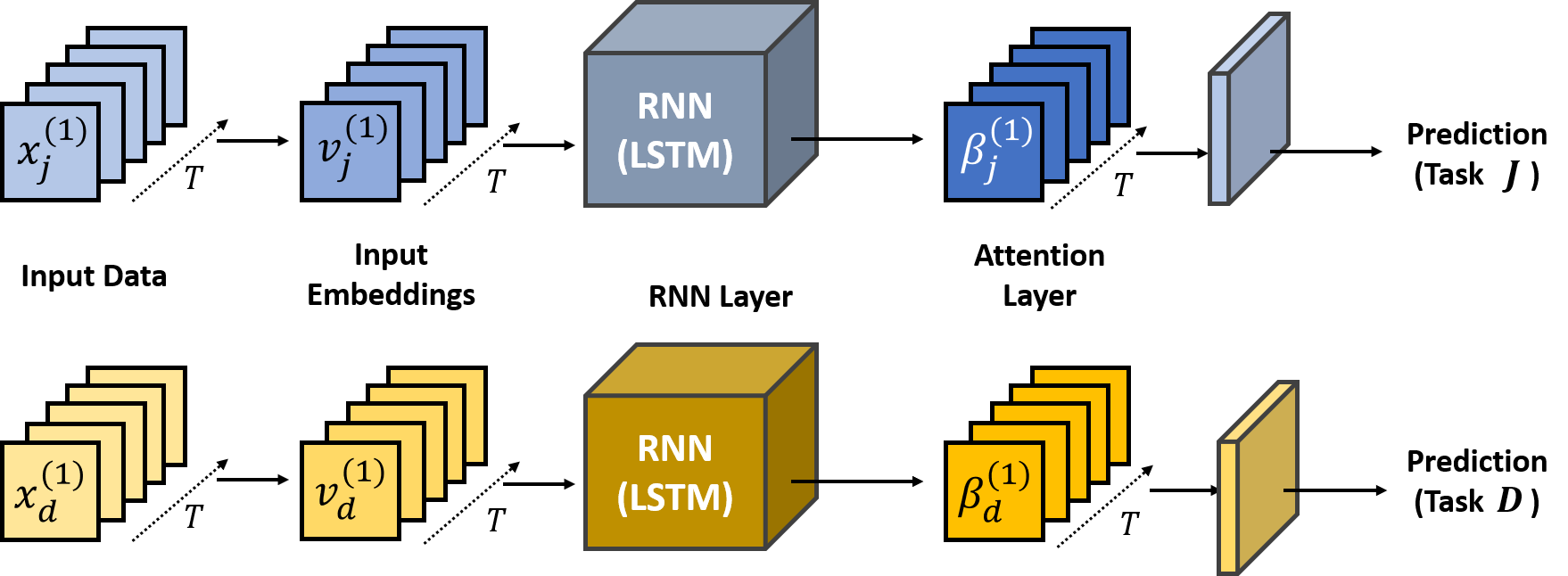}
		\caption{Single Task Learning with LSTM}
		\label{stl-lstm}
	\end{subfigure}
	\begin{subfigure}{.4\textwidth}
		\centering
		\includegraphics[width=\linewidth]{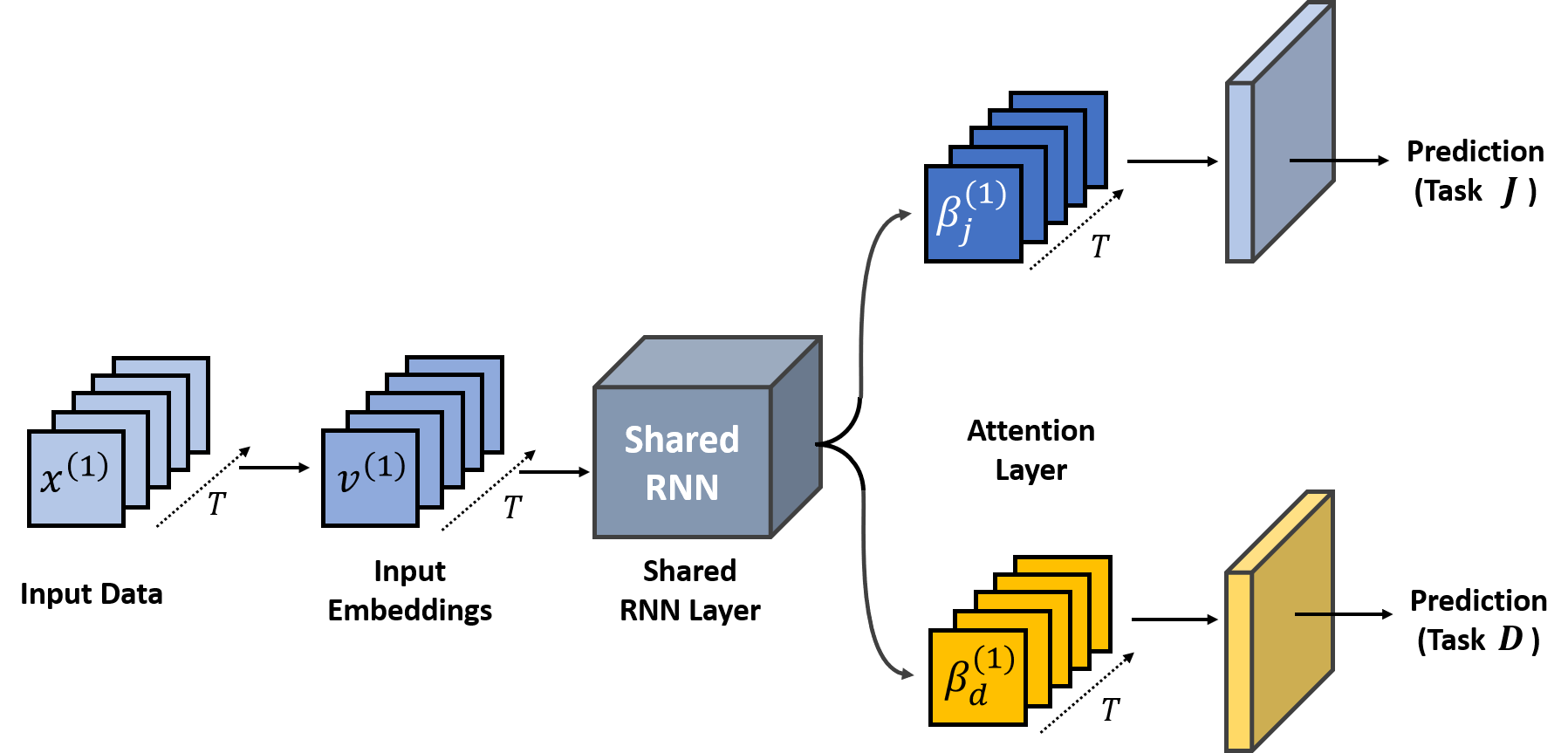}
		\caption{Multi-Task Learning with LSTM}
		\label{mtl-lstm}
	\end{subfigure}
    \caption{(a) Single- and (b) Multi-Task Learning with LSTM}
\end{figure}

\subsubsection{STL/MTL-Transformer} The single-task learning method which uses Transformer architecture \cite{vaswani2017attention} to capture the temporal dependencies:

\begin{gather*}
\mathbf{v} = \mathbf{x} \mathbf{W}_{emb} + POS\_ENC \in \mathbbm{R}^{T \times k} \\
\mathbf{f}_d = TRANS\_BLOCK_d(v) \in \mathbbm{R}^{T\times k} \\
\mathbf{c}_d = \frac{1}{T}(\mathbf{f}_d^{(1)}+\mathbf{f}_d^{(2)}+...+\mathbf{f}_d^{(T)}) \\
p(\widehat{y_d}|\mathbf{x})=Sigmoid\left(\mathbf{c}_i\mathbf{W}_d^o+b_d^o\right)
\end{gather*}

where $POS\_ENC$ is the positional encoding used in Transformer, $TRANS\_BLOCK$ is also the architecture used in the paper, which consists of two sublayers: $MULTI\_HEAD$ (with four heads) and $FFW$. We also used residual connection and layer norm after each sublayer as in the original paper.

\begin{figure}[ht!]
	\begin{center}
		\centerline{\includegraphics[width=.45\textwidth]{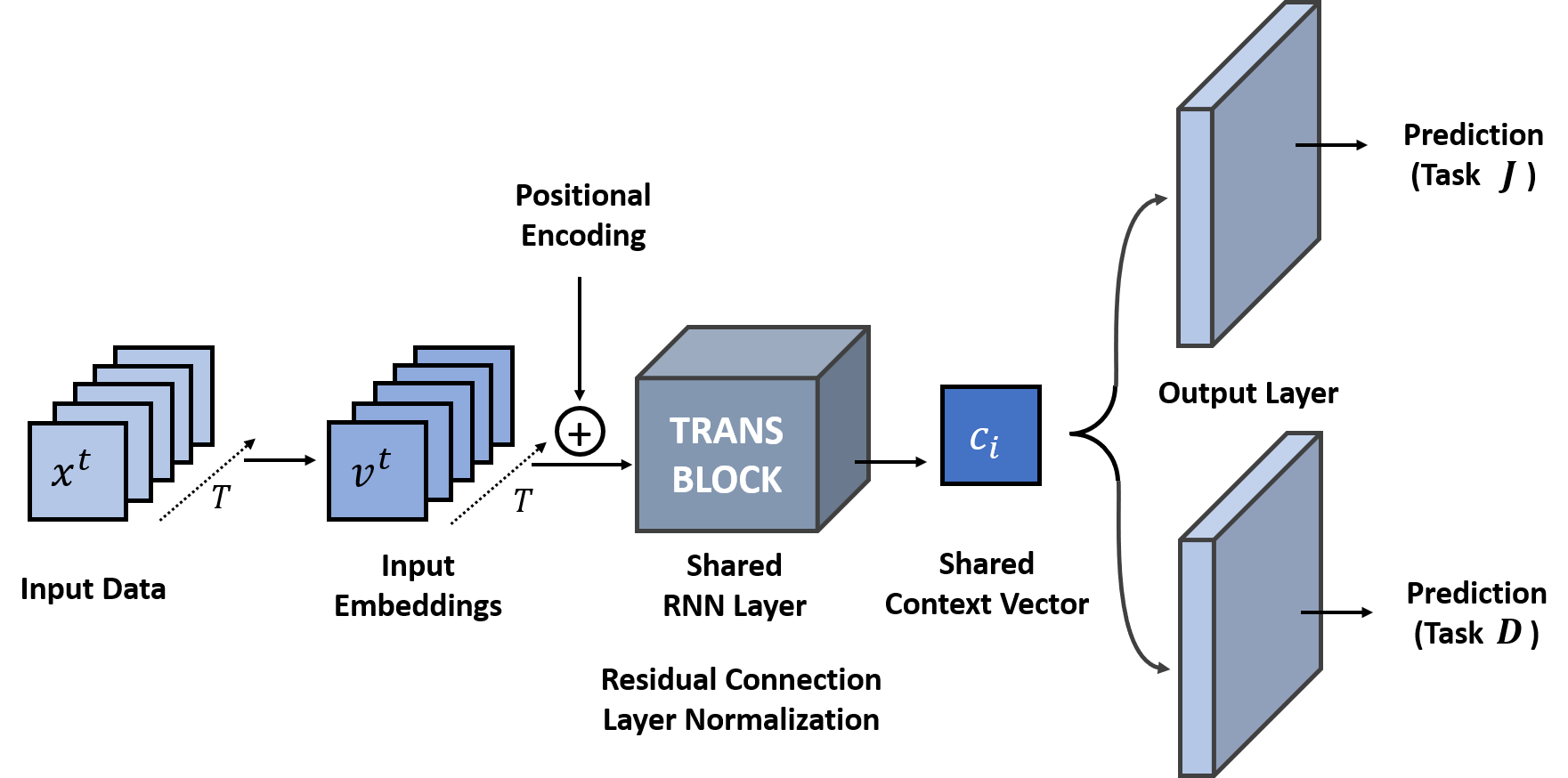}}
		\caption{Multi Task Learning with Transformer}
		\label{mtl-trans}
	\end{center}
\end{figure}

\subsubsection{MTL-RETAIN.} The same as MTL-LSTM, but with RETAIN\cite{choi2016retain} as the base network. Specifically, after getting the shared context vector $\mathbf{c}_i$, separated output layers will be applied to form the prediction for each task.

\subsubsection{MTL-UA.} The same as MTL-LSTM, but with UA\cite{heo2018uncertainty} as the base network. Specifically, after getting the shared context vector $\mathbf{c}_i$, separated output layers will be applied to form the prediction for each task. This can be seen as the probabilistic version of MTL-RETAIN.

\begin{align}
\mathbf{c}_i : \text{context vector from RETAIN, UA} \\ 
p(\widehat{y_d}|\mathbf{x})=Sigmoid\left(\mathbf{c}_i\mathbf{W}_d^o+b_d^o\right)
\end{align}

\begin{figure}[ht!]
	\begin{center}
		\centerline{\includegraphics[width=.45\textwidth]{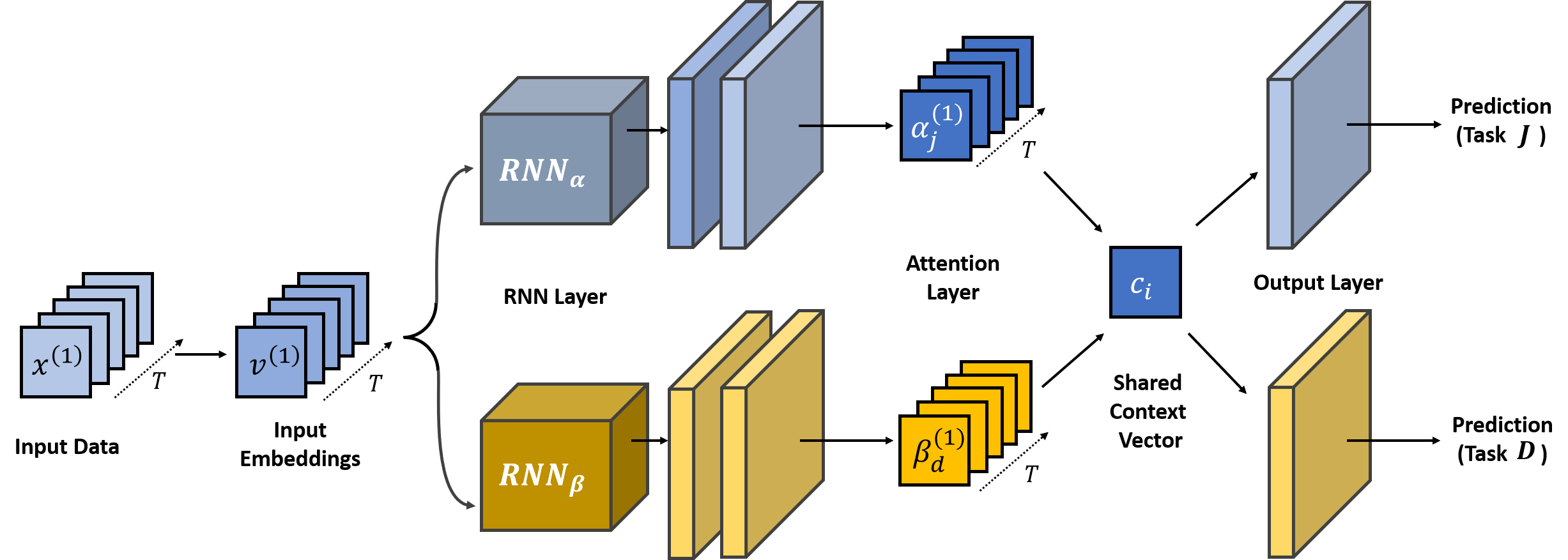}}
		\caption{Multi Task Learning with RETAIN, UA model}
		\label{mtl-ua}
	\end{center}
\end{figure}

\subsubsection{AMTL-LSTM.}  This is asymmetric multi-task learning~\cite{lee2016asymmetric} adopted for our time-series prediction framework, where we learn the knowledge transfer graph between task-specific parameters, which is learned to perform asymmetric knowledge transfer based on the task loss. The parameters for each task are shared across all timesteps, which will result in static asymmetric transfer between tasks. 

\subsubsection{MTL-RETAIN-Kendall}  This model is similar to MTL-RETAIN.
However, we followed the uncertainty-aware loss weighting scheme from~\cite{kendall2018multi} to weight the loss for each task by its uncertainty:

\begin{align}
\mathlarger{\sum}_{d=1}^D {\left(\frac{1}{\sigma_d^2}L_d+log(\sigma_d)\right)}
\end{align}

\subsubsection{TP-AMTL.}  Our probabilistic temporal asymmetric multi-task learning model that performs both intra-task and inter-task knowledge transfer.


\subsection{Details of models in Ablation study}
\subsubsection{AMTL-intratask.}  The probabilistic AMTL model with uncertainty-aware knowledge transfer, but performs knowledge transfer only within the same task at the transfer layer. Note that, however, this model can still share inter-task knowledge in a symmetrical manner since it still has shared lower layers (the embedding and the LSTM layers).

\begin{align}
\mathbf{C}_d^{(t)}=\mathbf{f}_d^{(t)}+\mathlarger{\sum}_{i=1}^t \alpha^{(i,t)}_{d,d}*G_d\left(\mathbf{f}_d^{(i)}\right) \forall t \in \{1,2,...,T\}
\end{align}

\subsubsection{AMTL-samestep.}  The probabilistic model with uncertainty-aware knowledge transfer, which performs knowledge transfer only between the features at the same timestep, at the transfer layer. Again, note that this model can still capture the temporal dependencies among the timesteps to certain degree, as it has shared lower layers.

\begin{align}
\mathbf{C}_d^{(t)}=\mathbf{f}_d^{(t)}+\mathlarger{\sum}^D_{j=1} G_{j,d}\left(\mathbf{f}_j^{(t)}\right) \forall t \in \{1,2,...,T\}
\end{align}

\begin{figure}[ht!]
	\begin{subfigure}{.45\textwidth}
		\centering
		\includegraphics[width=\linewidth]{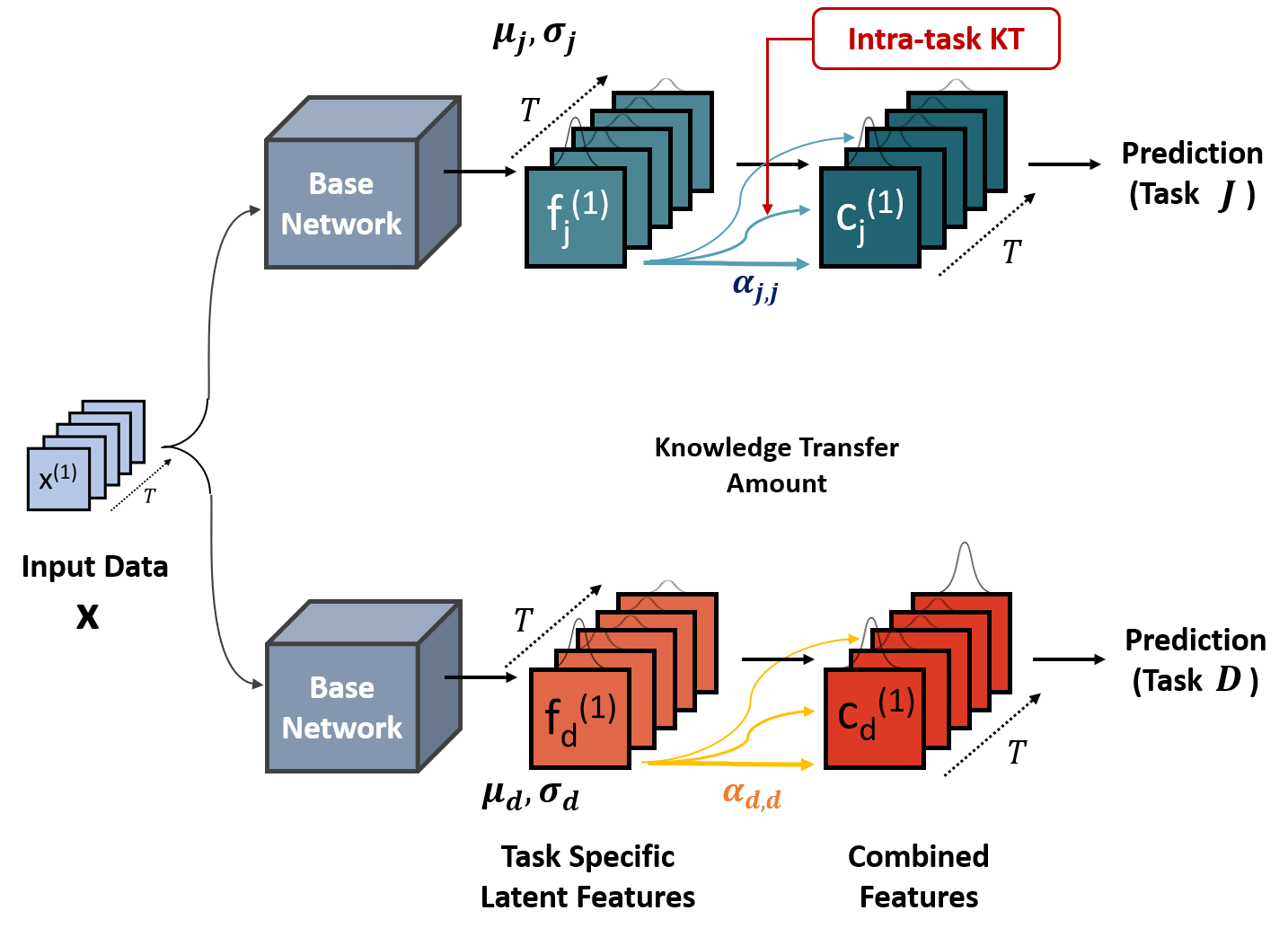}
		\caption{Intratask Knowledge Transfer}
		\label{amtl-intratask}
	\end{subfigure}
	\begin{subfigure}{.45\textwidth}
		\centering
		\includegraphics[width=\linewidth]{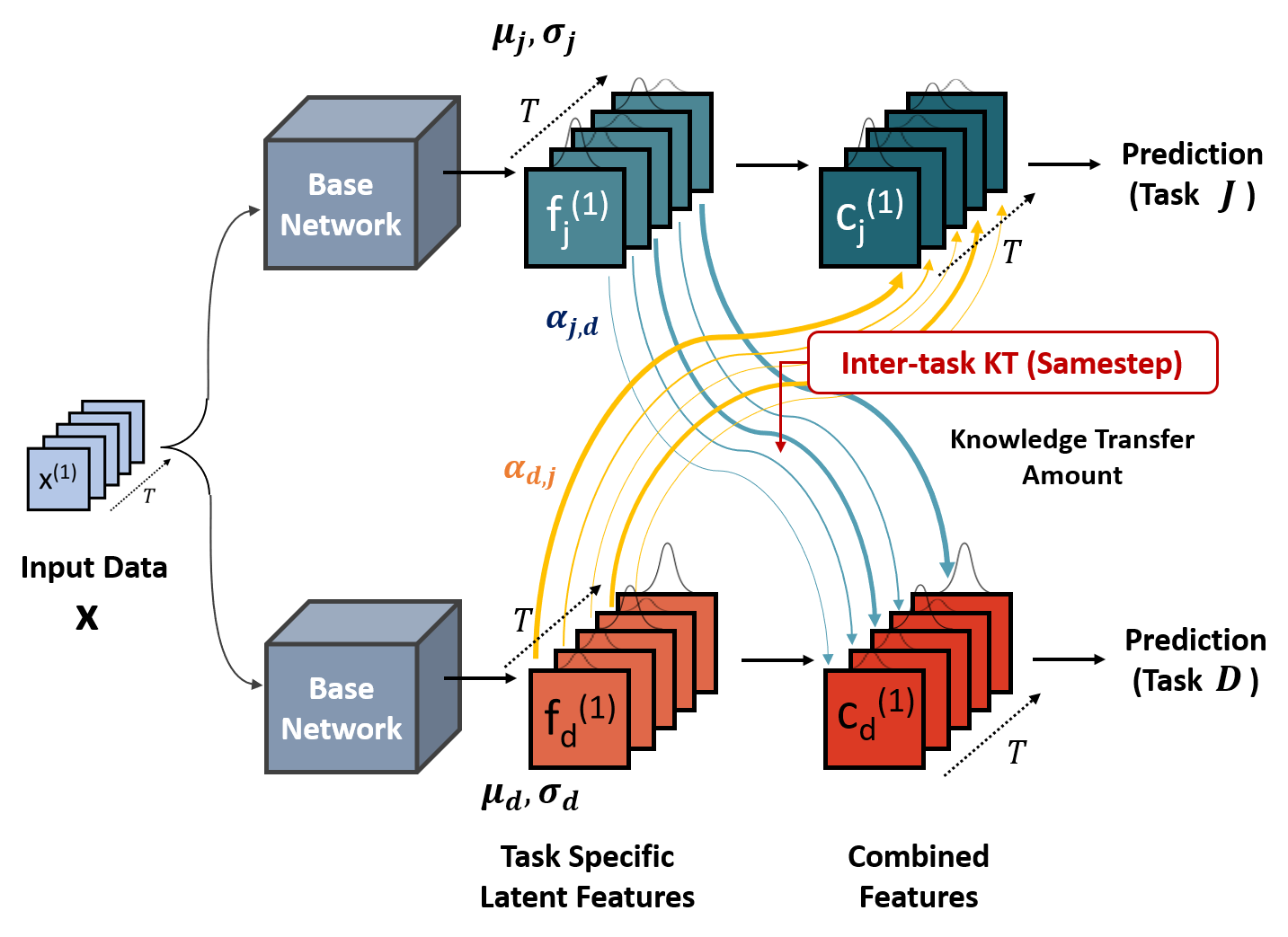}
		\caption{Samestep Knowledge Transfer}
		\label{amtl-samestep}
	\end{subfigure}
	\caption{Asymmetric Multi Task Learning with \textbf{(a) Intratask Knowledge Transfer} and \textbf{ (b) Samestep Knowledge Transfer}}
\end{figure}

\subsubsection{TD-TAMTL.}  The deterministic version of our model that does not make use of feature-level uncertainty when performing knowledge transfer.\\
\begin{align*}
  &\mathbf{v} = \mathbf{x} \mathbf{W}_{emb} \in \mathbbm{R}^{T\times k} \\
  &\mathbf{h} = (\mathbf{h}^{(1)},\mathbf{h}^{(2)},...,\mathbf{h}^{(T)}) = RNN(\mathbf{v}^{(1)},\mathbf{v}^{(2)},...,\mathbf{v}^{(T)})\\
  &\mathbf{h}_d = \sigma((...\sigma(\sigma(\mathbf{h} \mathbf{W}^1_d+\mathbf{b}^1_d)\mathbf{W}^2_d+\mathbf{b}^2_d)...)\mathbf{W}^L_d+\mathbf{b}^L_d) \\
  &(\mathbf{f}_d^{(1)},\mathbf{f}_d^{(2)},...,\mathbf{f}_d^{(T)}) = \mathbf{h}_d \in \mathbbm{R}^{T\times k}\\
  &\mathbf{C}_d^{(t)}=\mathbf{f}_d^{(t)}+\mathlarger{\sum}^D_{j=1} \mathlarger{\sum}_{i=1}^t \alpha^{(i,t)}_{j,d}*G\left(\mathbf{f}_j^{(i)}\right) \\
  &\forall t \in \{1,2,...,T\}\\
  &\bm{\beta}_d^{(t)}=tanh\left( \mathbf{C}_d^{(t)}\mathbf{W}_d^\beta+\mathbf{b}_d^\beta\right) \quad\forall t \in \{1,2,...,T\}\\
\end{align*}
\begin{align*}
  &p(\widehat{y_d}|\mathbf{x})=Sigmoid\left(\frac{1}{T}\left(\mathlarger{\sum}_{t=1}^T\beta_d^{(t)}\odot \mathbf{v}^{(t)}\right)\mathbf{W}_d^o+b_d^o\right)
\end{align*}

where $\alpha^{(i,t)}_{j,d}=F_\theta(\mathbf{f}_f^{(i)},\mathbf{f}_d^{(t)})$

\subsubsection{TP-TAMTL - no constraints.}
\begin{gather*}
    \mathbf{C}_d^{(t)}=\mathbf{f}_d^{(t)}+\mathlarger{\sum}^D_{j=1} \mathlarger{\sum}_{i=1}^T \alpha^{(i,t)}_{j,d}*G_{j,d}\left(\mathbf{f}_j^{(i)}\right) \\
    \forall t \in \{1,2,...,T\}
\end{gather*}

\subsection{Configuration and Hyperparameters}
\label{config, hyperparam}
We trained all the models using Adam optimizer. We set the maximum iteration for Adam optimizer as 100,000, and for other hyper-parameters, we searched for the optimal values by cross-validation, within predefined ranges as follows: Hidden units: \{8, 16, 32, 64\}, number of layers: \{2,3,6\}, mini batch size: \{32, 64, 128, 256\}, learning rate: \{0.01, 0.001, 0.0001\}, \textit{L2} regularization: \{0.02, 0.002, 0.0002\}, and dropout rate: \{0.1, 0.15, 0.2, 0.25, 0.3, 0.4, 0.5\}.

\subsection{Time Complexity of the model}
The complexity of each baselines and our final model is summarized in \textcolor{navyblue}{Table} \ref{Complexity}.

\begin{table}[ht!]
	\footnotesize
	\caption{Complexity of the Baseline Models}
 	\label{Complexity}
	\centering
	\resizebox{0.4\textwidth}{!}{
		\begin{tabular}{c|c}
			\toprule
			Models & Complexity \\
			\midrule
			Transformer \citep{vaswani2017attention} & $O(T^2)$ \\
			RETAIN \citep{choi2016retain} & $O(T)$ \\
			UA \citep{heo2018uncertainty}  & $O(T)$ \\
			SAnD \citep{song2018attend} & $O(T^2)$ \\
    		\midrule
		    \textbf{TP-AMTL (ours)} & $O(T)$ \\
			\bottomrule
		\end{tabular}
}
\end{table}

\begin{table*}[ht!]
	\small
	\caption{\textbf{MIMIC-III Heart Failure.} The reported numbers are average AUROC and standard error over \textbf{five runs}. The numbers colored in \textcolor{red}{red} for MTL models denote accuracies lower than those of their STL counterparts.}
	\label{heartfailure}
	\centering
	\resizebox{0.7\textwidth}{!}{
		\begin{tabular}{ccccccc}
			\toprule
			\multicolumn{2}{c}{Models} & \multicolumn{4}{c}{Tasks}       &             \\
			\cmidrule(r){3-6}
			&  & Ischemic & Valvular & Heart Failure & Mortality  & Average\\
			\cmidrule{1-7}
			\multicolumn{1}{c|}{} 
			& LSTM   
			& 0.7072 $\pm$ 0.01 
			& 0.7700 $\pm$ 0.02  
			& 0.6899 $\pm$ 0.02 
			& 0.7169 $\pm$ 0.03 
			& 0.7210 $\pm$ 0.01
			\\
			\multicolumn{1}{c|}{} 
			& Transformer\cite{vaswani2017attention}  
			& 0.6877 $\pm$ 0.01 
			& 0.7655 $\pm$ 0.01 
			& 0.7062 $\pm$ 0.01
			& 0.7183 $\pm$ 0.03 
			& 0.7194 $\pm$ 0.01 
			\\
			\multicolumn{1}{c|}{STL}
			& RETAIN \cite{choi2016retain}   
			& 0.6573 $\pm$ 0.03 
			& 0.7875 $\pm$ 0.01  
			& 0.6850 $\pm$ 0.01
			& 0.7027 $\pm$ 0.02 
			& 0.7081 $\pm$ 0.01
			\\
			\multicolumn{1}{c|}{} 
			& UA \cite{heo2018uncertainty}
			& 0.6843 $\pm$ 0.01 
			& 0.7728 $\pm$ 0.02 
			& \textbf{0.7090 $\pm$ 0.01} 
			& \textbf{0.7191 $\pm$ 0.01} 
			& 0.7213 $\pm$ 0.01
			\\
			\multicolumn{1}{c|}{} 
			& SAnD\cite{song2018attend}  
			& 0.7083 $\pm$ 0.01 
			& 0.7995 $\pm$ 0.01 
			& 0.7245 $\pm$ 0.01
			& 0.7438 $\pm$ 0.01
			& 0.7440 $\pm$ 0.01
			\\
			\multicolumn{1}{c|}{} 
			& AdaCare\cite{ma2020adacare}  
			& 0.6274 $\pm$ 0.02 
			& 0.7509 $\pm$ 0.01 
			& 0.6496 $\pm$ 0.02
			& 0.6775 $\pm$ 0.01 
			& 0.6764 $\pm$ 0.01 
			\\
			\cmidrule{1-7}
			\multicolumn{1}{c|}{} 
			& LSTM    
			& \textcolor{red}{0.6838 $\pm$ 0.02} 
			& 0.7808 $\pm$ 0.02  
			& 0.6965 $\pm$ 0.01 
			& \textcolor{red}{0.7093 $\pm$ 0.02} 
			& 0.7254 $\pm$ 0.02
			\\
			\multicolumn{1}{c|}{} 
			& TRANS\cite{vaswani2017attention}  
			& 0.6801 $\pm$ 0.01 
			& 0.7693 $\pm$ 0.01 
			& \textbf{0.7098 $\pm$ 0.02} 
			& 0.7008 $\pm$ 0.02 
			& 0.7150 $\pm$ 0.02 
			\\
			\multicolumn{1}{c|}{MTL} 
			& RETAIN  
			& 0.6649 $\pm$ 0.01 
			& \textcolor{red}{0.7532 $\pm$ 0.03}  
			& 0.6868 $\pm$ 0.02 
			& 0.7023 $\pm$ 0.03 
			& 0.7018 $\pm$ 0.02
			\\
			\multicolumn{1}{c|}{} 
			& UA 
			& 0.6917 $\pm$ 0.01 
			& 0.7868 $\pm$ 0.01 
			& \textcolor{red}{0.7073 $\pm$ 0.01} 
			& \textcolor{red}{0.7029 $\pm$ 0.01} 
			& 0.7222 $\pm$ 0.01 
			\\
			\multicolumn{1}{c|}{} 
			& SAnD\cite{song2018attend}  
			& 0.7206 $\pm$ 0.01
			& 0.8036 $\pm$ 0.01
			& 0.7169 $\pm$ 0.01
			& 0.7369 $\pm$ 0.01
			& 0.7445 $\pm$ 0.01
			\\
			\multicolumn{1}{c|}{} 
			& AdaCare\cite{ma2020adacare}  
			& 0.6261 $\pm$ 0.03
			& 0.6937 $\pm$ 0.01 
			& 0.6466 $\pm$ 0.02
			& 0.6059 $\pm$ 0.02 
			& 0.6431 $\pm$ 0.02
			\\
			\multicolumn{1}{c|}{} 
			& RETAIN-Kendall 
			& 0.6476 $\pm$ 0.03 
			& 0.7712 $\pm$ 0.02 
			& 0.6826 $\pm$ 0.01 
			& 0.7017 $\pm$ 0.02 
			&  0.7008 $\pm$ 0.01 
			\\
			\multicolumn{1}{c|}{} 
			& AMTL-LSTM\cite{lee2016asymmetric} 
			& 0.6963 $\pm$ 0.01 
			& \textbf{0.7997 $\pm$ 0.02} 
			& 0.7006 $\pm$ 0.01 
			& 0.7108 $\pm$ 0.01 
			& 0.7268 $\pm$ 0.01 
			\\
			\cmidrule{1-7}
			& TP-AMTL (our model) 
			& \textbf{0.7113 $\pm$ 0.01} 
			& \textbf{0.7979 $\pm$ 0.01} 
			& \textbf{0.7103 $\pm$ 0.01} 
			& \textbf{0.7185 $\pm$ 0.02} 
			& \textbf{0.7345 $\pm$ 0.01}
			\\
			\bottomrule
		\end{tabular}
	}
\end{table*}
\begin{table*}[ht!]
	\small
	\caption{\textbf{MIMIC-III Respiratory Failure (RF)}. The reported numbers are average AUROC and standard error over \textbf{five runs}. The numbers colored in \textcolor{red}{red} for MTL models denote accuracies lower than those of their STL counterparts.}
	\label{respiratoryfailure_original}
	\centering
	\resizebox{\textwidth}{!}{
		\begin{tabular}{cccccccccc}
			\toprule
			\multicolumn{2}{c}{Models} & \multicolumn{7}{c}{Tasks}       &             \\
			\cmidrule(r){3-9}
			&  & Hypoxemia & Hypercapnia & VQ Mismatch & Acidosis & RF & Cyanosis & Mortality & Average\\
			\cmidrule{1-10}
			\multicolumn{1}{c|}{} 
			& LSTM   
			& 0.7679 $\pm$ 0.01 
			& 0.7278 $\pm$ 0.02 
			& 0.8032 $\pm$ 0.02 
			& 0.8916 $\pm$ 0.01 
			& 0.7127 $\pm$ 0.02 
			& 0.6447 $\pm$ 0.02 
			& 0.8038 $\pm$ 0.01 
			& 0.7645 $\pm$ 0.03
			\\
			\multicolumn{1}{c|}{} 
			& Transformer \cite{vaswani2017attention} 
			& 0.7573 $\pm$ 0.02
			& 0.7326 $\pm$ 0.01
			& 0.8406 $\pm$ 0.02
			& 0.8919 $\pm$ 0.00 
			& 0.7750 $\pm$ 0.05
			& 0.8001 $\pm$ 0.01
			& 0.7069 $\pm$ 0.03
			& 0.7863 $\pm$ 0.01
			\\
			\multicolumn{1}{c|}{STL} 
			& RETAIN \cite{choi2016retain}   
			& 0.7695 $\pm$ 0.01 
			& 0.753 $\pm$ 0.01 
			& 0.8708 $\pm$ 0.01 
			& 0.8549 $\pm$ 0.03  
			& 0.7552 $\pm$ 0.01 
			& 0.6743 $\pm$ 0.01
			& 0.8060 $\pm$ 0.01  
			& 0.7834 $\pm$ 0.02 
			\\
			\multicolumn{1}{c|}{} 
			& UA \cite{heo2018uncertainty}	
			& 0.7494 $\pm$ 0.03  
			& 0.7469 $\pm$ 0.02  
			& 0.8931 $\pm$ 0.02  
			& 0.8975 $\pm$ 0.01  
			& 0.7873 $\pm$ 0.03 
			& 0.7069 $\pm$ 0.03 
			& 0.7970 $\pm$ 0.01 
			& 0.7959 $\pm$ 0.01 
			\\
			\multicolumn{1}{c|}{} 
			& SAnD \cite{song2018attend}	
			& 0.7919 $\pm$ 0.01
			& 0.7431 $\pm$ 0.01
			& 0.8868 $\pm$ 0.04
			& 0.9086 $\pm$ 0.01
			& 0.7824 $\pm$ 0.03
			& 0.8212 $\pm$ 0.00
			& 0.7075 $\pm$ 0.02
			& 0.8059 $\pm$ 0.01
			\\
			\multicolumn{1}{c|}{} 
			& AdaCare\cite{ma2020adacare}  
			& 0.6593 $\pm$ 0.02
			& 0.7062 $\pm$ 0.02 
			& 0.7793 $\pm$ 0.01
			& 0.8590 $\pm$ 0.01 
			& 0.6935 $\pm$ 0.04
			& 0.7500 $\pm$ 0.02
			& 0.6643 $\pm$ 0.04
			& 0.7302 $\pm$ 0.04
			\\
			\cmidrule{1-10}
			\multicolumn{1}{c|}{} & LSTM    
			& 0.7826 $\pm$ 0.00 
			& 0.7476 $\pm$ 0.01 
			& 0.8880 $\pm$ 0.01 
			& 0.8937 $\pm$ 0.00
			& 0.7948 $\pm$ 0.02 
			& 0.6992 $\pm$ 0.01 
			& \textcolor{red}{0.8030 $\pm$ 0.01} 
			& 0.8013 $\pm$ 0.01 
			\\
			\multicolumn{1}{c|}{} 
			& TRANS~\cite{vaswani2017attention}	
			& 0.7778 $\pm$ 0.01 
			& 0.7537 $\pm$ 0.01 
			& 0.8717 $\pm$ 0.02 
			& 0.8913 $\pm$ 0.00 
			& 0.7862 $\pm$ 0.01 
			& 0.7341 $\pm$ 0.04 
			& 0.8036 $\pm$ 0.01 
			& 0.8026 $\pm$ 0.01 
			\\
			\multicolumn{1}{c|}{} 
			& RETAIN~\cite{choi2016retain}  
			& \textbf{0.7902 $\pm$ 0.01} 
			& \textcolor{red}{0.7377 $\pm$ 0.01}  
			& 0.8835 $\pm$ 0.02 
			& 0.902 $\pm$ 0.00 
			& 0.7726 $\pm$ 0.01
			& 0.7246 $\pm$ 0.04 
			& 0.8106 $\pm$ 0.01 
			& 0.8030 $\pm$ 0.01 
			\\
			\multicolumn{1}{c|}{MTL}
			& RETAIN-Kendall~\cite{kendall2018multi} 
			& 0.7759 $\pm$ 0.01 
			& 0.7546 $\pm$ 0.01
			& 0.8714 $\pm$ 0.03 
			& 0.8949 $\pm$ 0.01 
			& \textbf{0.7953 $\pm$ 0.02} 
			& 0.6789 $\pm$ 0.02 
			& 0.7739 $\pm$ 0.01 
			& 0.7921 $\pm$ 0.01 
			\\
			\multicolumn{1}{c|}{} 
			& UA~\cite{heo2018uncertainty} 
			& 0.7646 $\pm$ 0.03 
			& 0.7479 $\pm$ 0.00 
			& \textbf{0.9271 $\pm$ 0.01} 
			& 0.8935 $\pm$ 0.00 
			& \textcolor{red}{0.7623 $\pm$ 0.03} 
			& \textcolor{red}{0.6952 $\pm$ 0.02} 
			& \textcolor{red}{0.7891 $\pm$ 0.01} 
			& 0.7971 $\pm$ 0.01 
			\\
			\multicolumn{1}{c|}{} 
			& SAnD \cite{song2018attend}	
			& 0.7889 $\pm$ 0.01  
			& 0.7402 $\pm$ 0.01  
			& 0.9000 $\pm$ 0.02  
			& 0.9066 $\pm$ 0.01
			& 0.7683 $\pm$ 0.03
			& 0.8157 $\pm$ 0.02
			& 0.7240 $\pm$ 0.03
			& 0.8062 $\pm$ 0.02
			\\
			\multicolumn{1}{c|}{} 
			& AdaCare\cite{ma2020adacare}  
			& 0.6991 $\pm$ 0.04
			& 0.6357 $\pm$ 0.04
			& 0.7907 $\pm$ 0.06
			& 0.8236 $\pm$ 0.02
			& 0.7189 $\pm$ 0.04
			& 0.6972 $\pm$ 0.03
			& 0.6987 $\pm$ 0.02
			& 0.7234 $\pm$ 0.01
			\\
			\multicolumn{1}{c|}{} 
			& AMTL-LSTM~\cite{lee2016asymmetric} 
			& 0.7577 $\pm$ 0.02 
			& 0.7436 $\pm$ 0.01  
			& 0.8667 $\pm$ 0.04 
			& 0.9049 $\pm$ 0.00 
			& 0.7246 $\pm$ 0.03 
			& 0.6928 $\pm$ 0.03 
			& 0.8073 $\pm$ 0.00 
			& 0.7854 $\pm$ 0.01 
			\\
			\cmidrule{1-10}
			& TP-AMTL (our model) 
			& \textbf{0.7943 $\pm$ 0.01}
			& \textbf{0.7786 $\pm$ 0.02} 
			& \textbf{0.9322 $\pm$ 0.00} 
			& \textbf{0.9113 $\pm$ 0.01} 
			& \textbf{0.7962 $\pm$ 0.01} 
			& \textbf{0.7894 $\pm$ 0.02} 
			& \textbf{0.819 $\pm$ 0.02} 
			& \textbf{0.8316 $\pm$ 0.01} \\
			\bottomrule
		\end{tabular}
	}
\end{table*}
\begin{table*}[ht!]
	\small
	\caption{\textbf{MIMIC-III Respiratory Failure (RF) - $37,818$ instances.} The reported numbers are average AUROC and standard error over \textbf{five runs}. The numbers colored in \textcolor{red}{red} for MTL models denote accuracies lower than those of their STL counterparts.}
	\label{respiratoryfailure_30000}
	\centering
	\resizebox{\textwidth}{!}{
		\begin{tabular}{cccccccccc}
			\toprule
			\multicolumn{2}{c}{Models} & \multicolumn{7}{c}{Tasks}       &             \\
			\cmidrule(r){3-9}
			&  & Hypoxemia & Hypercapnia & VQ Mismatch & Acidosis & RF & Cyanosis & Mortality & Average\\
			\cmidrule{1-10}
			\multicolumn{1}{c|}{} 
			& LSTM   
			& 0.7857 $\pm$ 0.01 
			& 0.7899 $\pm$ 0.00
			& 0.9248 $\pm$ 0.00 
			& 0.9203 $\pm$ 0.00 
			& 0.811 $\pm$ 0.01
			& 0.8333 $\pm$ 0.02
			& 0.8432 $\pm$ 0.01 
			& 0.8440 $\pm$ 0.05
			\\
			\multicolumn{1}{c|}{} 
			& Transformer \cite{vaswani2017attention} 
			& 0.7863 $\pm$ 0.01
			& 0.7808 $\pm$ 0.00
			& 0.9110 $\pm$ 0.01
			& 0.9286 $\pm$ 0.00 
			& 0.8029 $\pm$ 0.01
			& 0.8140 $\pm$ 0.01
			& 0.8416 $\pm$ 0.00
			& 0.8379 $\pm$ 0.00
			\\
			\multicolumn{1}{c|}{STL} 
			& RETAIN \cite{choi2016retain}   
			& 0.7968 $\pm$ 0.00 
			& 0.7802 $\pm$ 0.00 
			& 0.9276 $\pm$ 0.00 
			& 0.9219 $\pm$ 0.00  
			& 0.8120 $\pm$ 0.00 
			& 0.8253 $\pm$ 0.01 
			& 0.8503 $\pm$ 0.01 
			& 0.8449 $\pm$ 0.05 
			\\
			\multicolumn{1}{c|}{} 
			& UA \cite{heo2018uncertainty}	
			& 0.8315 $\pm$ 0.01  
			& \textbf{0.8271 $\pm$ 0.00}  
			& 0.846 $\pm$ 0.02  
			& 0.8002 $\pm$ 0.02  
			& 0.8020 $\pm$ 0.01 
			& 0.8304 $\pm$ 0.01 
			& 0.8651 $\pm$ 0.00 
			& 0.8432 $\pm$ 0.04 
			\\
			\multicolumn{1}{c|}{} 
			& SAnD \cite{song2018attend}	
			& 0.8195 $\pm$ 0.04
			& 0.7864 $\pm$ 0.00
			& 0.9341 $\pm$ 0.00
			& 0.9324 $\pm$ 0.00
			& 0.8214 $\pm$ 0.01
			& 0.8617 $\pm$ 0.00
			& 0.8565 $\pm$ 0.01
			& 0.8589 $\pm$ 0.01
			\\
			\multicolumn{1}{c|}{} 
			& AdaCare\cite{ma2020adacare}  
			& 0.6511 $\pm$ 0.01
			& 0.6673 $\pm$ 0.00
			& 0.7817 $\pm$ 0.02
			& 0.7588 $\pm$ 0.01
			& 0.7394 $\pm$ 0.02
			& 0.7089 $\pm$ 0.01
			& 0.6734 $\pm$ 0.03
			& 0.7115 $\pm$ 0.01
			\\
			\cmidrule{1-10}
			\multicolumn{1}{c|}{} 
			& LSTM    
			& 0.8021 $\pm$ 0.00 
			& \textcolor{red}{0.7843 $\pm$ 0.00} 
			& \textcolor{red}{0.8968 $\pm$ 0.05} 
			& \textcolor{red}{0.9197 $\pm$ 0.00} 
			& 0.816 $\pm$ 0.00 
			& \textcolor{red}{0.8261 $\pm$ 0.01} 
			& 0.8453 $\pm$ 0.00 
			& 0.8414 $\pm$ 0.05 
			\\
			\multicolumn{1}{c|}{} 
			& Transformer \cite{vaswani2017attention}	
			& 0.7937 $\pm$ 0.00 
			& 0.7781 $\pm$ 0.00 
			& 0.9247 $\pm$ 0.00 
			& 0.9234 $\pm$ 0.00 
			& 0.8231 $\pm$ 0.00 
			& 0.8139 $\pm$ 0.00 
			& 0.8330 $\pm$ 0.00 
			& 0.8414 $\pm$ 0.05 
			\\
			\multicolumn{1}{c|}{MTL} 
			& RETAIN  
			& 0.7992 $\pm$ 0.00 
			& \textcolor{red}{0.7794 $\pm$ 0.00}  
			& 0.9346 $\pm$ 0.00 
			& \textcolor{red}{0.9199 $\pm$ 0.00} 
			& 0.8139 $\pm$ 0.01 
			& \textcolor{red}{0.824 $\pm$ 0.00} 
			& \textcolor{red}{0.8313 $\pm$ 0.01} 
			& 0.8432 $\pm$ 0.06 
			\\
			\multicolumn{1}{c|}{} 
			& UA 
			& 0.8316 $\pm$ 0.00 
			& \textcolor{red}{0.8103 $\pm$ 0.00} 
			& 0.943 $\pm$ 0.00 
			& 0.9354 $\pm$ 0.00 
			& 0.8397 $\pm$ 0.00 
			& \textbf{0.8727 $\pm$ 0.00} 
			& \textbf{0.8754 $\pm$ 0.00} 
			& 0.8726 $\pm$ 0.05 
			\\
			\multicolumn{1}{c|}{} 
			& RETAIN-Kendall 
			& 0.8028 $\pm$ 0.00
			& 0.7794 $\pm$ 0.00 
			& 0.9274 $\pm$ 0.00 
			& 0.9168 $\pm$ 0.00 
			& 0.8207 $\pm$ 0.00 
			& 0.8117 $\pm$ 0.00 
			& 0.8352 $\pm$ 0.01 
			& 0.8420 $\pm$ 0.05 
			\\
			\multicolumn{1}{c|}{} 
			& SAnD \cite{song2018attend}	
			& 0.8253 $\pm$ 0.00
			& 0.8108 $\pm$ 0.00
			& 0.9421 $\pm$ 0.00
			& 0.9355 $\pm$ 0.00
			& 0.8366 $\pm$ 0.01
			& 0.8646 $\pm$ 0.00
			& 0.8593 $\pm$ 0.00
			& 0.8678 $\pm$ 0.00
			\\
			\multicolumn{1}{c|}{} 
			& AdaCare\cite{ma2020adacare}  
			& 0.7683 $\pm$ 0.02
			& 0.7301 $\pm$ 0.02
			& 0.8920 $\pm$ 0.00
			& 0.8805 $\pm$ 0.01
			& 0.7671 $\pm$ 0.02
			& 0.7810 $\pm$ 0.02
			& 0.7585 $\pm$ 0.01
			& 0.7968 $\pm$ 0.01
			\\
			\multicolumn{1}{c|}{} 
			& AMTL-LSTM \cite{lee2016asymmetric} 
			& 0.8049 $\pm$ 0.01 
			& 0.7937 $\pm$ 0.01  
			& 0.9146 $\pm$ 0.00 
			& 0.9228 $\pm$ 0.00 
			& 0.7962 $\pm$ 0.02 
			& 0.8434 $\pm$ 0.02 
			& 0.8510 $\pm$ 0.01 
			& 0.8467 $\pm$ 0.05 
			\\
			\cmidrule{1-10}
			& TP-AMTL (our model) 
			& \textbf{0.8435 $\pm$ 0.01}
			& \textbf{0.8291 $\pm$ 0.02} 
			& \textbf{0.9456 $\pm$ 0.00} 
			& \textbf{0.9399 $\pm$ 0.01} 
			& \textbf{0.8405 $\pm$ 0.01} 
			& \textbf{0.8757 $\pm$ 0.02} 
			& \textbf{0.8761 $\pm$ 0.02} 
			&  \textbf{0.8739 $\pm$ 0.01}
			\\
			\bottomrule
		\end{tabular}
	}
\end{table*}
\begin{table*}[ht!]
	\small
	\caption{\textbf{MIMIC-III Respiratory Failure (RF) - $7$ day time interval.} The reported numbers are average AUROC and standard error over \textbf{five runs}. The numbers colored in \textcolor{red}{red} for MTL models denote accuracies lower than those of their STL counterparts.}
	\label{respiratoryfailure_7days}
	\centering
	\resizebox{\textwidth}{!}{
		\begin{tabular}{cccccccccc}
			\toprule
			\multicolumn{2}{c}{Models} & \multicolumn{7}{c}{Tasks}       &             \\
			\cmidrule(r){3-9}
			&  & Hypoxemia & Hypercapnia & VQ Mismatch & Acidosis & RF & Cyanosis & Mortality & Average\\
			\cmidrule{1-10}
			\multicolumn{1}{c|}{} 
			& LSTM   
			& 0.7894 $\pm$ 0.03
			& 0.7461 $\pm$ 0.01
			& 0.9095 $\pm$ 0.02
			& 0.8732 $\pm$ 0.01
			& 0.7709 $\pm$ 0.03 
			& \textbf{0.8090 $\pm$ 0.01}
			& 0.7566 $\pm$ 0.05
			& 0.8078 $\pm$ 0.01
			\\
			\multicolumn{1}{c|}{} 
			& Transformer \cite{vaswani2017attention} 
			& 0.7823 $\pm$ 0.02
			& \textbf{0.9070 $\pm$ 0.08}
			& \textbf{0.9370 $\pm$ 0.02} 
			& 0.8760 $\pm$ 0.00 
			& 0.7701 $\pm$ 0.03 
			& 0.7911 $\pm$ 0.01 
			& 0.7225 $\pm$ 0.04 
			& 0.8266 $\pm$ 0.02
			\\
			\multicolumn{1}{c|}{STL} 
			& RETAIN \cite{choi2016retain}   
			& 0.7902 $\pm$ 0.01  
			& 0.7597 $\pm$ 0.01
			& 0.9036 $\pm$ 0.02
			& 0.8048 $\pm$ 0.01 
			& 0.7820 $\pm$ 0.02
			& \textbf{0.8062 $\pm$ 0.01}
			& \textbf{0.7884 $\pm$ 0.06}
			& 0.8050 $\pm$ 0.02
			\\
			\multicolumn{1}{c|}{} 
			& UA \cite{heo2018uncertainty}	
			& 0.7697 $\pm$ 0.00 
			& \textbf{0.7729 $\pm$ 0.02}
			& 0.9166 $\pm$ 0.03
			& 0.8816 $\pm$ 0.02
			& 0.7915 $\pm$ 0.04
			& \textbf{0.8073 $\pm$ 0.01}
			& \textbf{0.7872 $\pm$ 0.02}
			& 0.8181 $\pm$ 0.01
			\\
			\multicolumn{1}{c|}{} 
			& SAnD \cite{song2018attend}	
			& 0.7895 $\pm$ 0.01
			& 0.7608 $\pm$ 0.01
			& 0.7589 $\pm$ 0.03
			& \textbf{0.9646 $\pm$ 0.02}
			& \textbf{0.8856 $\pm$ 0.01}
			& 0.7611 $\pm$ 0.05
			& 0.6952 $\pm$ 0.04
			& 0.8022 $\pm$ 0.01
			\\
			\multicolumn{1}{c|}{} 
			& AdaCare\cite{ma2020adacare}  
			& 0.7494 $\pm$ 0.01 
			& 0.7410 $\pm$ 0.01
			& 0.8786 $\pm$ 0.01
			& 0.8923 $\pm$ 0.00 
			& 0.6945 $\pm$ 0.05
			& 0.7771 $\pm$ 0.01
			& \textbf{0.7922 $\pm$ 0.01}
			& 0.7893 $\pm$ 0.02
			\\
			\cmidrule{1-10}
			\multicolumn{1}{c|}{} 
			& LSTM    
			& 0.7916 $\pm$ 0.02
			& 0.7530 $\pm$ 0.01
			& 0.9287 $\pm$ 0.03
			& 0.8789 $\pm$ 0.01
			& 0.7963 $\pm$ 0.03
			& 0.8004 $\pm$ 0.01
			& 0.7728 $\pm$ 0.03
			& 0.8174 $\pm$ 0.01
			\\
			\multicolumn{1}{c|}{} 
			& Transformer \cite{vaswani2017attention}	
			& 0.7746 $\pm$ 0.01  
			& \textcolor{red}{0.7403 $\pm$ 0.01}
			& 0.9519 $\pm$ 0.01
			& 0.8778 $\pm$ 0.01
			& 0.7855 $\pm$ 0.01
			& 0.7872 $\pm$ 0.01
			& 0.7153 $\pm$ 0.02
			& \textcolor{red}{0.8047 $\pm$ 0.00}
			\\
			\multicolumn{1}{c|}{MTL} 
			& RETAIN  
			& \textcolor{red}{0.7686 $\pm$ 0.01} 
			& 0.7541 $\pm$ 0.01
			& \textcolor{red}{0.8803 $\pm$ 0.04}
			& 0.8787 $\pm$ 0.01
			& 0.7738 $\pm$ 0.02
			& 0.8043 $\pm$ 0.01
			& 0.7781 $\pm$ 0.02
			& 0.8054 $\pm$ 0.01
			\\
			\multicolumn{1}{c|}{} 
			& UA 
			& \textbf{0.8170 $\pm$ 0.01}  
			& 0.7684 $\pm$ 0.01
			& \textbf{0.9446 $\pm$ 0.02}
			& 0.8856 $\pm$ 0.01
			& 0.8437 $\pm$ 0.01
			& \textbf{0.8190 $\pm$ 0.02}
			& \textbf{0.8000 $\pm$ 0.02}
			& \textbf{0.8398 $\pm$ 0.01}
			\\
			\multicolumn{1}{c|}{} 
			& RETAIN-Kendall 
			& \textcolor{red}{0.7784 $\pm$ 0.01}
			& 0.7556 $\pm$ 0.01
			& 0.9163 $\pm$ 0.04
			& 0.8718 $\pm$ 0.01 
			& 0.7859 $\pm$ 0.03
			& \textcolor{red}{0.7884 $\pm$ 0.02}
			& 0.7603 $\pm$ 0.02
			& 0.8081 $\pm$ 0.01
			\\
			\multicolumn{1}{c|}{} 
			& SAnD \cite{song2018attend}	
			& 0.7802 $\pm$ 0.01
			& 0.7708 $\pm$ 0.01
			& \textbf{0.9611 $\pm$ 0.01 }
			& \textcolor{red}{0.8839 $\pm$ 0.01}
			& \textcolor{red}{0.7821 $\pm$ 0.02}
			& \textbf{0.8062 $\pm$ 0.01}
			& 0.7416 $\pm$ 0.02
			& 0.8180 $\pm$ 0.01
			\\
			\multicolumn{1}{c|}{} 
			& AdaCare\cite{ma2020adacare}  
			& 0.6926 $\pm$ 0.02
			& 0.6707 $\pm$ 0.02
			& 0.7893 $\pm$ 0.07
			& 0.7939 $\pm$ 0.03
			& 0.7437 $\pm$ 0.02
			& 0.7080 $\pm$ 0.03
			& 0.6845 $\pm$ 0.03
			& 0.7261 $\pm$ 0.02
			\\
			\multicolumn{1}{c|}{} 
			& AMTL-LSTM \cite{lee2016asymmetric} 
			& 0.7898 $\pm$ 0.01  
			& 0.7527 $\pm$ 0.02
			& 0.8717 $\pm$ 0.04
			& 0.8754 $\pm$ 0.01
			& 0.7274 $\pm$ 0.04
			& 0.7923 $\pm$ 0.02
			& 0.7659 $\pm$ 0.04
			& 0.7965 $\pm$ 0.01
			\\
			\cmidrule{1-10}
			& TP-AMTL (our model) 
			& \textbf{0.8048 $\pm$ 0.01}
			& \textbf{0.7717 $\pm$ 0.02} 
			& \textbf{0.9465 $\pm$ 0.00} 
			& \textbf{0.9051 $\pm$ 0.01} 
			& \textbf{0.8613 $\pm$ 0.01} 
			& \textbf{0.8060 $\pm$ 0.02} 
			& \textbf{0.7813 $\pm$ 0.02} 
			&  \textbf{0.8395 $\pm$ 0.01} 
			\\
			\bottomrule
		\end{tabular}
	}
\end{table*}

\section{Quantitative evaluation on clinical time-series prediction tasks: MIMIC III-Heart Failure, Respiratory Failure}
\label{additionaldata}
Here, we provide the experimental results of our model and other baselines on the additional dataset: MIMIC III-Heart Failure, Respiratory Failure. \textcolor{navyblue}{Table}~\ref{heartfailure} and \textcolor{navyblue}{Table}~\ref{respiratoryfailure_original} shows that our model still outperforms other baselines, which indicates that our method can generalize well on a variety of time-series datasets. Furthermore, \textcolor{navyblue}{Table}~\ref{respiratoryfailure_30000} and \textcolor{navyblue}{Table}~\ref{respiratoryfailure_7days} shows that our model also generalize well to larger datasets. 

\section{Clinical Interpretation of generated uncertainty and knowledge transfer between tasks}
\label{sec: interpret}
In this section, we further describe the interpretation of several example patients using generated uncertainty and knowledge transfer across timesteps.

\subsection{MIMIC-III Heart Failure} 
\begin{figure*}[ht!]
	\begin{subfigure}{.5\textwidth}
		\centering
		\includegraphics[width=\linewidth]{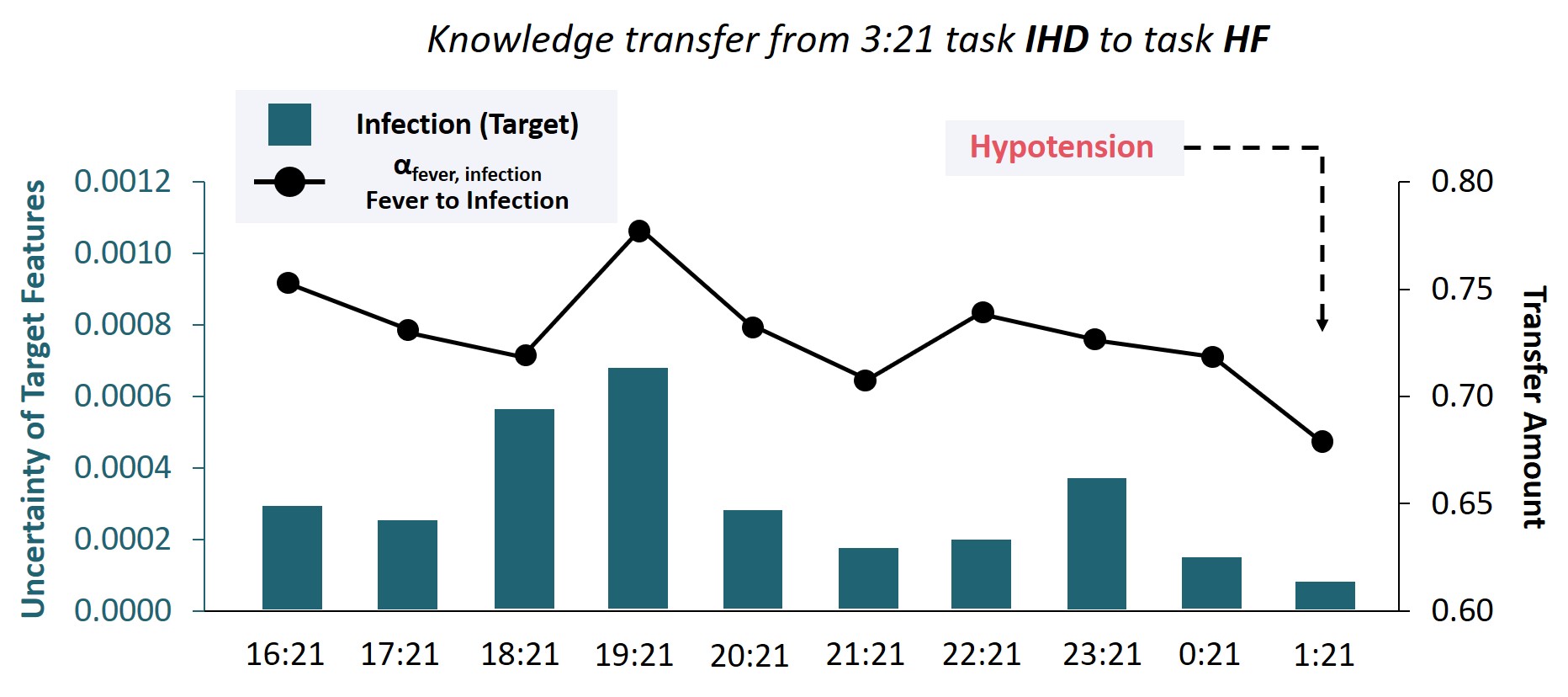}
		\caption{KT from single timestep and Target Uncertainty}
		\label{Target_Heartfailure}
	\end{subfigure}
	\begin{subfigure}{.5\textwidth}
		\centering
		\includegraphics[width=\linewidth]{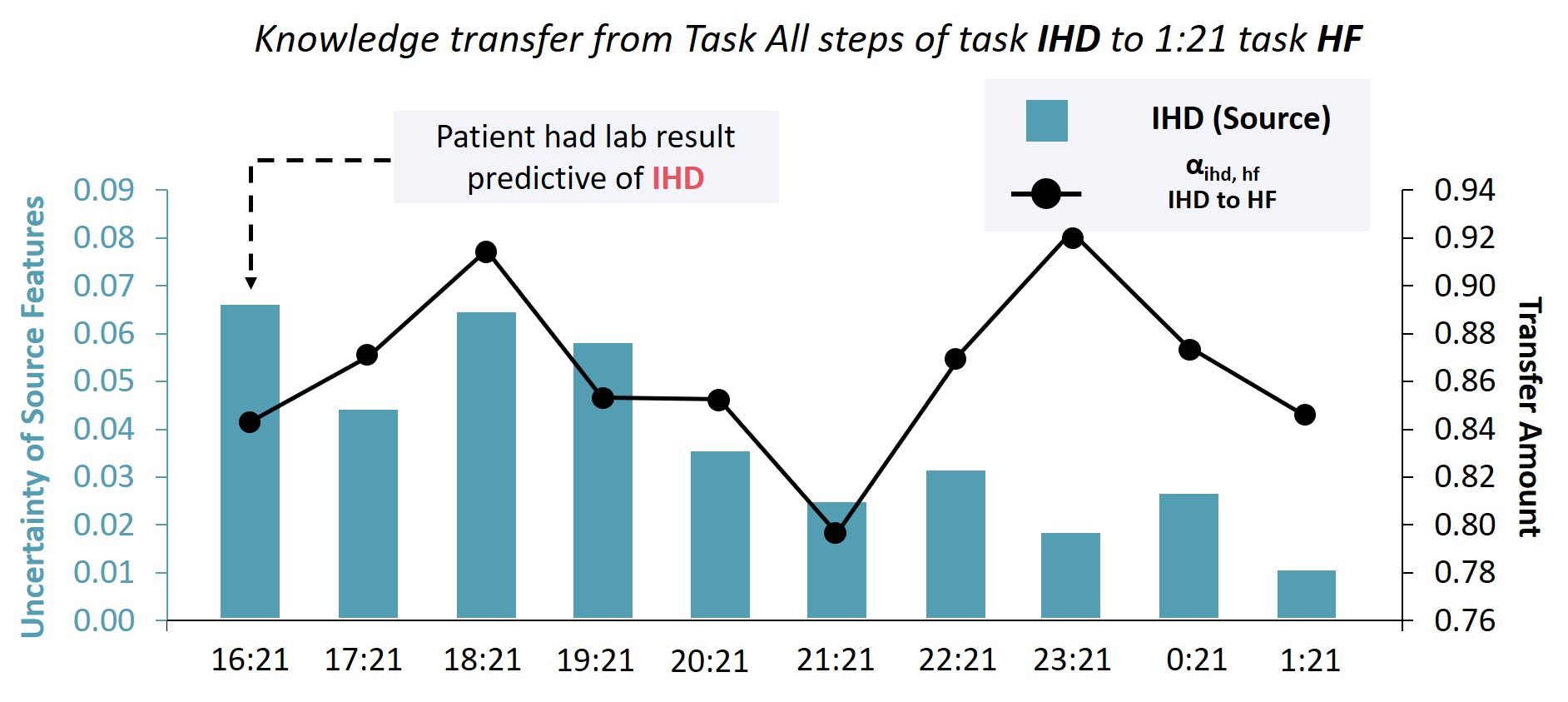}
		\caption{KT to single timestep and Source Uncertainty}
		\label{Source_Heartfailure}
	\end{subfigure}
	\caption{\small \textbf{Uncertainty and Knowledge Transfer(KT) : Example case of MIMIC III - Heart Failure dataset} where the changes in the amount of uncertainty at certain timesteps are correlated with clinical events. We denote the timesteps with noticeable changes in uncertainty and knowledge transfer with blue boxes.}
	\label{fig:HFInterpretation}
\end{figure*}

Interpretation on another example patient from MIMIC III-heart failure dataset is plotted on \textcolor{navyblue}{Figure} ~\ref{fig:HFInterpretation}. This example patient is finally diagnosed with congestive heart failure on Chest X-ray. During the admission period, the troponin level of this patient was elevated, which is not diagnostic \cite{reichlin2009early}, but implying that this patient had a cardiac event. Given cardiac events, hypotension occurred in $1:21$ (\textcolor{navyblue}{Table} ~\ref{Clinicalevents_HF}, \textcolor{navyblue}{Figure} ~\ref{fig:HFInterpretation}) can be explained to be related to final diagnosis heart failure. As the patient SBP decreases to $90$ and DBP to $30$ around $1:21$ (\textcolor{navyblue}{Table} ~\ref{Clinicalevents_HF}), the uncertainty of target task \textit{Heart Failure} decreases in \textcolor{navyblue}{Figure} ~\ref{Target_Heartfailure}. Knowledge transfer starts to drop as the knowledge from the target task becomes more important than that of the source task. We can also see that the trend of knowledge transfer follows the trend of target uncertainty. Furthermore, troponin increased in $16:21$ implies ongoing myocardial stress, which can be expressed as constantly lowered uncertainty of source task \textit{ischemic heart disease} among the window period we plotted on \textcolor{navyblue}{Figure} ~\ref{Source_Heartfailure}. As the uncertainty of source task decreases, knowledge transfer to target task \textit{heart failure} kept increasing till $23:21$. However, as a patient condition related to heart function, especially blood pressure starts to decrease and knowledge from the target task gets important, knowledge  transfer starts to decrease after $23:21$.

\begin{table*}[ht!]
	\caption{\small Clinical Events in selected medical records for case studies. \textbf{HR} - Heart Rate, \textbf{RR} - Respiratory Rate, \textbf{SBP} - Systolic arterial blood pressure, \textbf{DBP} - Diastolic arterial blood pressure}
	\label{Clinicalevents_HF}
	\hspace{-0.25in}
	\resizebox{15cm}{!}{
		\begin{tabular}{lll|llllllll|llllllllll}
			&  &  & HR & RR & SBP & DBP & Troponin-c &  &  &  & HR & RR & SBP & DBP &  &  &  &  &  &  \\ \cline{3-8} \cline{11-20}
			&  & 16:21 & 93 & 18 & 139 & 59 & 1.26 &  &  & 23:21 & 120 & 13 & 97 & 36 &  &  &  &  &  &  \\
			&  & 18:21 & 84 & 21 & \textbf{98} & \textbf{38} &  &  &  & 0:21 & 113 & 23 & 102 & 36 &  &  &  &  &  &  \\
			&  & 19:21 & 81 & 21 & \textbf{95} & \textbf{36} &  &  &  & \textbf{1:21} & \textbf{128} & \textbf{26} & \textbf{91} & \textbf{30} &  &  &  &  &  & 
		\end{tabular}
	}
\end{table*}

\subsection{MIMIC-III Respiratory Failure} 
\begin{figure*}[ht!]
	\begin{subfigure}{.5\textwidth}
		\centering
		\includegraphics[width=\linewidth]{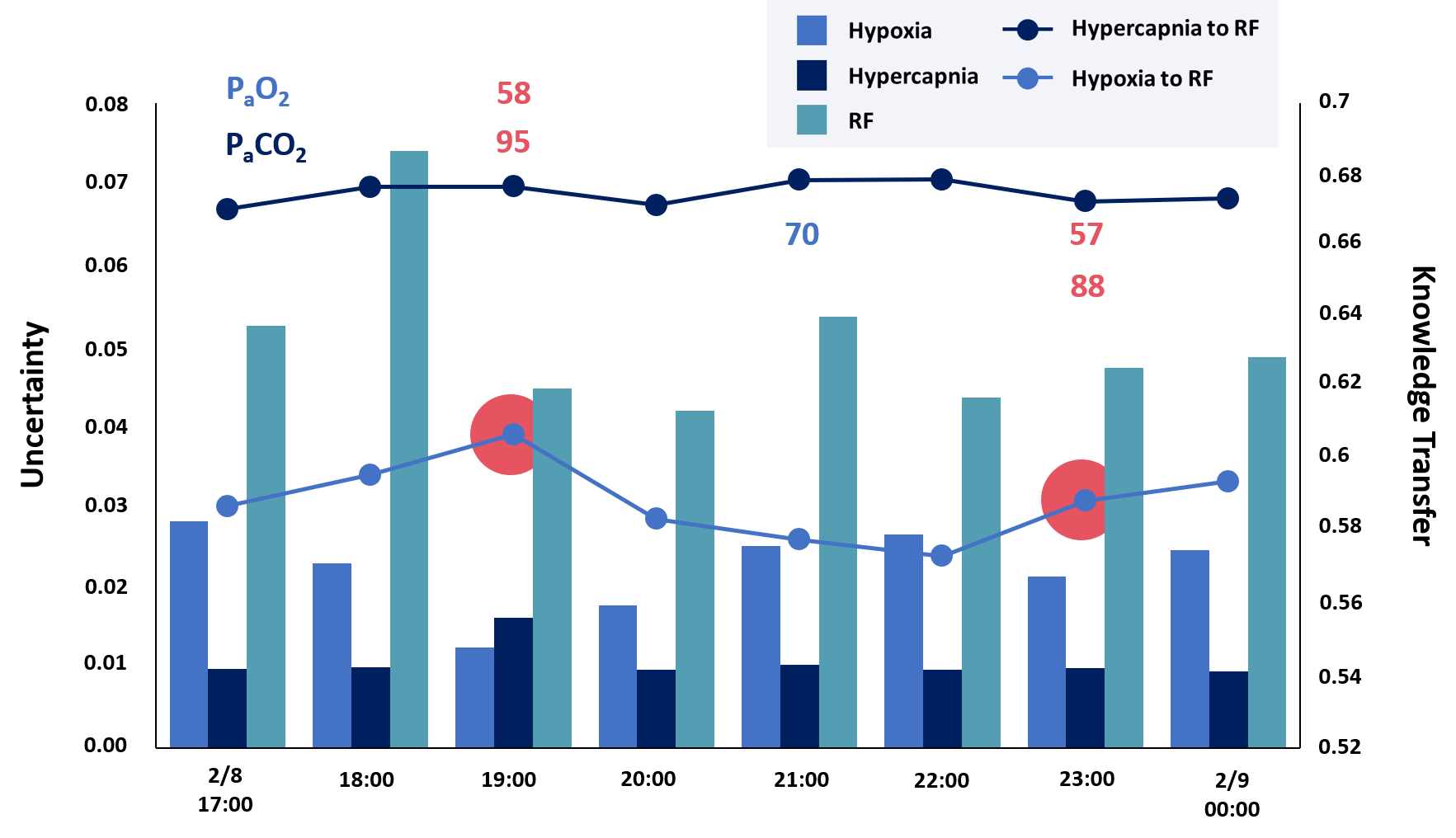}
		\caption{KT from Hypoxemia, Hypercapnia to RF}
		\label{Knowledge_Transfer_RF}
	\end{subfigure}
	\begin{subfigure}{.5\textwidth}
		\centering
		\includegraphics[width=\linewidth]{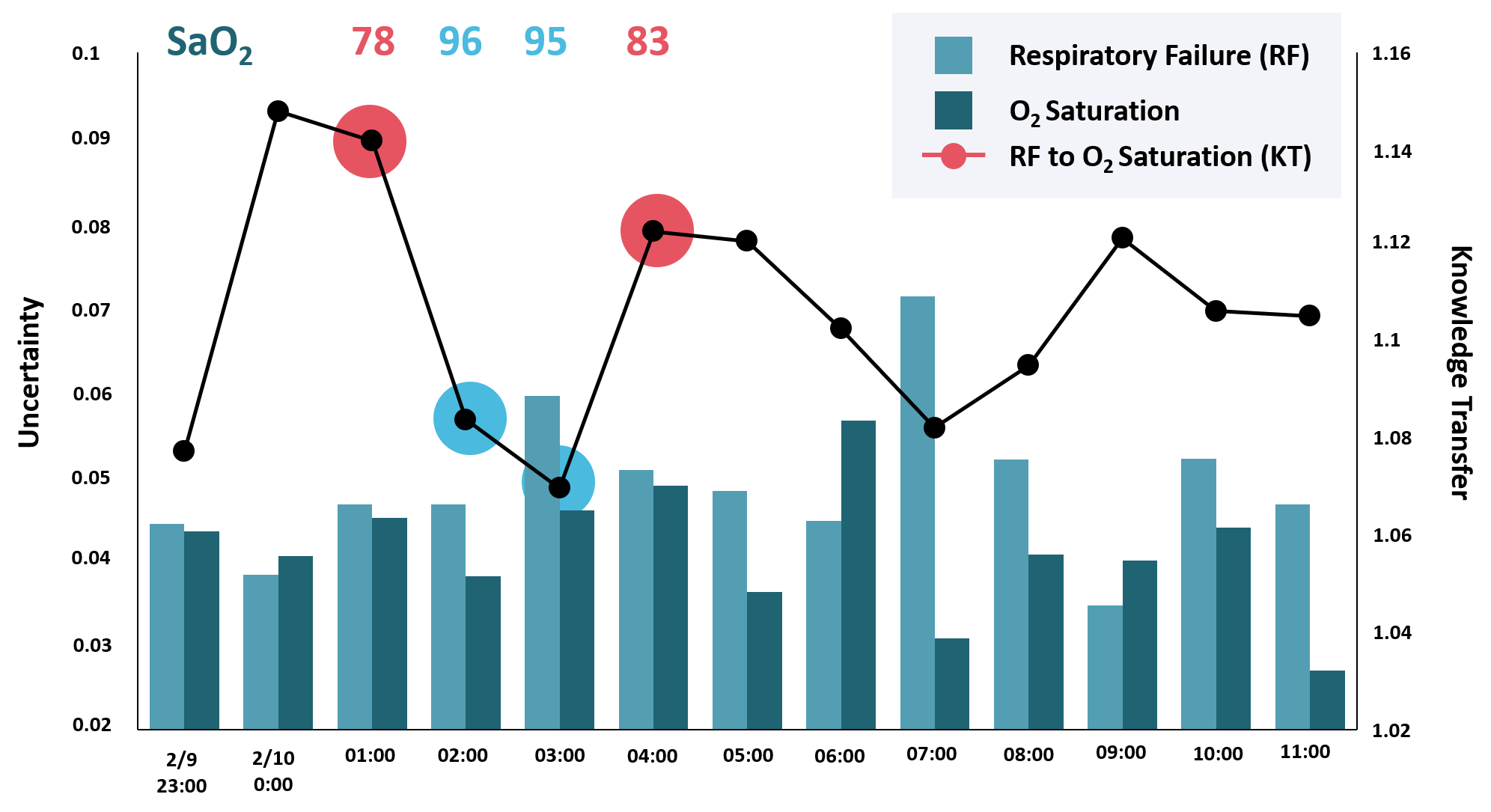}
		\caption{KT from RF to $S_aO_2$}
		\label{Knowledge_Transfer_From_RF_to_SaO2}
	\end{subfigure}
	\caption{\small \textbf{Visualizations of the amount of uncertainty and normalized knowledge transfer for example cases} where the changes in the amount of uncertainty at certain timesteps are correlated with clinical events. We denote the timesteps with noticeable changes in uncertainty and knowledge transfer with Red(fatal clinical condition predictive of target task), Blue(improved patient condition)}
	\label{fig:Interpretation_RF}
\end{figure*}

With the help of a physician, we further analyze how generated transfer weights and uncertainties between tasks can be used to track changes in the relationships between clinical events and  (see \textcolor{navyblue}{Figure}~\ref{fig:Interpretation_RF}). We first consider an example record of a patient from the MIMIC-III Respiratory Failure(RF) dataset who was found to be in hypercarbic(high arterial $CO_2$ pressure (high $P_aCO_2$, hypercapnia)) respiratory failure. \textcolor{navyblue}{Figure}~\ref{Knowledge_Transfer_RF} shows the amount of knowledge transfer from task \emph{Hypoxia} and \emph{Hypercapnia} at $5:00AM$ to all later timesteps of task \emph{RF}. As patient experience hypoxia (low arterial $O_2$ pressure (low $P_aO_2$)) during 12 ($19:00$) to 17 ($23:00$ AM) hours of admission, the uncertainty of hypoxia (blue bar chart of \textcolor{navyblue}{Figure}~\ref{Knowledge_Transfer_RF}) drops accordingly, and knowledge transfers from task \emph{Hypoxia} to task \emph{RF} (dark blue line graph in \textcolor{navyblue}{Figure}~\ref{Knowledge_Transfer_RF}). As patient was hypercarbic on admission, uncertainty of the task \emph{Hypercapnia} kept its low value (dark blue bar chart in \textcolor{navyblue}{Figure}~\ref{Knowledge_Transfer_RF}) and knowledge transfer to task \emph{RF} is constantly high (dark blue line chart in \textcolor{navyblue}{Figure}~\ref{Knowledge_Transfer_RF}). After $23:00$, $P_aO_2$ increases back to knowledge transfer drops as the uncertainty of task \emph{hypoxia} increases. Change in learned knowledge graph can also be used for treatment planning, such as deciding whether the patient needs prompt oxygen infusion or not. Later in timestep $20$ ($01:00$), the $O_2$ Saturation of this patient drops to $78$ (see \textcolor{navyblue}{Figure} ~\ref{Knowledge_Transfer_From_RF_to_SaO2}) and uncertainty of task $S_aO_2$ drops accordingly as a result of respiratory failure (dark green bar chart in \textcolor{navyblue}{Figure} ~\ref{Knowledge_Transfer_From_RF_to_SaO2}). As the target uncertainty drops, knowledge transfer drops accordingly. After treatment $S_aO_2$ increases up to $94$ and then drops again to $83$ in timestep $23$ ($4:00$), where knowledge transfer from source task \emph{RF} increases and drops accordingly in \textcolor{navyblue}{Figure} ~\ref{Knowledge_Transfer_From_RF_to_SaO2} (black line).

\section{Broader Impact}

In this paper, we present Temporal Probabilistic Asymmetric Multi-Task Learning (TPAMTL) model, which is appropriate for various multi-task time-series predictions with strong performance while preventing negative transfer. Our model is especially applicable to safety-critical clinical risk predictions, as it provides a reliable explanation of its final prediction on each task at each timestep. Here in this section, we introduce the clinical and social impact of our model. 

\paragraph{Safe and Reliable AI for Clinical risk prediction}
With the actual knowledge transfer graph plotted with uncertainty obtained at each timestep, healthcare professionals could interpret the behaviors of model according to actual clinical events in multiple clinical risk prediction tasks (\textcolor{navyblue}{Figure} 4, \textcolor{navyblue}{Supplementary Figure} 11, 12, 13). By checking the knowledge transfer graph and the amount of uncertainty measured for each task at each timestep, our model can provide medical professionals with a better explanation of the underlying model behavior. Based on this explanation, medical experts can have a clue on the direction of knowledge transfer, or say interaction, between tasks, which might help them decide future action: whether to consult other departments for medical opinion, to run a further test to confirm the predicted event, or to come up with better treatment plan. This interpretability of our model will be useful in building a safe time-series analysis system for large-scale settings where both the number of time-series data instances and timestep are extremely large, such that manual analysis is impractical. 

\paragraph{Monitoring clinical events of critical ill patient}
Due to the unprecedented pandemic outbreak of coronavirus infection (COVID-19) this year, hospitals worldwide are suffering from the lack of healthcare professionals and facilities to monitor and care for patients \cite{xie2020critical}. In this era of difficulty, machine learning models can aid the diagnosis~\cite{li2020artificial, wang2020covid}, drug repositioning \cite{richardson2020baricitinib}, predicting patient condition, and epidemiological trends of disease~\cite{mccall2020covid}. One example of this large-scale setting would be the situation to monitor COVID-19, where the number of newly diagnosed patients is increasing daily~\cite{novel2020epidemiological}, and there is a need to follow-up on multiple outcomes (pneumonia~\cite{woo2004relative}, cardiac disease related to myocardial injury~\cite{guo2020cardiovascular}, etc.) of this infection, also in relation to underlying disease each patient had, where no pathophysiological and case information is available. In this situation when the number of patients exceeds the capacity of the healthcare system, a prediction model with a better capacity of handling large-scale data can help reduce the burden of healthcare staff. Not only the acute clinical conditions (e.g., infection), but also the chronic clinical conditions, such as conditions and abnormalities related to heart failure (Task MIMIC-III Heart Failure) or respiratory failure (Task MIMIC-III Respiratory Failure), can be monitored with our model, as our model provides long-term prediction without degeneration of performance, still providing reliable information.

\begin{table*}[ht!]
	\centering
	\footnotesize
	\caption{Feature information of MIMIC III - Infection dataset}
	\label{mimic3feature}
	\begin{tabular}{ccc}
		\toprule
		\multicolumn{1}{c}{\textbf{Features}} & 
		\multicolumn{1}{c}{\textbf{Item-ID}} & 
		\multicolumn{1}{c}{\textbf{Name of Item}} \\ \hline
		Age & NA & \begin{tabular}[c]{@{}l@{}}initime\\ dob\end{tabular} \\ \hline
		Sex & NA & gender \\ \hline
		& 211 & Heart Rate \\
		\multirow{-2}{*}{Heart Rate} & 22045 & Heart Rate \\ \hline
		& 51 & Systolic Blood Pressure \\
		& 442 & Systolic Blood Pressure \\
		& 455 & Systolic Blood Pressure \\
		& 6701 & Systolic Blood Pressure \\
		& 220179 & Systolic Blood Pressure \\
		\multirow{-6}{*}{Systolic Blood Pressure} & 220050 & Systolic Blood Pressure \\ \hline
		& 8368 & Diastolic Blood Pressure \\
		& 8440 & Diastolic Blood Pressure \\
		& 8441 & Diastolic Blood Pressure \\
		& 8555 & Diastolic Blood Pressure \\
		& 220051 & Diastolic Blood Pressure \\
		\multirow{-5}{*}{Diastolic Blood Pressure} & 220180 & Diastolic Blood Pressure \\ \hline
		& 223900 & GCS-Verbal Response \\ 
		& 223901 & GCS-Motor Response \\ 
		\multirow{-3}{*}{Glasgow Coma Scale} & 220739 & GCS-Eye Opening \\ \hline
		& 225433 & Chest Tube Placed \\
		& 5456 & Chest Tube \\
		& 225445 & Paracentesis \\
		& 225446 & PEG Insertion \\
		& 225399 & Lumbar Puncture \\
		& 5939 & Lumbar drain \\
		& 225469 & OR Received \\
		& 225442 & Liver Biopsy \\
		& 224264 & PICC Line \\
		& 224560 & PA Catheter \\
		& 225430 & Cardiac Cath \\
		& 225315 & Tunneled (Hickman) Line \\
		& 226475 & Intraventricular Drain Inserted \\
		\multirow{-14}{*}{Invasive procedures}& 5889 & Bladder cath \\ \hline
		& 225434 & Colonoscopy \\
		& 225439 & Endoscopy \\
		\multirow{-3}{*}{Endoscopic Procedure} & 227550 & ERCP \\ \hline
		& 224385 & Intubation \\
		& 225448 & Percutaneous Tracheostomy \\
		& 225468 & Unplanned Extubation (patient-initiated) \\
		& 225477 & Unplanned Extubation (non-patient initiated) \\
		& 226237 & Open Tracheostomy \\
		\multirow{-6}{*}{Intubation / Unplanned Extubation} & 225792 & Invasive Ventilation \\ \hline
		& 772 & Albumin (\textgreater{}3.2) \\
		& 1521 & Albumin \\
		& 227456 & Albumin \\
		& 3727 & Albumin (3.9-4.8) \\
		& 226981 & Albumin\_ApacheIV \\
		\multirow{-6}{*}{Albumin}  & 226982 & AlbuminScore\_ApacheIV \\ \hline
		& 220650 & Total Protein(6.5-8) \\  
		& 849 & Total Protein(6.5-8) \\
		& 3807 & Total Protein \\
		& 1539 & Total Protein(6.5-8) \\
		\multirow{-5}{*}{Total Protein} & 220650 & Total Protein(6.5-8) \\
		\bottomrule
	\end{tabular}
\end{table*}

\begin{table*}[ht!]
	\centering
	\footnotesize
	\caption{Feature information of MIMIC III - Infection dataset: continued}
	\begin{tabular}{ccc}
		\toprule
		\multicolumn{1}{c}{\textbf{Features}} & 
		\multicolumn{1}{c}{\textbf{Item-ID}} & 
		\multicolumn{1}{c}{\textbf{Name of Item}} \\ \hline
		& 225441 & Hemodialysis \\
		& 225805 & Peritoneal Dialysis \\
		& 226477 & Temporary Pacemaker Wires Inserted \\
		& 224270 & Dialysis Catheter \\
		& 225802 & Dialysis - CRRT \\
		\multirow{-6}{*}{Dialysis} & 225805 & Peritoneal Dialysis \\ \hline
		& 4929 & Prednisolone \\
		& 7772 & Predisolone \\
		& 6753 & Prednisilone gtts \\
		& 6111 & prednisone \\
		& 8309 & prednisolone gtts \\ 
		& 5003 & prednisolone \\
		& 1878 & methylprednisolone \\
		& 2656 & SOLUMEDROL MG/KG/HR \\
		& 2657 & SOLUMEDROL CC/H \\
		& 2629 & SOLUMEDROL DRIP \\
		& 2983 & solumedrol mg/hr \\
		& 7425 & Solu-medrol mg/hr \\
		& 6323 & solumedol \\
		& 7592 & Solumedrol cc/h \\
		& 30069 & Solumedrol \\
		& 2959 & Solumedrolmg/kg/hr \\
		& 1878 & methylprednisolone \\
		& 5395 & Beclamethasone \\
		& 4542 & Tobradex \\
		& 5612 & Dexamethasone gtts \\
		& 3463 & Hydrocortisone \\
		\multirow{-22}{*}{Intravenous Steroid} & 8070 & dexamethasone gtts \\ 
		\bottomrule
	\end{tabular}
\end{table*}

\begin{table*}[ht!]
	\centering
	\footnotesize
	\caption{Disease information of MIMIC III - Heart Failure dataset}
	\label{hf_label}
	\begin{tabular}{cc}
		\toprule
		\multicolumn{1}{c}{\textbf{Task Name}} & 
		\multicolumn{1}{c}{\textbf{ICD9 Disease Code}} \\ \hline
        & 412, 4110, 4148, 4149, 41000, 41001, 41002, 41010, 41011, 41012, \\
        & 41020, 41021, 41022, 41030, 41031, 41032, 41040, 41041, 41042, 41050, \\
        & 41051, 41052, 41060, 41061, 41062, 41070, 41071, 41072, 41080, 41082, \\
        \multirow{-4}{*}{Ischemic Heart Disease (IHD)} & 41090, 41091, 41092, 41181, 41189, 41406, 41407 \\  \hline
        & 3940, 3942, 3949, 3952, 3960, 3961, 3962, 3963, 3968, 3969, \\ 
		\multirow{-2}{*}{Valvular Heart Disease (VHD)} & 3970, 3971, 4240, 4241, 4242, 4243, V422, V433 \\ \hline
		& 4280, 4281, 39831, 40201, 40211, 40291, 40401, 40403, 40411, 40413, \\ 
		& 40491, 40493, 42820, 42821, 42822, 42823, 42830, 42831, 42832, 42833, \\
		\multirow{-3}{*}{Heart Failure (HF)} & 42840, 42841, 42842, 42843 \\ \hline
		Mortality & NA \\ 
		\bottomrule
	\end{tabular}
\end{table*}

\begin{table*}[ht!]
	\centering
	\tiny
	\caption{Feature information of MIMIC III - Heart Failure dataset}
	\label{hf_feature}
	\begin{tabular}{ccc}
		\toprule
		\multicolumn{1}{c}{\textbf{Features}} & 
		\multicolumn{1}{c}{\textbf{Item-ID}} & 
		\multicolumn{1}{c}{\textbf{Name of Item}} \\ \hline
		Age & NA & \begin{tabular}[c]{@{}l@{}}initime\\ dob\end{tabular} \\ \hline
		Sex & NA & gender \\ \hline
		& 211 & Heart Rate \\
		\multirow{-2}{*}{Heart Rate} & 22045 & Heart Rate \\ \hline
		& 618 & Respiratory Rate \\
		& 619 & Respiratory Rate \\
		& 220210 & Respiratory Rate \\
		& 224688 & Respiratory Rate \\
		& 224689 & Respiratory Rate \\
		\multirow{-6}{*}{Respiratory Rate}& 224690 & Respiratory Rate \\ \hline
		& 51 & Systolic Blood Pressure \\
		& 442 & Systolic Blood Pressure \\
		& 455 & Systolic Blood Pressure \\
		& 6701 & Systolic Blood Pressure \\
		& 220179 & Systolic Blood Pressure \\
		\multirow{-6}{*}{Systolic Blood Pressure} & 220050 & Systolic Blood Pressure \\ \hline
		& 8368 & Diastolic Blood Pressure \\
		& 8440 & Diastolic Blood Pressure \\
		& 8441 & Diastolic Blood Pressure \\
		& 8555 & Diastolic Blood Pressure \\
		& 220051 & Diastolic Blood Pressure \\
		\multirow{-5}{*}{Diastolic Blood Pressure} & 220180 & Diastolic Blood Pressure \\ \hline
		& 676 & Body Temperature \\  
		& 677 & Body Temperature \\
		& 8537 & Body Temperature \\
		& 223762 & Body Temperature \\
		\multirow{-5}{*}{Body Temperature} & 226329 & Body Temperature \\ \hline
		& 189 & Fi$O_2$ \\
		& 190 & Fi$O_2$ \\
		& 2981 & Fi$O_2$ \\
		& 3420 & Fi$O_2$ \\
		& 3422 & Fi$O_2$ \\
		\multirow{-6}{*}{Fraction of inspired oxygen (Fi$O_2$)} & 223835 & Fi$O_2$  \\ \hline
		& 823 & $S_vO_2$ \\
		& 2396 & $S_vO_2$ \\
		& 2398 & $S_vO_2$ \\
		& 2574 & $S_vO_2$ \\
		& 2842 & $S_vO_2$ \\
		& 2933 & $S_vO_2$ \\
		& 2955 & $S_vO_2$ \\
		& 3776 & $S_vO_2$ \\
		& 5636 & $S_vO_2$ \\
		& 6024 & $S_vO_2$ \\
		& 7260 & $S_vO_2$ \\
		& 7063 & $S_vO_2$ \\
		& 7293 & $S_vO_2$ \\
		& 226541 & $S_vO_2$ \\
		& 227685 & $S_vO_2$ \\
		& 225674 & $S_vO_2$ \\
		\multirow{-17}{*}{Mixed venous Oxygen Saturation ($S_vO_2$)} & 227686 & $S_vO_2$ \\ \hline
		& 834 & $S_aO_2$ \\
		& 3288 & $S_aO_2$ \\
		& 8498 & $S_aO_2$ \\
		\multirow{-4}{*}{Oxygen Saturation of arterial blood ($S_aO_2$)} & 220227 & $S_aO_2$ \\ \hline
		& 7294 & BNP \\
		& 227446 & BNP \\
		\multirow{-3}{*}{Brain Natriuretic Peptide (BNP)} & 225622 & BNP \\ \hline
		Ejection Fraction (EF) & 227008 & EF \\ \hline		
		Glasgow Coma Scale (GCS) - Verbal Response & 223900 & GCS-Verbal Response \\ \hline
		Glasgow Coma Scale (GCS) - Motor Response & 223901 & GCS-Motor Response \\ \hline
		Glasgow Coma Scale (GCS) - Eye Opening & 220739 & GCS-Eye Opening \\ 
		\bottomrule
	\end{tabular}
\end{table*}

\begin{table*}[ht!]
	\centering
	\footnotesize
	\caption{Disease information of MIMIC III - Respiratory Failure dataset}
	\label{rf_label}
	\begin{tabular}{ccc}
		\toprule
		\multicolumn{1}{c}{\textbf{Task Name (Clinical Condition)}} & 
		\multicolumn{1}{c}{\textbf{Item-ID}} & 
		\multicolumn{1}{c}{\textbf{Name of Item}} \\ \hline
		& 779 & Arterial $P_aO_2$ \\
		& 3785 & $P_aO_2$ (ABG's) \\
		& 3837 & $P_aO_2$ (ABG's) \\
		& 3838 & $P_aO_2$ (other) \\
		& 4203 & $P_aO_2$ (cap) \\
		\multirow{-6}{*}{Hypoxia ($P_aO_2 < 60mmHg$)}& 220224 & $P_aO_2$ (Arterial $O_2$ Pressure) \\ \hline
		& 778 & Arterial $P_aCO_2$ \\
		& 3784 & $P_aCO_2$ (ABG's) \\
		& 3835 & $P_aCO_2$ \\
		& 3836 & $P_aCO_2$ (other) \\
		& 4201 & $P_aCO_2$ (cap) \\
		\multirow{-6}{*}{Hypercapnia (Hypercarbia, $P_aCO_2 > 50mmHg$)}& 220224 & $P_aCO_2$ (Arterial $CO_2$ Pressure) \\ \hline
		VQ Mismatch & 26 & Alveolar-arterial oxygen gradient ($A_aDO_2$) \\ \hline
		& 780 & Arterial pH \\
		& 1126 & Art.pH \\
		& 3839 & ph (other) \\
		& 4202 & ph (cap) \\
		\multirow{-5}{*}{Acidosis (Arterial $pH < 7.35$)}& 4753 & ph (Art) \\ \hline
		& 51881 & Acute Respiratory Failure (ARF) \\
		& 51883 & Chronic Respiratory Failure (CRF) \\
		\multirow{-3}{*}{Respiratory Failure (RF) }& 51884 & ARF, CRF \\ \hline
		& 834 & $S_aO_2$ \\
		& 3288 & $O_2$ sat [Pre] \\
		& 4833 & $S_aO_2$ (post) \\
		& 8498 & $O_2$ sat [Post] \\
		\multirow{-5}{*}{Cyanosis: Patient requiring $O_2$ infusion ($O_2$ Saturation($S_aO_2$) $< 94$)}& $220227$ & Arterial $O_2$ Saturation \\ \hline
		Mortality & NA \\ 
		\bottomrule
	\end{tabular}
\end{table*}

\begin{table*}[ht!]
	\centering
	\tiny
	\caption{Feature information of MIMIC III - Respiratory Failure dataset}
	\label{rf_feature}
	\begin{tabular}{ccc}
		\toprule
		\multicolumn{1}{c}{\textbf{Features}} &
		\multicolumn{1}{c}{\textbf{Item-ID}} &
		\multicolumn{1}{c}{\textbf{Name of Item}} \\ \hline
		Age & NA & \begin{tabular}[c]{@{}l@{}}initime\\ dob\end{tabular} \\ \hline
		Sex & NA & gender \\ \hline
		& 211 & Heart Rate \\
		\multirow{-2}{*}{Heart Rate} & 22045 & Heart Rate \\ \hline
		& 618 & Respiratory Rate \\
		& 619 & Respiratory Rate \\
		& 220210 & Respiratory Rate \\
		& 224688 & Respiratory Rate \\
		& 224689 & Respiratory Rate \\
		\multirow{-6}{*}{Respiratory Rate}& 224690 & Respiratory Rate \\ \hline
		& 51 & Systolic Blood Pressure \\
		& 442 & Systolic Blood Pressure \\
		& 455 & Systolic Blood Pressure \\
		& 6701 & Systolic Blood Pressure \\
		& 220179 & Systolic Blood Pressure \\
		\multirow{-6}{*}{Systolic Blood Pressure} & 220050 & Systolic Blood Pressure \\ \hline
		& 8368 & Diastolic Blood Pressure \\
		& 8440 & Diastolic Blood Pressure \\
		& 8441 & Diastolic Blood Pressure \\
		& 8555 & Diastolic Blood Pressure \\
		& 220051 & Diastolic Blood Pressure \\
		\multirow{-5}{*}{Diastolic Blood Pressure} & 220180 & Diastolic Blood Pressure \\ \hline
		& 676 & Body Temperature \\
		& 677 & Body Temperature \\
		& 8537 & Body Temperature \\
		& 223762 & Body Temperature \\
		\multirow{-5}{*}{Body Temperature} & 226329 & Body Temperature \\ \hline
		Bicarbonate ($HCO_3^-$) & 812 & $HCO_3^-$ \\ \hline
		Base Excess (BE) & 812 & BE \\ \hline
		& 189 & Fi$O_2$ \\
		& 190 & Fi$O_2$ \\
		& 2981 & Fi$O_2$ \\
		& 3420 & Fi$O_2$ \\
		& 3422 & Fi$O_2$ \\
		\multirow{-6}{*}{Fraction of inspired oxygen (Fi$O_2$)} & 223835 & Fi$O_2$  \\ \hline
		& 823 & $S_vO_2$ \\
		& 2396 & $S_vO_2$ \\
		& 2398 & $S_vO_2$ \\
		& 2574 & $S_vO_2$ \\
		& 2842 & $S_vO_2$ \\
		& 2933 & $S_vO_2$ \\
		& 2955 & $S_vO_2$ \\
		& 3776 & $S_vO_2$ \\
		& 5636 & $S_vO_2$ \\
		& 6024 & $S_vO_2$ \\
		& 7260 & $S_vO_2$ \\
		& 7063 & $S_vO_2$ \\
		& 7293 & $S_vO_2$ \\
		& 226541 & $S_vO_2$ \\
		& 227685 & $S_vO_2$ \\
		& 225674 & $S_vO_2$ \\
		\multirow{-17}{*}{Mixed venous Oxygen Saturation ($S_vO_2$)} & 227686 & $S_vO_2$ \\ \hline
		Partial Pressure of Oxygen in the Alveoli ($P_AO_2$) & 490 & $P_AO_2$ \\ \hline
		Arterial Oxygen Content ($C_aO_2$) & 114 & $C_aO_2$ \\ \hline
		Venous Oxygen Content ($C_vO_2$) & 143 & $C_vO_2$ \\ \hline
		& 1390 & $DO_2$ \\
		& 1391 & $DO_2$ \\
		\multirow{-3}{*}{Delivered Oxygen ($DO_2$)} & 2740 & $DO_2$ \\ \hline
		& 504 & $PCWP$ \\
		\multirow{-2}{*}{Pulmonary Capillary Wedge Pressure (PCWP)} & 223771 & PCWP \\ \hline
		Positive End-Expiratory Pressure (PEEP) & 505 & PEEP \\
		\bottomrule
	\end{tabular}
\end{table*}

\begin{table*}[ht!]
	\centering
	\footnotesize
	\caption{Feature information of MIMIC III - Respiratory Failure dataset: continued}
	\begin{tabular}{ccc}
		\toprule
		\multicolumn{1}{c}{\textbf{Features}} & 
		\multicolumn{1}{c}{\textbf{Item-ID}} & 
		\multicolumn{1}{c}{\textbf{Name of Item}} \\ \hline
		& 814 & $Hb$ \\
		\multirow{-2}{*}{Hemoglobin ($Hb$)} & 220228 & $Hb$ \\ \hline
		Red Blood Cell (RBC) & 223901 & RBC \\ \hline
		& 861 & WBC \\  
		& 1127 & WBC \\
		& 1542 & WBC \\
		\multirow{-4}{*}{White Blood Cell (WBC)} & 220546 & WBC \\ \hline
		& 828 & Platelet \\  
		& 30006 & Platelet \\
		\multirow{-3}{*}{Platelet} & 225170 & Platelet \\ \hline
		& 1162 & BUN \\  
		& 5876 & BUN \\
		\multirow{-3}{*}{Blood Urine Nitrogen (BUN)} & 225624 & BUN \\ \hline
		& 1525 & Creatinine \\
		\multirow{-2}{*}{Creatinine} & 220615 & Creatinine \\ \hline
		& 7294 & BNP \\
		& 227446 & BNP \\
		\multirow{-3}{*}{Brain Natriuretic Peptide (BNP)} & 225622 & BNP \\ \hline
		Ejection Fraction (EF) & 227008 & EF \\ \hline		
		Glasgow Coma Scale (GCS) - Verbal Response & 223900 & GCS-Verbal Response \\ \hline
		Glasgow Coma Scale (GCS) - Motor Response & 223901 & GCS-Motor Response \\ \hline
		Glasgow Coma Scale (GCS) - Eye Opening & 220739 & GCS-Eye Opening \\ 
		\bottomrule
	\end{tabular}
\end{table*}

\clearpage

\end{document}